\documentclass[10pt]{article} 
\usepackage[accepted]{tmlr}

\usepackage{afterpage}
\usepackage{hyperref}
\usepackage{url}
\usepackage{amsmath, amssymb, amsthm}
\usepackage{algorithm}
\usepackage{enumitem}
\usepackage{algorithmic}
\usepackage{wrapfig}
\usepackage{capt-of}
\usepackage{booktabs}
\usepackage{float}
\usepackage{comment}
\usepackage{subcaption}
\usepackage{setspace} 
\usepackage{multirow}
\usepackage{mathtools}
\usepackage[dvipsnames]{xcolor}

\usepackage[capitalize,nameinlink]{cleveref}
\usepackage[most]{tcolorbox}
\usepackage{bbm}
\usepackage{pifont} 
\usepackage{icomma}
\usepackage[%
    prefixes-as-symbols=false,
    ]{siunitx}
\usepackage{tablefootnote}
\usepackage{makecell}
\usepackage{arydshln}

\usepackage{amsmath,amsfonts,bm}

\def\eqref#1{equation~\ref{#1}}

\def\1{\bm{1}}

\DeclareMathAlphabet{\mathsfit}{\encodingdefault}{\sfdefault}{m}{sl}
\SetMathAlphabet{\mathsfit}{bold}{\encodingdefault}{\sfdefault}{bx}{n}

\newcommand{\E}{\mathbb{E}}

\newcommand{\Var}{\mathrm{Var}}

\newcommand{\Cov}{\mathrm{Cov}}

\DeclareMathOperator*{\argmax}{arg\,max}

\theoremstyle{plain}
\newtheorem{thm}{Theorem}[section]

\usepackage{romannum}

\definecolor{Red}{rgb}{0.768, 0.054, 0.054}
\definecolor{Blue}{rgb}{0.152, 0.294, 0.925}
\definecolor{Green}{rgb}{0,0.4,0.7}
\definecolor{Gray}{gray}{0.9}
\definecolor{linkgreen}{hsb}{0.85,0.85,0.85} 
\definecolor{linkgray}{HTML}{555555} 

\usepackage{xspace}
\newcommand{\sysstyle}[1]{\textsf{#1}}  

\newcommand{\DisORM}{\sysstyle{dORM}\xspace}
\newcommand{\DisPRM}{\sysstyle{dPRM}\xspace}
\newcommand{\GenORM}{\sysstyle{gORM}\xspace}
\newcommand{\GenPRM}{\sysstyle{gPRM}\xspace}

\hypersetup{
    colorlinks=true,
    citecolor=teal,
    linkcolor=Red,
    urlcolor=Green,
    }

\tcbset{
    llmprompt/.style={
        colback=gray!5,
        colframe=gray!40,
        width=\textwidth,
        boxrule=0.8pt,
        arc=2mm,
        left=3mm,
        right=3mm,
        top=2mm,
        bottom=2mm
    }
}

\AtBeginDocument{\pagenumbering{arabic}}

\DeclarePairedDelimiterX{\infdivx}[2]{(}{)}{%
  #1\;\delimsize\|\;#2%
}

\newcommand{\ie}{\textit{i.e.}}
\newcommand{\cf}{\textit{cf.}}
\newcommand{\eg}{\textit{e.g.}}

\crefname{table}{Table}{Tables}
\Crefname{figure}{Fig.}{Figs.}
\Crefname{equation}{Eq.}{Eqs.}
\Crefformat{section}{#2\S#1#3}
\Crefrangeformat{section}{#3\S#1#4--#5\S#2#6}
\Crefmultiformat{section}{#2\S#1#3}{ and #2\S#1#3}{, #2\S#1#3}{, and #2\S#1#3}
\Crefname{thm}{Theorem}{Theorem}
\creflabelformat{equation}{#2(#1)#3}            
\crefrangelabelformat{equation}{#3(#1)#4--#5(#2)#6} 

\crefname{appendix}{Appendix}{Appendices}
\Crefname{appendix}{Appendix}{Appendices}
\crefformat{appendix}{#2Appendix~#1#3}
\Crefformat{appendix}{#2Appendix~#1#3}
\crefrangeformat{appendix}{#3Appendices~#1#4--#5#2#6}
\Crefrangeformat{appendix}{#3Appendices~#1#4--#5#2#6}

\usepackage[table]{xcolor}
\usepackage{array,colortbl}
\definecolor{Gray}{gray}{0.92} 
\newcolumntype{g}{>{\columncolor{Gray}}c} 

\title{Rethinking Reward Models for Multi-Domain Test-Time Scaling}

\author{%
      Dong Bok Lee\textsuperscript{1,*,\dag},
      Seanie Lee\textsuperscript{1,*},
      Sangwoo Park\textsuperscript{1},
      Minki Kang\textsuperscript{1},
      Jinheon Baek\textsuperscript{1}, \\
      Dongki Kim\textsuperscript{1},
      Dominik Wagner\textsuperscript{3},
      Jiongdao Jin\textsuperscript{1},
      Heejun Lee\textsuperscript{1,4},
      Tobias Bocklet\textsuperscript{3}, \\
      Jinyu Wang\textsuperscript{2},
      Jingjing Fu\textsuperscript{2},
      Sung Ju Hwang\textsuperscript{1,4},
      Jiang Bian\textsuperscript{2},
      Lei Song\textsuperscript{2} \\
      \addr \textsuperscript{1}KAIST
      \quad \textsuperscript{2}Microsoft Research Asia
      \quad \textsuperscript{3}TH N\"{u}rnberg
      \quad \textsuperscript{4}DeepAuto.ai \\
      \addr \textsuperscript{*}Equal contribution. Correspondence to \texttt{markhi@kaist.ac.kr}. \\
      \quad \textsuperscript{\dag}Work done during an internship at Microsoft Research Asia.
}

\def\month{07}  
\def\year{2026} 
\def\openreview{\url{https://openreview.net/forum?id=PgouBhL7IR}} 

\begin{document}

\maketitle
\begin{abstract}
The reliability of large language models (LLMs) during test-time scaling is often assessed with \emph{external verifiers} or \emph{reward models} that distinguish correct reasoning from flawed logic. 
Prior work has studied both outcome reward models (ORMs), which assess only the final answer, and process reward models (PRMs), which score intermediate reasoning steps. Although PRMs are often viewed as advantageous due to their finer-grained supervision, much of the supporting evidence comes from math-adjacent settings, and their relative benefits across broader domains remain unclear.
We present the first unified evaluation of four reward model variants, discriminative ORM and PRM (\DisORM, \DisPRM) and generative ORM and PRM (\GenORM, \GenPRM), across 14 diverse domains. Contrary to conventional wisdom, we find that (i) \DisORM performs on par with \DisPRM, (ii) \GenPRM is not competitive, and (iii) overall, \GenORM is the most robust, yielding significant and consistent gains across every tested domain.
We attribute the worse performance of \GenPRM to the stepwise scoring process, which inherits label noise from LLM-based automatic labeling, leading to difficulties in evaluating long reasoning trajectories, including those involving self-correcting reasoning. 
Both our theoretical analysis and empirical observations indicate that stepwise aggregation compounds errors as reasoning length increases.
These findings challenge the common assumption that fine-grained supervision is always better and support generative outcome verification for multi-domain deployment. 
Our \href{https://github.com/db-Lee/Multi-RM}{\underline{code}} is publicly available to facilitate future research in multi-domain settings.

\end{abstract}
\section{Introduction}\label{sec:introduction}
Test-time scaling (TTS) enables large language models (LLMs) to generate diverse, reliable solutions via chain-of-thought reasoning \citep[CoT;][]{wei2022chain, kojima2022large, yao2023react, madaan2023self} and has shown strong results on challenging reasoning tasks~\citep{yao2023tree, snell2024scaling, wu2024inference}.
A widely adopted TTS approach uses \emph{external verifiers} that select the best among the candidates~\citep{snell2024scaling}. 
A common external verifier is the outcome reward model (ORM), typically implemented as a discriminative classifier that assigns a scalar \emph{reward} to a CoT~\citep{cobbe2021training, uesato2022solving, yu2023ovm}.
ORMs are trained only on outcome-level signals, providing a single label per trajectory rather than feedback at individual reasoning steps. 
Recent work has introduced process reward models \citep[PRMs;][]{prm800k, wang2024mathshepherd, setlur2024rewarding, zheng2024processbench} that score each step of a CoT and aggregate the scores into a trajectory-level reward. Supervised with high-quality, carefully constructed process labels, \eg, manual annotation~\citep{prm800k} or Monte Carlo rollouts~\citep{wang2024mathshepherd}, PRMs have been shown to outperform ORMs as TTS verifiers.

Beyond discriminative verifiers, several studies have shown that the generative ability of LLMs can improve CoT verification, such as \emph{LLM-as-a-judge}~\citep{wang2023chatgpt, liu2023g, zheng2023judging}. Based on this idea, other works fine-tune LLMs to generate a verification rationale for a CoT and compute the final reward from token probabilities~\citep{zhang2025generativeverifiers,thinkprm,zhao2025genprm}. To obtain verification CoTs for training, most previous work adopts \emph{consensus-filtering}: (i) generate verification CoTs, and (ii) retain the verification CoT if its parsed verdict aligns with outcome or process labels. After training, these generative verifiers have shown strong performance in math-adjacent reasoning tasks, outperforming discriminative verifiers.

However, external verifiers for TTS have been studied primarily in math-adjacent domains. 
This narrow scope limits the potential for LLM deployment in high-stakes real-world applications, such as the legal~\citep{guha2023legalbench,cui2023chatlaw,fei2023lawbench} and medical~\citep{singhal2023large,kung2023performance,singhal2025toward} domains, where trustworthiness is paramount and rigorous verification of LLM outputs is especially important. Recently, \cite{versaprm} proposed multi-domain PRMs trained on the graduate-level benchmark \citep[MMLU-Pro;][]{mmlu-pro}, covering 14 diverse domains, and showed that multi-domain training for PRMs significantly improves TTS performance across diverse domains. However, the study is \emph{limited} to discriminative PRMs and the broader potential of different verifier types (\eg, ORMs vs.\ PRMs, discriminative vs.\ generative) in the multi-domain setting remains \emph{underexplored}.

\begin{wrapfigure}{r}{0.50\textwidth}
\vspace{-0.15in}
\centering
\includegraphics[width=\linewidth]{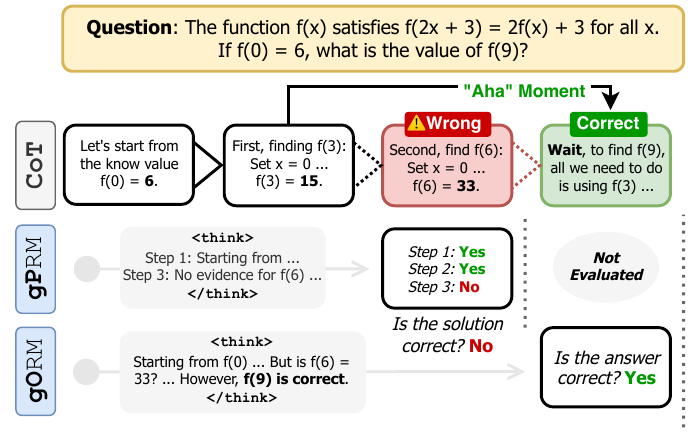}
\vspace{-0.275in}
\caption{\small \textbf{Evaluating CoTs using} \GenORM \textbf{and} \GenPRM.}
\label{fig:aha_reward_models}
\end{wrapfigure}
To this end, we present the first controlled multi-domain TTS evaluation of four verifier variants, discriminative ORM and PRM (\DisORM, \DisPRM), and generative ORM and PRM (\GenORM, \GenPRM), trained and evaluated under a common protocol across 14 diverse domains. We review these variants in \Cref{sec:related_work} and, under controlled conditions, evaluate them on math \citep[PRM800K, ProcessBench;][]{prm800k, zheng2024processbench}, multi-domain \citep[MMLU-Pro;][]{mmlu-pro}, and specialized-domain \citep[GPQA-Diamond, MedQA, LEXam;][]{rein2024gpqa, jin2021disease, fan2025lexam} in \Cref{sec:experiments}. In the math domain, trends across the four variants are consistent with prior work~\citep{prm800k,zhang2025generativeverifiers,thinkprm}. \DisPRM outperforms \DisORM, and generative variants outperform discriminative ones. In the multi-domain and specialized domain setting, however, we observe contrasting results. \DisORM performs on par with \DisPRM, \GenPRM is not competitive, and overall, \GenORM delivers \textbf{consistent and significant gains} over the others.  

In \Cref{sec:analysis}, we identify two factors underlying the weaker performance of \GenPRM.
First, on more difficult multi-domain problems, LLMs tend to produce longer CoTs that PRMs struggle to evaluate.
As illustrated in~\Cref{fig:aha_reward_models}, stepwise aggregation in PRMs often fails to reward long CoTs that recover from earlier errors \citep[``aha'' moments;][]{guo2025deepseek}, because verification stops at the first mistake. 
In \Cref{sec:cot_length}, we analyze how this stepwise aggregation compounds errors as the chain length increases, and confirm this effect empirically.
Second, label noise is prevalent in multi-domain datasets.
Given that step annotation in specialized domains is costly, prior work such as \citet{versaprm} depends on LLM-based auto-labeling, which can introduce noise.
In \Cref{sec:label_noise}, using a simulated label-noise analysis in the math domain, we show that \DisPRM is particularly sensitive to noisy step labels, whereas \GenORM remains robust.
Although \GenPRM is robust to label noise in the math domain, it degrades in the multi-domain setting. We attribute this degradation to a severe shift in the CoT-length distribution induced by consensus filtering. Based on this analysis, we present practical guidelines for selecting among the four variants and discuss limitations and future work in \Cref{sec:conclusion}.

Our contributions and findings are summarized as follows:
\begin{itemize}[itemsep=1mm,parsep=1pt,topsep=2pt,leftmargin=*]
\item We present the first controlled multi-domain TTS evaluation of four verifier variants (\DisORM, \DisPRM, \GenORM, and \GenPRM) trained and evaluated under a common protocol across 14 diverse domains.

\item Contrary to conventional wisdom established in math-adjacent settings, we observe that (i) \DisORM performs similarly to \DisPRM, (ii) \GenPRM is not competitive, and (iii) overall, \GenORM \textbf{delivers consistent gains} over the others.

\item To explain the empirical observations, we provide two perspectives: (i) a theoretical analysis, with empirical support, showing that \emph{PRM risk increases with CoT length}, and (ii) evidence of \emph{process label risk} in the multi-domain setting with length-distribution shift induced by \emph{consensus filtering}.
\end{itemize}

\section{Background and Related Work}\label{sec:related_work}
\begin{figure}[t]
\centering
\includegraphics[width=0.95\textwidth]{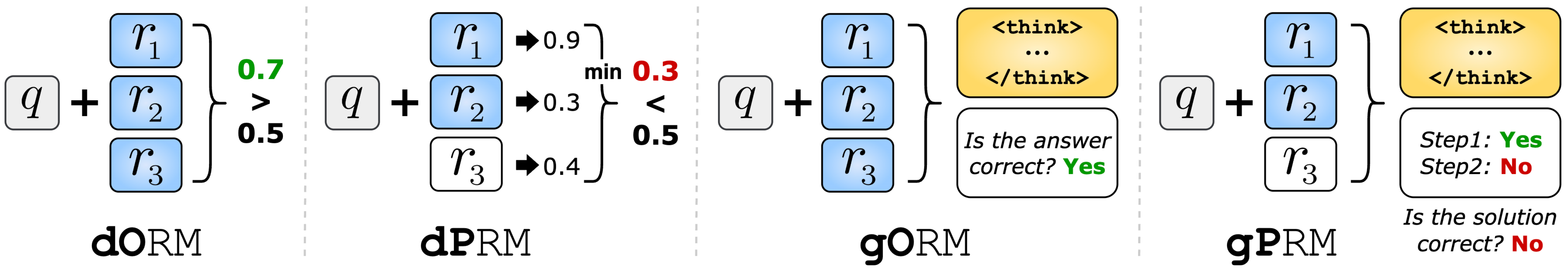}
\caption{\small \textbf{Conceptual illustration of reward models}: $r_2$ is the first incorrect step; the final answer is correct.
}
\label{fig:summary_of_reward_models}
\end{figure}

In this section, we review background and related work. We first formalize notation and test-time scaling in \Cref{sec:problem_formulation}, and then discuss reward-model variants in \Cref{sec:reward_models}, summarized in \Cref{fig:summary_of_reward_models}.

\subsection{Problem Formulation}\label{sec:problem_formulation}
\paragraph{Notation.}
For a given question $q$ with the corresponding ground-truth (GT) answer $a$, we leverage the reasoning ability of LLMs to predict $a$ by generating a CoT, \ie, $r_{1:T}\coloneqq(r_1,\ldots,r_T) \sim p_{\mathrm{LLM}}(\cdot \mid q)$. Following \citet{versaprm}, we segment the reasoning steps $r_{1:T}$ using the delimiter ``\texttt{\textbackslash n\textbackslash n}'', where $T$ is the number of reasoning steps. 
Let $x\coloneqq(q,r_{1:T})\in\mathcal{X}$, where $\mathcal{X}$ denotes the space of questions and reasoning chains, and let $x_{1:t}\coloneqq(q,r_{1:t})$ be the prefix up to the $t$-th step.
We consider two types of labels: (1) the \emph{outcome label} $y=\mathbbm{1}(\hat{a}(r_T)=a)\in\{0,1\}$, where $\hat{a}(r_T)$ is the predicted answer retrieved from the last reasoning step $r_T$ and $\mathbbm{1}$ is the indicator function; and (2) the \emph{process labels} $z_{1:T}=(z_1,\ldots,z_{T})\in\{0,1\}^{T}$, where each $z_t$ indicates whether the corresponding reasoning step $r_t$ is correct. Note that $y$ represents the correctness label for the last reasoning step, so $y=z_T$. 
We use $y$ (and $z_t$) to denote both the label and the corresponding random variable, where the event $y=1$ corresponds to the model generating the positive verdict token (\eg, ``\texttt{Yes}''). Thus, expressions such as $p(y=1\mid x)$ denote the predicted probability of this event and are well-defined at test time without access to the true label.

\paragraph{TTS with reward models.}
Reward models have many applications, including LLM training via reinforcement learning~\citep{rl1,rl2,achiam2023gpt,llama3,team2024gemma,yang2025qwen3}, preference labeling~\citep{preference1,preference2,preference3}, rejection sampling~\citep{rejection1,rejection2}, and data filtering~\citep{llama3,albalak2024survey,yang2025qwen3}.
In this work, we focus on parallel or sampling-based~\citep{wu2024inference} TTS with reward models, such as Best-of-$N$ \citep[Bo$N$;][]{charniak-johnson-2005-coarse,khalifa2023grace,snell2024scaling}, which allocates more compute at test time (\ie, generates $N$ CoTs) and selects the candidate $\hat{a}(r^{(i_\star)}_T)$ with the highest \emph{reward}:
\begin{equation}
  i_\star=\argmax_{i \in \{1,\ldots,N\}} f(x^{(i)}),
\quad \text{where } 
x^{(i)} \coloneqq (q, r_{1:T}^{(i)}), 
\;\text{and }\;
r_{1:T}^{(i)} \overset{\text{i.i.d.}}{\sim} p_{\mathtt{LLM}}(\cdot \mid q).
\label{eq:problem_formulation}
\end{equation}
Here, $f : \mathcal{X} \to [0,1]$ is the \emph{true (unknown) reward function} that assigns higher scores to CoTs that yield more reasonable and correct answers. However, $f$ is unknown, so we train an \emph{external verifier} $\hat f : \mathcal{X} \to [0,1]$ to approximate $f$ and use $\hat f$ as a surrogate in \Cref{eq:problem_formulation}, which is detailed in \Cref{sec:reward_models}.

\subsection{Reward Models}\label{sec:reward_models}
\paragraph{Discriminative outcome reward model (\DisORM).} 
Early studies on reward models~\citep{cobbe2021training,uesato2022solving,yu2023ovm} train a binary classifier $\hat f_{\DisORM}:\mathcal{X}\mapsto[0,1]$ on \emph{outcome labels $y\in \{0,1\}$ only}, without requiring the intermediate process labels $(z_1,\ldots, z_{T-1})$. Specifically, they sample CoTs and answers for given questions, construct a training dataset $\mathcal{D}_{\DisORM}\coloneqq\{(x,y)\}$, and train $\hat f_{\DisORM}$ with the binary cross-entropy (BCE) loss to approximate true $p(y=1\mid x)$:
\begin{equation}
\mathcal{L}_{\DisORM}\coloneqq\frac{1}{|\mathcal{D}_{\DisORM}|}\sum_{(x,y)\in\mathcal{D}_{\DisORM}}\ell_\mathtt{BCE}\left(\hat f_{\DisORM}(x), y\right),
\label{eq:disorm_train}
\end{equation}
with $\ell_\mathtt{BCE}(x,y)=-\bigl[y\log x + (1-y)\log(1-x)\bigr]$. It is important to note that while \DisORM takes the full chain-of-thought as input and can leverage intermediate reasoning to predict final correctness, its fundamental characteristic is \emph{outcome-level supervision and global scoring}. Despite this coarser supervision, \citet{uesato2022solving} demonstrated that outcome-based feedback achieves comparable final-answer accuracy to process supervision when training the generator LLM $p_\mathtt{LLM}$. While their focus is on policy training rather than external test-time verification, their finding highlights the strong baseline efficacy of outcome signals.

\paragraph{Discriminative process reward model (\DisPRM).} \DisPRM seeks to improve the reward signal by training on fine-grained feedback for intermediate reasoning steps, \ie, \emph{process labels} $z_{1:T}$. For \DisPRM, the quality of these labels is the primary factor. Accordingly, prior work has proposed collecting process labels for sampled CoTs via manual annotation~\citep{prm800k}, Monte Carlo (MC) rollouts~\citep{wang2024mathshepherd}, automatically generated labels from LLMs~\citep{versaprm}, or combinations thereof~\citep{zhang2025lessons}. After collecting the process labels, we construct the training set $\mathcal{D}_{\DisPRM}\coloneqq\{(x,z_{1:T})\}$ and train $\hat f_{\DisPRM}$ using the BCE loss at each step:
\begin{equation}
\mathcal{L}_{\DisPRM}
\coloneqq\frac{1}{|\mathcal{D}_{\DisPRM}|}\sum_{(x,z_{1:T})\in\mathcal{D}_{\DisPRM}}\frac{1}{{T}^\prime} \sum_{t=1}^{{T}^\prime}
\ell_\mathtt{BCE}\left(\hat f_{\DisPRM}(x_{1:t}),z_t\right),
\label{eq:disprm_train}
\end{equation}
where ${T}^\prime$ is the index of the first incorrect reasoning step, \ie, ${T}^\prime \coloneqq \min(\{t \in \{1,\dots,T\}: z_t = 0\}\cup\{T\})$. Training up to the ${T}^\prime$-th step reflects a common assumption in the literature~\citep{prm800k,wang2024mathshepherd,zheng2024processbench,versaprm}: once a reasoning step is incorrect, \emph{subsequent steps are also incorrect}, \ie, if $z_t=0$ then $z_{t'}=0$ for all $t^{\prime}\in\{t+1, \ldots,T\}$. 
At test time, we approximate $f$ in \Cref{eq:problem_formulation} by aggregating the step rewards with the \emph{minimum}\footnote{\Cref{tab:mmlu_pro_aggregation_smollm3_3b,tab:mmlu_pro_aggregation_qwen2_5_7b,tab:mmlu_pro_aggregation_llama3_1_8b,tab:mmlu_pro_aggregation_gemma2_9b,tab:mmlu_pro_aggregation_llama3_1_70b} show that minimum, average, product, and last-step aggregation yield only marginal differences, with minimum aggregation slightly outperforming the others overall, consistent with \citet{versaprm}.}~\citep{versaprm}. Thus, in contrast to the global scoring of ORMs, the defining characteristic of PRMs is \emph{process-level supervision and step-score aggregation}.

\paragraph{LLM-as-a-judge.} \citet{wang2023chatgpt, liu2023g, zheng2023judging} show that the task-generalization ability of LLMs can extend to verification (\ie, zero-shot CoT verification). 
However, LLMs often ``overthink''~\citep{bavaresco2024llms} and, without additional training, remain practically limited~\citep{zheng2024processbench}, implying the need for LLMs explicitly trained for verification.

\paragraph{Generative outcome reward model (\GenORM).} \citet{zhang2025generativeverifiers} proposed \GenORM, trained to generate a \emph{verification CoT} together with a binary verdict, \eg, ``\texttt{Verification: Is the answer correct? Yes}'' or ``\texttt{No}''. Because GT verification CoTs are unavailable, they synthesize training data via a consensus-filtering mechanism~\citep{wang2024self,zhu2023judgelm}. We first sample a verification CoT and verdict from an LLM-as-a-judge, \ie, $v_{1:L}\sim p_{\mathtt{LLM\text{-}j}}(\cdot\mid x)$ using the prompt format in \Cref{prompt:orm_style_user}. Here, $ v_{1:L}\in\mathcal{V}^L$ denotes the verification-CoT token sequence (including the verdict tokens), $\mathcal{V}$ is the vocabulary, and let $\hat{y}\in\{0,1\}$ be the parsed verdict ($1$ for ``\texttt{Yes}'', $0$ for ``\texttt{No}''). We then include $(x, v_{1:L})$ in the training set $\mathcal{D}_{\GenORM}$ only if $\hat y$ agrees with the known outcome label $y$. We train $p_{\GenORM}$ with the next-token prediction over verification CoTs $v_{1:L}$:
\begin{equation}
\mathcal{L}_{\GenORM}\coloneqq\frac{1}{|\mathcal{D}_{\GenORM}|}\sum_{(x,v_{1:L})\in\mathcal{D}_{\GenORM}}\frac{1}{L} \sum_{i=1}^{L} -\log p_{\GenORM}(v_i\mid x,v_{<i}).
\label{eq:genorm_train}
\end{equation}
$-\log p_{\GenORM}$ is implemented as the cross-entropy loss over $\mathcal{V}$. At test time, we approximate $f$ with:
\begin{equation}
\hat f_\GenORM(x)\coloneqq \E_{v_{1:L}\sim p_\GenORM(\cdot\mid x)}\!\bigl[p_\GenORM(y=1\mid  x,v_{1:L})\bigr]\;\approx\; \frac{1}{M}\sum_{i=1}^M p_\GenORM\left(y=1\mid  x,v_{1:L}^{(i)}\right),
\label{eq:genorm_test}
\end{equation}
where $v_{1:L}^{(i)}\overset{\text{i.i.d.}}{\sim}p_\GenORM(\cdot\mid x)$. Here, the expectation is approximated with $M$ MC samples and the model’s normalized probability of predicting the verdict ``\texttt{Yes}'' at the last verdict position:
\begin{equation}
p_\GenORM(y=1\mid  v_{1:L}, x)\coloneqq \dfrac{p_\GenORM(\text{``\texttt{Yes}''}\mid x, v_{1:(L-1)})}{p_\GenORM(\text{``\texttt{Yes}''}\mid x, v_{1:(L-1)})+p_\GenORM(\text{``\texttt{No}''}\mid x, v_{1:(L-1)})}.
\label{eq:genorm_token_prob}
\end{equation}

\paragraph{Generative Process Reward Model (\GenPRM).}
Beyond \GenORM, \citet{thinkprm} proposed \GenPRM, which is trained to generate verification CoTs $v_{1:L}$ with \emph{stepwise process verdicts}, \eg, ``\texttt{Step t: The step is \textbackslash boxed\{correct\}}'' or ``\texttt{\textbackslash boxed\{incorrect\}}''. Let the predicted verdict sequence be $\hat z_{1:{T}^\prime}\in\{0,1\}^{{T}^\prime}$, defined up to the first predicted incorrect step ${T}^\prime$\footnote{As shown in \Cref{prompts:prm_style_data_generation}, when generating verification CoTs for \GenPRM (\ie, \(v_{1:L}\sim p_{\mathtt{LLM\text{-}j}}(\cdot\mid x)\)), \citet{thinkprm} instruct the LLM-as-a-judge \(p_{\mathtt{LLM\text{-}j}}\) to stop once it detects the first incorrect step.
}. Following \citet{thinkprm}, we append a final verdict prompt, yielding the token sequence $v_{1:L^+}$ by concatenating ``\texttt{Is the solution correct? Yes}'' if all predicted process labels are $1$ ($\hat z_{1:{T}^\prime}=\mathbf{1}_{{T}^\prime}$), and ``\texttt{No}'' otherwise. We then construct $\mathcal{D}_\GenPRM\coloneqq \{(x, v_{1:L^+})\}$ only when the predicted prefix agrees with the GT ($\hat z_{1:{T}^\prime}=z_{1:{T}^\prime}$). We train $p_\GenPRM$ with $v_{1:L^+}$:
\begin{equation}
\mathcal{L}_{\GenPRM}\coloneqq \frac{1}{|\mathcal{D}_{\GenPRM}|}\sum_{(x,v_{1:L^+})\in\mathcal{D}_{\GenPRM}}\frac{1}{L^+}\sum_{i=1}^{L^+}-\log p_{\GenPRM}(v_i\mid  x_{1:{T}^\prime},v_{<i}).
\label{eq:genprm_train}
\end{equation}
We condition on $x_{1:T^\prime}$ rather than the full input $x$ for training~\citep{thinkprm}, since the model $p_{\GenPRM}$ is prompted to stop verification once it reaches the first incorrect step, analogous to the data-generation process (\Cref{prompt:prm_style_user}).
At test time, consistent with \Cref{eq:genorm_test,eq:genorm_token_prob}, we approximate $f$ in \Cref{eq:problem_formulation} by sampling from $p_\GenPRM$ and computing the normalized probability of a positive final verdict:
\begin{equation}
\hat f_\GenPRM(x)\coloneqq \mathbb{E}_{v_{1:L^+}\sim p_\GenPRM(\cdot\mid x)}\!\left[p_\GenPRM( y=1\mid  x,v_{1:L^+})\right]\approx \frac{1}{M}\sum_{i=1}^M p_\GenPRM\left(y=1\mid x,v_{1:L^+}^{(i)}\right),
\label{eq:genprm_test}
\end{equation}
\begin{equation}
p_\GenPRM(y=1\mid  x, v_{1:L^+}) \coloneqq \dfrac{p_\GenPRM(\text{``\texttt{Yes}''}\mid x, v_{1:(L^+-1)})}{p_\GenPRM(\text{``\texttt{Yes}''}\mid x, v_{1:(L^+-1)})+p_\GenPRM(\text{``\texttt{No}''}\mid x, v_{1:(L^+-1)})}, 
\end{equation}
where $v_{1:L^+}^{(i)}\overset{\text{i.i.d.}}{\sim} p_\GenPRM(\cdot\mid x)$ and we now condition on the full input $x$ at test time~\citep{thinkprm}. 
Recent work has proposed more advanced \GenPRM architectures to improve generalization and mitigate evaluation artifacts, such as incorporating code verification~\citep{zhao2025genprm} or reasoning-driven generative evaluations~\citep{she2025rprm}. However, these methods are primarily optimized for the math domain and do not directly extend to multi-domain data (\eg, legal or medical domains). Therefore, we follow the approach of \citet{thinkprm} in this work.
\section{Experiments}\label{sec:experiments}
In this section, we evaluate \DisORM, \DisPRM, \GenORM, and \GenPRM in the math domain and the multi-domain setting. We describe experimental setups (\Cref{sec:experimental_setups}), and present results (\Cref{sec:experimental_results}).

\subsection{Experimental Setups}\label{sec:experimental_setups}

\paragraph{Math datasets.}
For the math domain, we use \textbf{PRM800K}~\citep{prm800k} for training, where the process labels $z_{1:T}$ are human-annotated. As a testbed, we use \href{https://huggingface.co/datasets/Qwen/ProcessBench}{\textbf{ProcessBench}}~\citep{zheng2024processbench} with four splits: GSM8K, Math, Omni-Math, and OlympiadBench. We generate $N{=}16$ CoTs per question in GSM8K and Math with \href{https://huggingface.co/Qwen/Qwen2.5-7B-Instruct}{Qwen2.5-7B-Instruct}~\citep{qwen2.5} for TTS; since we only seek to verify that a controlled evaluation reproduces prior findings, we restrict TTS to this setting.

\paragraph{Multi-domain and specialized-domain datasets.} Following \citet{versaprm}, we primarily use \textbf{MMLU-Pro}~\citep{mmlu-pro}, a 10-choice benchmark covering 14 domains. Each question is paired with 16 CoTs for training and  128 for evaluation, generated by \href{https://huggingface.co/meta-llama/Llama-3.1-8B-Instruct}{Llama-3.1-8B-Instruct}~\citep{llama3}, where process labels $z_{1:T}$ are \emph{automatically annotated} by \href{https://huggingface.co/meta-llama/Llama-3.1-70B-Instruct}{Llama-3.1-70B-Instruct}. 
To assess generalization across different $p_{\mathtt{LLM}}$, we generate $N{=}16$ CoTs per question using \href{https://huggingface.co/HuggingFaceTB/SmolLM3-3B}{SmolLM3-3B}~\citep{bakouch2025smollm3}, \href{https://huggingface.co/Qwen/Qwen2.5-7B-Instruct}{Qwen2.5-7B-Instruct}, \href{https://huggingface.co/google/gemma-2-9b-it}{gemma-2-9b-it}~\citep{team2024gemma}, and \href{https://huggingface.co/meta-llama/Llama-3.1-70B-Instruct}{Llama-3.1-70B-Instruct}. 
For broader assessment, we also include additional specialized-domain benchmarks: the graduate-level science benchmark \citep[\textbf{GPQA-Diamond};][]{rein2024gpqa}, the medical benchmark \citep[\textbf{MedQA};][]{jin2021disease}, and the legal benchmark \citep[\textbf{LEXam};][]{fan2025lexam}.

\paragraph{Implementation details.}
For the reward-model backbones, we use the \href{https://huggingface.co/collections/deepseek-ai/deepseek-r1-678e1e131c0169c0bc89728d}{R1-Distill models}~\citep{guo2025deepseek}: \href{https://huggingface.co/deepseek-ai/DeepSeek-R1-Distill-Qwen-1.5B}{Qwen-1.5B} and \href{https://huggingface.co/deepseek-ai/DeepSeek-R1-Distill-Qwen-7B}{Qwen-7B} for the math domain, and \href{https://huggingface.co/deepseek-ai/DeepSeek-R1-Distill-Llama-8B}{Llama-8B} and \href{https://huggingface.co/deepseek-ai/DeepSeek-R1-Distill-Qwen-14B}{Qwen-14B} for the multi-domain setting, respectively. We also use \href{https://huggingface.co/Qwen/Qwen3-8B}{Qwen3-8B} as the backbone to assess whether the results hold for non-distilled backbones in the multi-domain setting.
We follow \citet{zhang2025generativeverifiers} for the \GenORM prompt template (\Cref{prompt:orm_style_user}) and \citet{thinkprm} for \GenPRM (\Cref{prompt:prm_style_user}).
We optimize reward models using AdamW~\citep{adamw} with LoRA~\citep{hu2022lora}. 
For \GenORM and \GenPRM, we sample  $M = 16$ verification CoTs in the math setting and $M = 10$ in the multi-domain setting (\cf\ \Cref{eq:genorm_test,eq:genprm_test}), using \href{https://github.com/vllm-project/vllm}{vLLM}~\citep{kwon2023efficient}.
See \Cref{sec:implementation_details}, \Cref{tab:hyperparameters} and \href{https://github.com/db-Lee/Multi-RM}{this repository} for more details.

\paragraph{Verification CoTs.}
Following \citet{zhang2025generativeverifiers} and \citet{thinkprm}, we construct verification-CoT datasets for \GenORM and \GenPRM by prompting \href{https://huggingface.co/Qwen/QwQ-32B}{QwQ-32B}~\citep{qwq32b} with the formats in \Cref{prompt:orm_style_user,prompts:prm_style_data_generation}. We discard any verification CoT whose parsed labels are inconsistent with the targets (\eg, $y$ or $z_{1:T}$), corresponding to the \emph{consensus filtering} in \Cref{sec:reward_models}. The training sets of \GenORM/\GenPRM contain 34{,}286/35{,}666 and 171{,}780/94{,}156 verification CoTs for the math and multi-domain settings. See \Cref{sec:implementation_details,prompt:orm_style_example,prompt:prm_style_example} for more details and examples.

\subsection{Experimental Results}\label{sec:experimental_results}

\begin{figure}[t!]
\vspace{-0.3in}
\includegraphics[height=0.7cm]{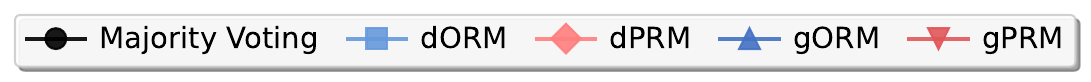}
\medskip
\vspace{-0.12in}
\centering
\includegraphics[width=1\textwidth]{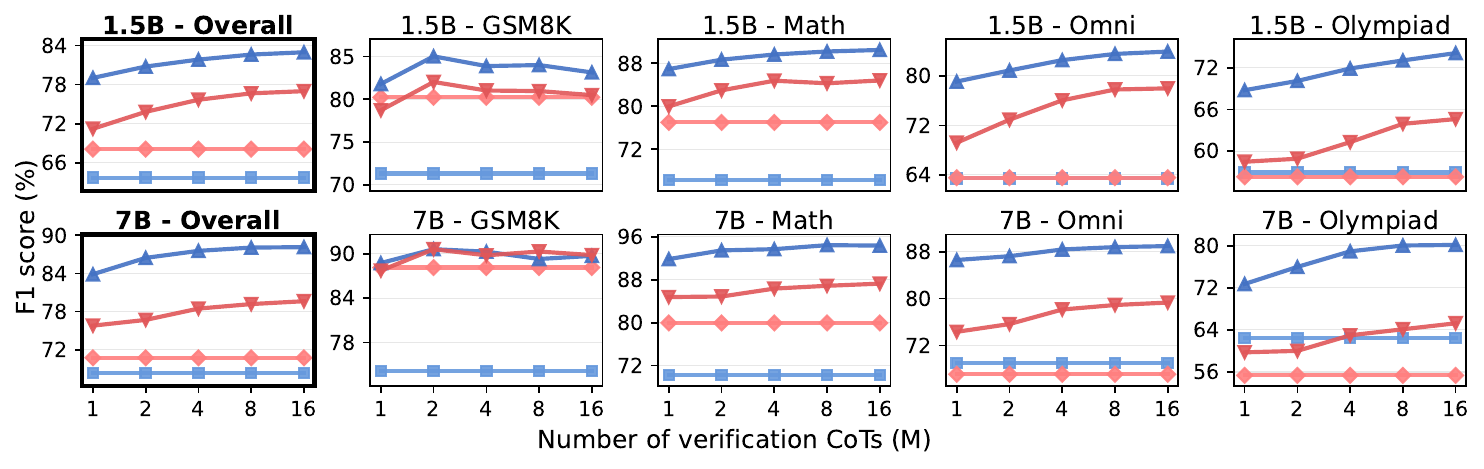}
\vspace{-0.3in}
\caption{\small \textbf{Outcome verification results} on ProcessBench in the math domain.}
\label{fig:math_outcome_verification}
\end{figure}
\begin{figure}[t!]
\centering
\includegraphics[width=0.8\textwidth]{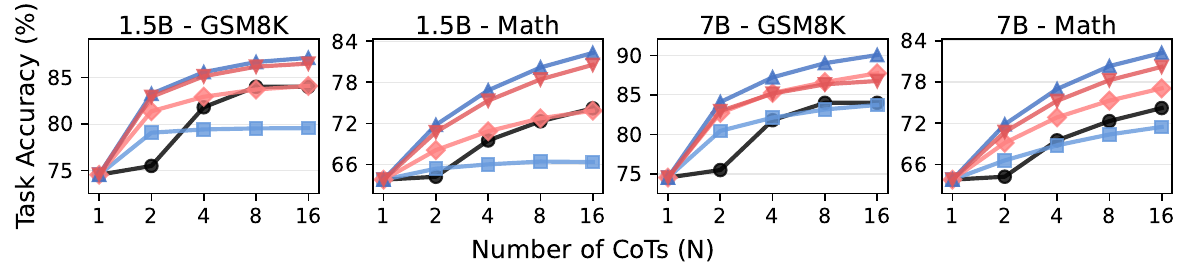}
\vspace{-0.15in}
\caption{\small \textbf{Best-of-$N$ results using \href{https://huggingface.co/Qwen/Qwen2.5-7B-Instruct}{Qwen2.5-7B-Instruct}} on GSM8K and Math in the math domain.}
\label{fig:math_tts}
\end{figure}
\paragraph{Math-domain results.}
First, we evaluate the four verifier variants in the math domain. We compare outcome-verification performance with a 0.5 decision threshold, \ie, $\hat{y}\coloneqq\mathbbm{1}(\hat{f}(x)>0.5)$. \Cref{fig:math_outcome_verification} reports F1 score (\%) on ProcessBench splits.
\DisPRM outperforms \DisORM overall, consistent with prior findings~\citep{prm800k}, and shows a slight drop in Omni-Math/OlympiadBench with 7B backbones. For \GenORM/\GenPRM, the overall performance improves with $M$. At small $M$, \GenPRM may lag behind discriminative models (\eg, OlympiadBench). \GenORM generally outperforms \GenPRM (except 7B-GSM8K), and the gap widens on Omni-Math/OlympiadBench.

\begin{figure}[t]
\vspace{-0.3in}
\centering
\includegraphics[width=1\textwidth]{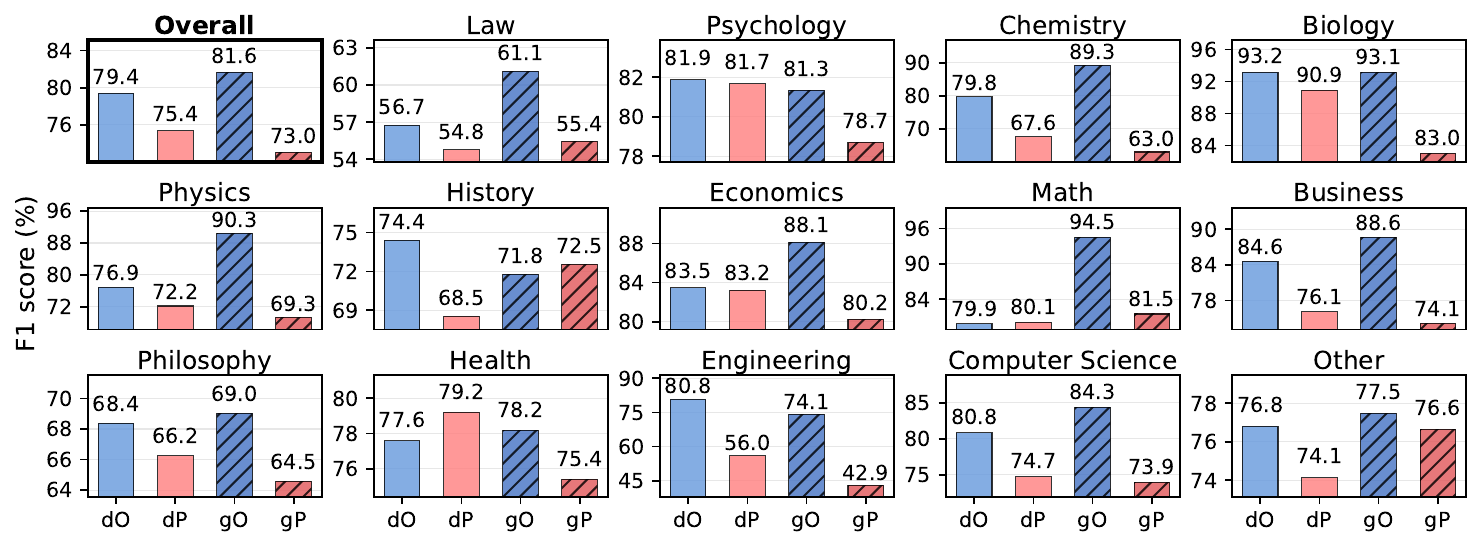}
\vspace{-0.3in}
\caption{\small \textbf{Outcome verification results} on MMLU-Pro in the multi-domain setting.}
\label{fig:outcome_verification}
\end{figure}



\begin{figure}[!t]
\centering

\includegraphics[height=0.7cm]{images/main_legend.pdf}

\begin{subfigure}{\textwidth}
    \centering
    \includegraphics[width=\textwidth]{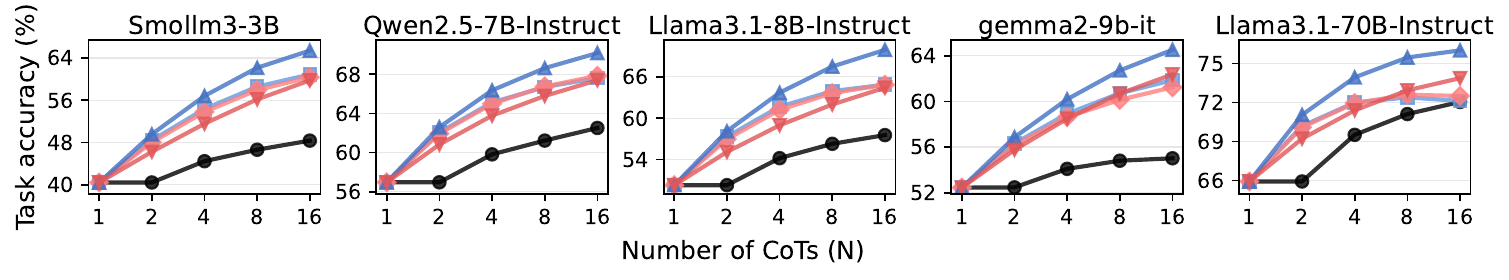}
    \caption{\textbf{MMLU-Pro}}
    \label{fig:overall_bon}
\end{subfigure}

\vspace{0.05in}

\begin{subfigure}{\textwidth}
    \centering
    \includegraphics[width=\textwidth]{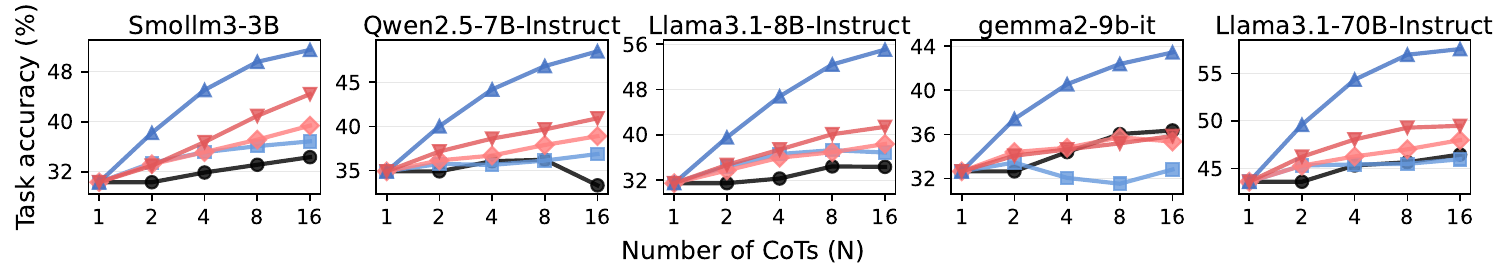}
    \caption{\textbf{GPQA-diamond}}
    \label{fig:gpqa_bon}
\end{subfigure}

\vspace{0.05in}

\begin{subfigure}{\textwidth}
    \centering
    \includegraphics[width=0.85\textwidth]{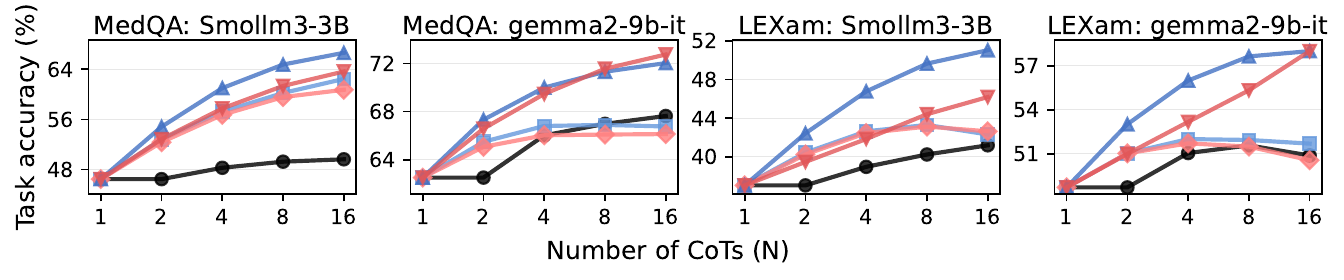}
    \caption{\textbf{MedQA and LEXam}}
    \label{fig:medqa_lexam}
\end{subfigure}

\caption{\small Best-of-$N$ results using different $p_\mathtt{LLM}$ on \textbf{MMLU-Pro}, \textbf{GPQA-diamond}, \textbf{MedQA}, and \textbf{LEXam}.}
\label{fig:overall_bon_all}
\end{figure}

Although TTS has been well studied in the math domain, evaluations are not fully controlled: (i) models are rarely compared with a shared backbone, and (ii) \GenORM and \GenPRM have not been directly compared. We therefore evaluate the reward models with Bo$N$ under controlled conditions. As shown in \Cref{fig:math_tts}, and consistent with the findings of~\citet{prm800k}, \DisPRM outperforms \DisORM. Notably, \DisORM even underperforms majority voting (MV) with 1.5B backbones, demonstrating the limitations of coarse outcome-level supervision in the math domain. In line with \citet{zhang2025generativeverifiers} and~\citet{thinkprm}, generative models outperform discriminative ones, with \GenORM slightly surpassing \GenPRM.

\paragraph{Multi-domain and specialized-domain results.}
Next, we compare the four variants in the multi-domain setting. \Cref{fig:outcome_verification} reports F1 scores (\%) for outcome-verification, with a 0.5 decision threshold, using \href{https://huggingface.co/deepseek-ai/DeepSeek-R1-Distill-Qwen-14B}{R1-Distill-Qwen-14B} as the reward model backbone. dO/dP/gO/gP denote \DisORM/\DisPRM/\GenORM/\GenPRM. 
In contrast to the math domain results in \Cref{fig:math_outcome_verification}, ORM variants achieve higher F1 scores than PRM variants.

\Cref{fig:overall_bon} shows the overall Bo$N$ performance using five different $p_\mathtt{LLM}$ and \href{https://huggingface.co/deepseek-ai/DeepSeek-R1-Distill-Qwen-14B}{R1-Distill-Qwen-14B} as the reward model backbone. In this setting, \DisORM \emph{performs comparably} to \DisPRM, while \GenPRM \emph{lags behind the other variants}, which is contrary to \citet{prm800k,thinkprm} and our math-domain results in \Cref{fig:math_tts}. Overall, \GenORM outperforms \DisORM/\DisPRM/\GenPRM, without notable degradation in any domain relative to the others (see \Cref{sec:additional_experiments} for detailed per-domain results).
The same pattern holds for a smaller backbone (\href{https://huggingface.co/deepseek-ai/DeepSeek-R1-Distill-Llama-8B}{DeepSeek-R1-Distill-Llama-8B}; \Cref{fig:llama_8B_8B_bon}) and a non-distilled backbone (\href{https://huggingface.co/Qwen/Qwen3-8B}{Qwen3-8B}; \Cref{fig:overall_bon_qwen}).
\Cref{tab:hyperparameter_search} also shows that this trend is not due to insufficient hyperparameter tuning: sweeping the learning rate and LoRA rank yields only marginal changes in PRM performance.

To verify that the above observations generalize across datasets, we take the reward models trained on the MMLU-Pro training split and evaluate them on GPQA-Diamond, MedQA, and LEXam by generating 
$N{=}16$ CoTs for each question. For MedQA and LEXam, we include only \href{https://huggingface.co/HuggingFaceTB/SmolLM3-3B}{SmolLM3-3B} and \href{https://huggingface.co/google/gemma-2-9b-it}{gemma-2-9b-it}, since the other $p_\mathtt{LLM}$ exhibit severe degradation, even compared to random guessing. As shown in \Cref{fig:gpqa_bon}, \GenORM outperforms \DisORM/\DisPRM/\GenPRM on GPQA-Diamond, consistent with the results on MMLU-Pro (\Cref{fig:overall_bon}). In specialized domains (\Cref{fig:medqa_lexam}), generative variants significantly outperform discriminative variants and \GenORM outperforms \GenPRM on CoTs generated by \href{https://huggingface.co/HuggingFaceTB/SmolLM3-3B}{SmolLM3-3B} and performs comparably on CoTs generated by \href{https://huggingface.co/google/gemma-2-9b-it}{gemma-2-9b-it}.

\begin{figure}[t]
\centering
\includegraphics[width=1\textwidth]{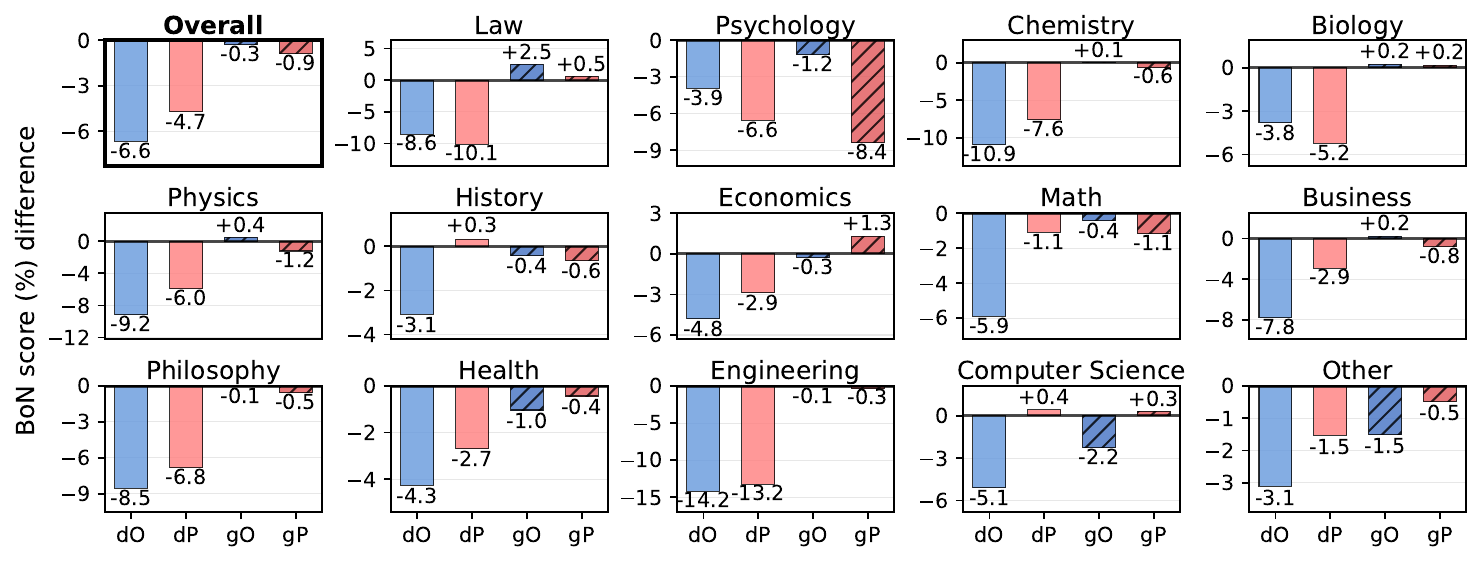}
\caption{\small \textbf{Best-of-$N$ performance gap between all-domain and single-domain training} on MMLU-pro.}
\label{fig:domain_difference}
\end{figure}
\paragraph{Effect of multi-domain training.}
To assess the effect of multi-domain training, we train and evaluate all four reward model variants \emph{only} on each MMLU-Pro domain and compare each variant to its multi-domain counterpart.
\Cref{fig:domain_difference} shows the degradation in Bo$N$ performance with $N{=}16$ under domain-specialized training.
We observe \emph{large drops} for \DisORM and \DisPRM under single-domain training relative to their multi-domain counterparts, with a slightly larger decline for \DisORM, likely because outcome-only supervision is sparser than step-level supervision and both discriminative variants require substantial training data.

In contrast, \GenORM and \GenPRM appear more \emph{sample-efficient}: even without multi-domain training, their performance decreases only modestly (or in some cases improves).
This also explains the results on MedQA and LEXam (\Cref{fig:medqa_lexam}): the generative variants show strong gains over the discriminative variants in these sample-inefficient specialized domains (medical and legal).
We defer complete results of single-domain training for the four reward models to \Cref{fig:domain_specialization_bon,fig:domain_specialization_wmv} in \Cref{sec:additional_experiments}.
\section{Analysis on Why PRMs Fail in the Multi-Domain Settings}\label{sec:analysis}
In this section, we analyze the failure modes of PRMs observed in the multi-domain setting of \Cref{sec:experiments}.

\subsection{Risk of PRMs with CoT Length}\label{sec:cot_length}

\begin{figure}[!t]
    \vspace{-0.3in}
    \centering
    \begin{subfigure}[t]{0.49\textwidth}
        \centering
        \includegraphics[width=\linewidth]{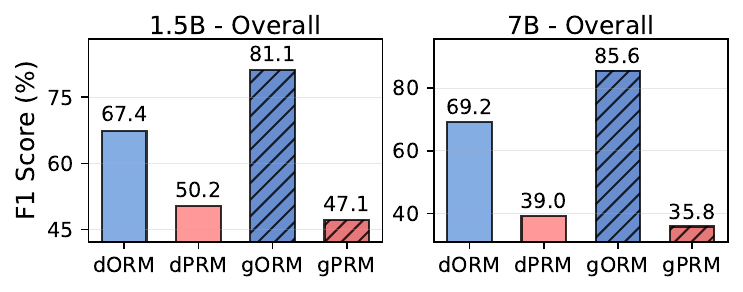}
        \caption{Outcome-verification results on \textbf{``aha'' CoTs}.}
        \label{fig:math_aha_overall}
    \end{subfigure}
    \hfill
    \begin{subfigure}[t]{0.49\textwidth}
        \centering
        \includegraphics[width=\linewidth]{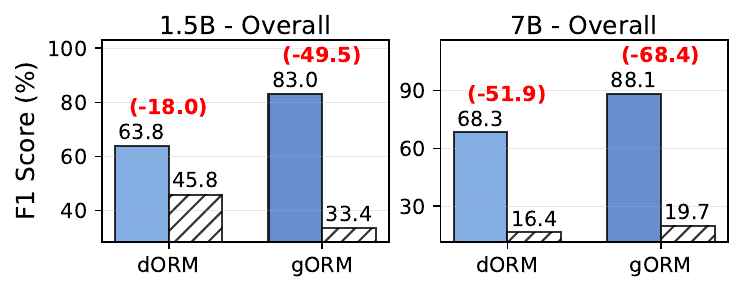}
        \caption{Results on \textbf{randomly shuffled ``aha'' CoTs}.}
        \label{fig:math_overfit}
    \end{subfigure}
    \caption{\textbf{(a)}: ORMs outperform PRMs on ``aha'' CoTs; however, \textbf{(b)}: their performance drops when intermediate steps are randomly shuffled. This suggests that ORMs \textbf{do not simply memorize question-answer pairs}.}
\end{figure}
\paragraph{``Aha'' CoTs.}
As noted in \Cref{sec:reward_models}, PRMs typically assume that once a reasoning step is incorrect, \emph{all subsequent steps are incorrect}. However, recent reasoning models can recover from earlier mistakes and still arrive at the correct answer \citep[an \textbf{``aha'' moment};][]{guo2025deepseek}. In such cases, PRMs can miss the recovery due to a monotonicity bias induced by their training data.
To demonstrate this, we evaluate on ``aha'' CoTs from \href{https://huggingface.co/datasets/Qwen/ProcessBench}{ProcessBench}\footnote{The ``aha'' experiments are conducted only on \href{https://huggingface.co/datasets/Qwen/ProcessBench}{ProcessBench}, which provides human-annotated step-level labels. For MMLU-Pro, LLM-annotated process labels would make the rate of ``aha'' CoTs depend on annotator reliability.} that contain at least one incorrect step ($\exists\,t\in\{1,\ldots,T\}: z_t=0$) but a correct outcome ($y=1$). 
Overall, ``aha'' CoTs account for 15.3\% of the cases.
In \Cref{fig:math_aha_overall}, we report F1 scores (\%) for the ``aha'' CoTs using $M{=}16$ for \GenORM/\GenPRM.
We observe that PRM variants perform particularly \textbf{poorly on ``aha'' CoTs}. Moreover, scaling the backbone from \href{https://huggingface.co/deepseek-ai/DeepSeek-R1-Distill-Qwen-1.5B}{1.5B} to \href{https://huggingface.co/deepseek-ai/DeepSeek-R1-Distill-Qwen-7B}{7B} improves ORM performance, whereas PRM performance degrades with larger backbones, possibly because larger PRMs are more likely to follow the PRM assumption inherent in their training data and objective (\cf\ \Cref{eq:disprm_train,eq:genprm_train}).

\paragraph{Do ORMs overfit on ``aha'' CoTs?}
A natural concern about ORM results on “aha” CoTs in \Cref{fig:math_aha_overall} is \emph{overfitting}: ORMs might only memorize questions and their answers, thereby correctly verifying “aha” CoTs without checking the correctness of intermediate reasoning steps. This memorization issue in the math domain has recently been studied by \citet{wu2025reasoning}. To investigate this, we conduct the following test: (i) \textbf{replace the intermediate reasoning steps} $r_{1:T-1}$ with $r^\prime_{1:T-1}$ \textbf{taken from other CoTs}, and (ii) evaluate ORMs on these perturbed CoTs. If ORMs only memorize the answer in the final reasoning step $r_T$, their performance should remain largely unaffected.
However, \Cref{fig:math_overfit} shows a \textbf{significant drop} for ORMs (dashed), indicating the reliance on intermediate steps. Interestingly, the degradation is greater with the \href{https://huggingface.co/deepseek-ai/DeepSeek-R1-Distill-Qwen-7B}{7B backbone} than with the \href{https://huggingface.co/deepseek-ai/DeepSeek-R1-Distill-Qwen-1.5B}{1.5B backbone} for both \DisORM and \GenORM. This suggests that larger models rely \emph{more heavily} on intermediate reasoning steps during verification.

\begin{figure}[t]
    \vspace{-0.3in}
    \centering
    \begin{subfigure}[t]{0.58\textwidth}
        \centering
        \includegraphics[width=\linewidth]{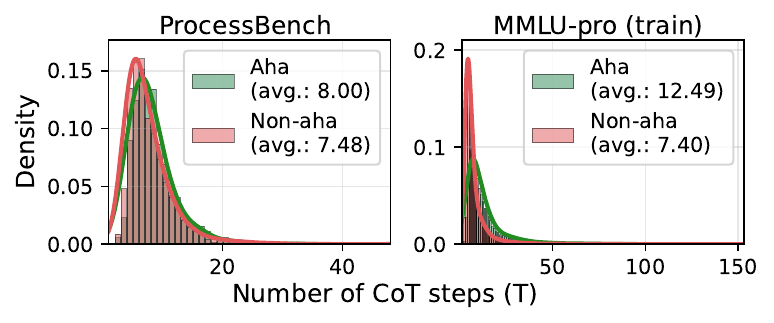}
        \vspace{-0.2in}
        \caption{\textbf{Length distribution of ``aha'' CoTs}.}
        \label{fig:length_of_aha_cots}
    \end{subfigure}
    \hfill
    \begin{subfigure}[t]{0.41\textwidth}
        \centering
        \includegraphics[width=\linewidth]{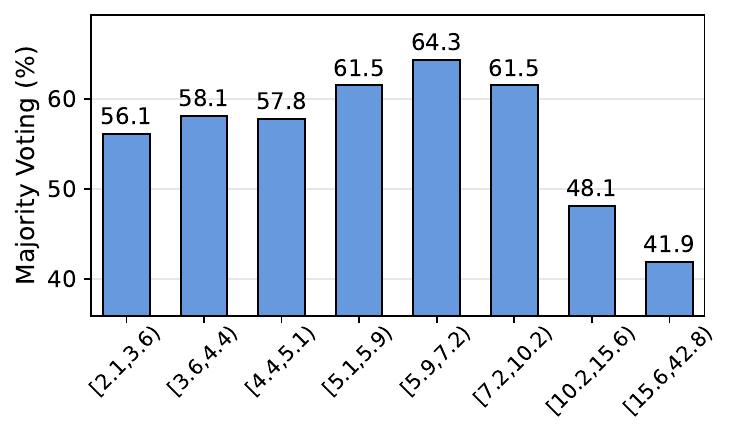}
        \vspace{-0.2in}
        \caption{\textbf{MV} vs. \textbf{CoT length} on MMLU-Pro.}
        \label{fig:mj_by_length}
    \end{subfigure}
    \caption{\textbf{(a)}: ``Aha'' moments lengthen CoTs, an effect pronounced in the multi-domain setting (MMLU-Pro); and \textbf{(b)}: majority voting results degrade significantly with increasing CoT length.}
\end{figure}

\paragraph{Risk increases with CoT length.}
``Aha'' moments can also lengthen CoTs, an effect especially pronounced in the multi-domain setting (\Cref{fig:length_of_aha_cots}), where LLMs struggle more than in the math domain. As shown in \Cref{fig:mj_by_length}, majority voting results degrade significantly with increasing CoT length in the multi-domain setting.
Consistent with the outcome-verification failures of PRMs on ``aha'' CoTs, we argue that \emph{the error of PRM variants grows with CoT length} ($T$). Intuitively, as a CoT grows longer, the chance that a PRM misclassifies at least one intermediate step rises, making it more likely to \emph{prematurely} conclude the CoT is incorrect. Longer CoTs also create more opportunities for ``aha'' recoveries that PRMs systematically miss. We formalize this as follows:

\begin{thm}[Informal: Log-error bound of \DisORM and \GenORM; \Cref{thm:orm}]
\label{thm:main_paper_orm}
Under mild assumptions on the variance of the reward model's error, the expected squared log-error of \DisORM and \GenORM is bounded by a constant that is \textbf{independent} of the CoT length $T$.
\end{thm}

\begin{thm}[Informal: Log-error lower bound of \DisPRM; \Cref{thm:dprm}]
\label{thm:main_paper_dprm}
Assuming a minimum average per-step error variance and bounded local error correlation, the expected squared log-error of \DisPRM grows \textbf{at least linearly} with the CoT length $T$.
\end{thm}

\begin{thm}[Informal: Log-error lower bound of sampled \GenPRM; \Cref{thm:gprm-log}]
\label{thm:main_paper_gprm-log}
Under the same assumptions as \Cref{thm:main_paper_dprm}, the expected squared log-error of \GenPRM also grows \textbf{at least linearly} with $T$. Furthermore, the generative sampling process introduces additional per-step variance, which strictly worsens this lower bound compared to \DisPRM.
\end{thm}

All formal definitions, assumptions, and proofs are deferred to \Cref{sec:theoretical_analysis}. 
\Cref{thm:main_paper_orm,thm:main_paper_dprm,thm:main_paper_gprm-log} establish that while \DisORM/\GenORM error bounds remain stable regardless of length, \DisPRM/\GenPRM lower bounds grow linearly with $T$. In \Cref{thm:meanprm}, we additionally show that for \GenPRM with Monte Carlo estimation (\cf\ \Cref{eq:genprm_test}), the log-error lower bound also increases linearly with $T$.

\begin{wrapfigure}{r}{0.55\textwidth}
\vspace{-0.2in}
\includegraphics[width=\linewidth, trim=0 0 0 0, clip]{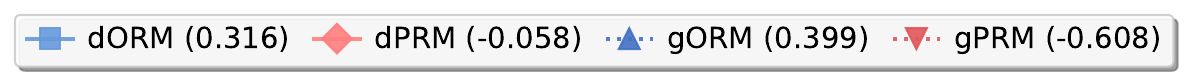}
\vspace{0.02em}
\includegraphics[width=\linewidth, trim=0 0 0 0, clip]{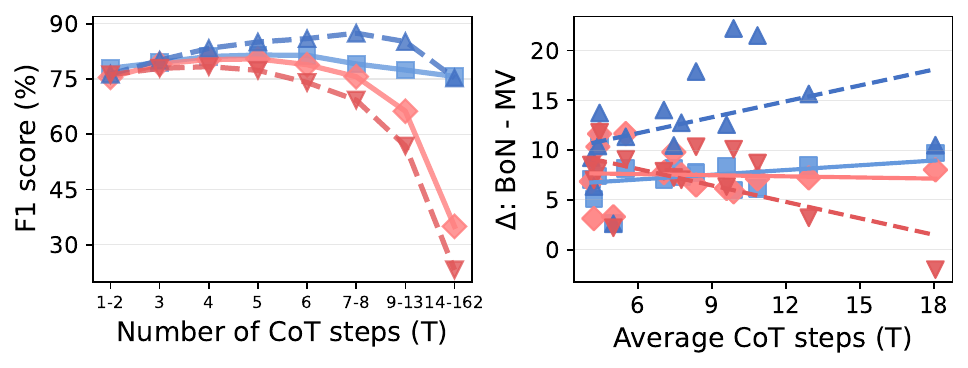}
\vspace{-0.32in}
\caption{\small \textbf{(Left)}: Outcome verification vs.\ CoT length; \textbf{(Right)}: TTS improvement vs.\ average CoT length.
}
\label{fig:length_analysis}
\end{wrapfigure}

\paragraph{Empirical support.}
To empirically support \Cref{thm:main_paper_orm,thm:main_paper_dprm,thm:main_paper_gprm-log}, we plot the F1 score (\%) for outcome-verification in the multi-domain setting as a function of the number of CoT steps ($T$) in \Cref{fig:length_analysis}-(Left). We divide CoTs into eight bins: 1-2, 3, 4, 5, 6, 7-8, 9-13, and 14-162 steps. 
As $T$ increases, \DisPRM/\GenPRM degrade considerably relative to \DisORM/\GenORM. \Cref{fig:length_analysis}-(Right) shows the performance improvements over majority voting for different categories with respect to the average number of CoT steps. We observe negative correlations for \DisPRM (-0.058) and \GenPRM (-0.608), while \DisORM (0.316) and \GenORM (0.399) show positive correlations. These results not only provide empirical support for \Cref{thm:main_paper_orm,thm:main_paper_dprm,thm:main_paper_gprm-log} but also demonstrate that increasing CoT length can degrade TTS performance for \DisPRM and \GenPRM in the multi-domain setting.

\subsection{Label Noise and Consensus Filtering of PRMs}\label{sec:label_noise}

\begin{figure}[!t]
\vspace{-0.3in}
\centering
\hspace{0.39in}
\includegraphics[width=0.44\textwidth]{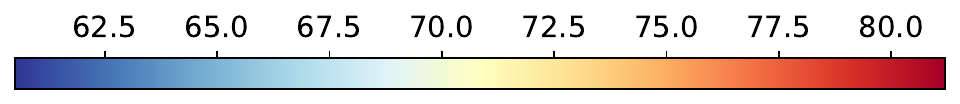}
\hspace{0.09in}
\includegraphics[width=0.44\textwidth]{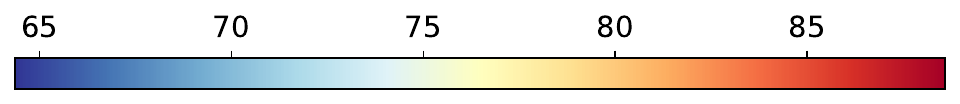}
\medskip
\vspace{-0.16in}
\centering
\includegraphics[width=0.95\textwidth]{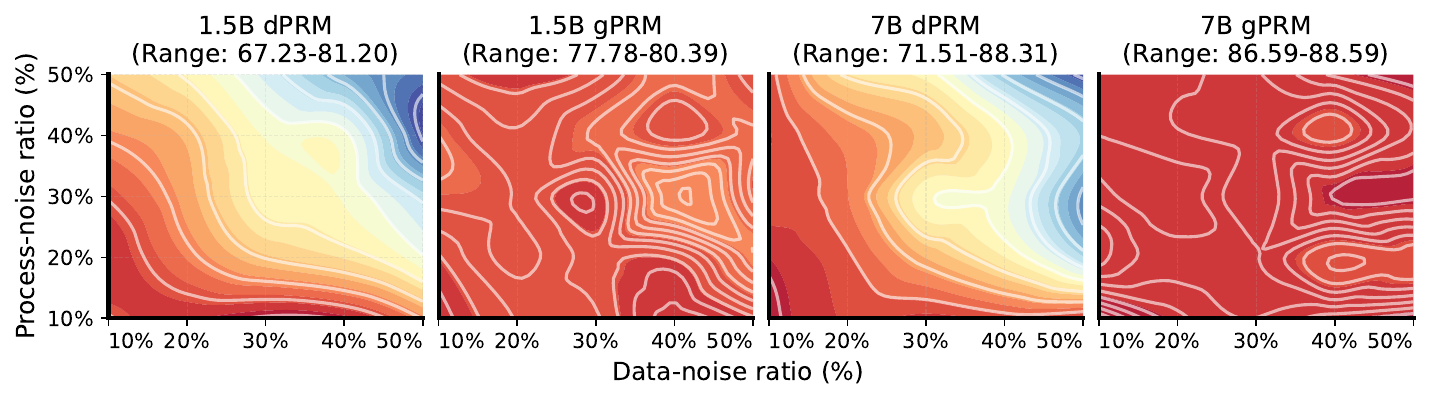}
\vspace{-0.15in}
\caption{\small \textbf{Outcome-verification results of PRMs vs.\ label noise} on GSM8K.}
\label{fig:full_math_noise_analysis}
\end{figure}
\paragraph{Label noise risk.}
Beyond CoT-length effects, \emph{label noise} poses an additional risk, especially in multi-domain settings. 
Since human annotation of long CoTs is more costly in specialized domains such as law and medicine than in math, prior work often relies on LLMs to automatically label process steps~\citep{versaprm}, which introduces noise that can degrade PRM performance. 
Indeed, recent evaluations in the math domain document that automated annotation pipelines, such as Monte Carlo estimation, can inject substantial label noise and evaluation artifacts~\citep{zhang2025lessons}. We study this by injecting synthetic noise into the process labels of PRM800K. We vary the level of noise along two axes: (i) \emph{process-noise ratio} (the per-step probability of flipping a process label) and (ii) \emph{data-noise ratio} (the fraction of examples to which noise is applied). 
We report the outcome-verification F1 score (\%) in \Cref{fig:full_math_noise_analysis} for \href{https://huggingface.co/deepseek-ai/DeepSeek-R1-Distill-Qwen-1.5B}{1.5B} and \href{https://huggingface.co/deepseek-ai/DeepSeek-R1-Distill-Qwen-7B}{7B} backbones, using greedy decoding for generative variants.
\DisPRM is \textbf{highly sensitive} to label noise, demonstrating its potential vulnerability in multi-domain settings. In contrast, \GenPRM is more robust, which is consistent with reports that LLM memorization can make random label noise act as a regularizer in math~\citep{wu2025reasoning}. 

\begin{figure}[!t]
\centering
\includegraphics[width=0.95\textwidth]{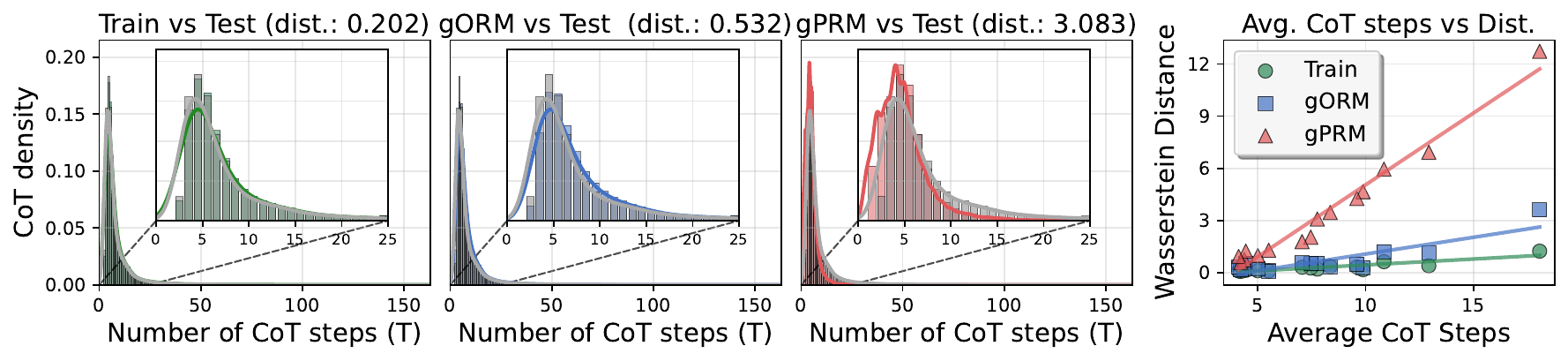}
\vspace{-0.1in}
\caption{\small \textbf{Length distribution shift} on MMLU-Pro (overall/per-domain) measured by Wasserstein distance.
}
\label{fig:training_distribution}
\end{figure}

\paragraph{Length shift hurts \GenPRM.}
We further analyze why \GenPRM degrades in the multi-domain setting, despite its robustness to label noise in math. As CoTs become longer, the imperfect $p_\texttt{LLM-j}$ struggles to align stepwise verification rationales with process labels. As a result, \emph{consensus filtering} \textbf{prunes long CoTs}, shifting the training CoT-length distribution away from the test set (\Cref{fig:training_distribution}).
\color{black}

We quantify the above length distribution shift with the Wasserstein distance~\citep{kantorovich1960mathematical}, reporting distances from the test set to the unfiltered pool (\textbf{Train}), the \GenORM training set, and the \GenPRM  training set.
In the math domain (\Cref{tab:wasserstein_distance_math}), \GenPRM has the smallest distance (\eg, overall: 2.760/2.430/1.600 for Train/\GenORM/\GenPRM), whereas in the multi-domain setting (\Cref{fig:training_distribution,tab:wasserstein_distance}) it has \textbf{the largest} distance (\eg, overall: 0.202/0.532/3.083 for Train/\GenORM/\GenPRM).

\begin{figure}[!t]
\centering
\includegraphics[width=0.95\textwidth]{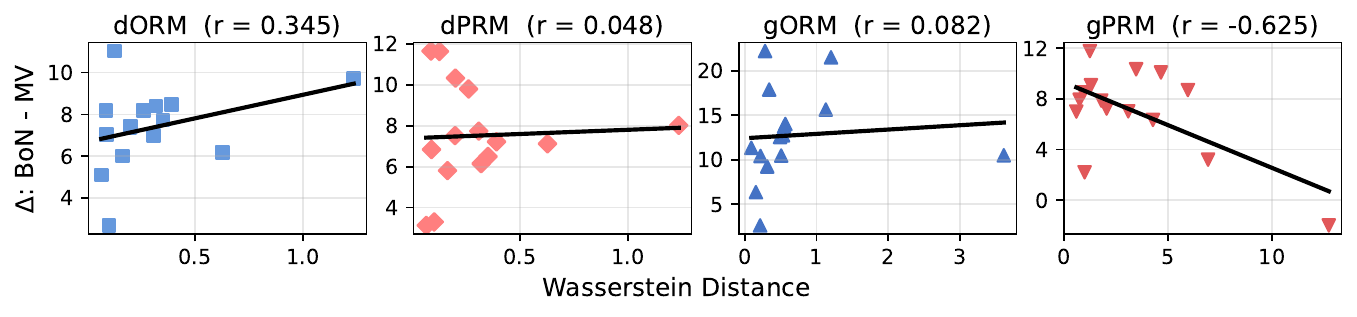}
\vspace{-0.11in}
\caption{\small \textbf{Per-domain Bo$N$ improvement over majority voting} vs. \textbf{Wasserstein distance} on MMLU-Pro.}
\label{fig:distance_vs_improvement}
\end{figure}

The distribution shift of \GenPRM also corresponds to its degradation across domains, observed in the multi-domain setting (\Cref{fig:overall_bon}). \Cref{fig:distance_vs_improvement} shows a strong negative correlation between the Wasserstein distance and per-domain improvement over majority voting with $N{=}16$ for \GenPRM (\(-0.625\)), whereas correlations are weak for the other methods (\(0.345/0.048/0.082\) for \DisORM/\DisPRM/\GenORM). Together, these results suggest that \emph{consensus filtering} induces a length-distribution shift that disproportionately affects \GenPRM in the multi-domain setting, despite its robustness to label noise.

\vspace{-0.1in}

\paragraph{Label refinement and relaxed filtering.}
To mitigate the CoT-length distribution shift for \GenPRM, we further test label refinement with \href{https://blog.google/technology/google-deepmind/google-gemini-ai-update-december-2024/}{Gemini-2.0 Flash}~\citep{comanici2025gemini} and relaxed \textit{consensus filtering}. Although both increase the CoT survival rate, they neither substantially reduce the Wasserstein distance nor improve downstream performance. See \Cref{tab:wasserstein_distance,tab:survival_proportion,tab:label_refinement_and_relaxed_filtering} in \Cref{sec:additional_analysis} for more details.

\section{Practical Guidelines, Limitations, and Future Work}\label{sec:conclusion}
This section suggests practical guidance, clarifies limitations, and outlines future directions. Building on the empirical and theoretical results of \Cref{sec:experiments,sec:analysis}, we summarize when each reward-model variant is preferable in the table below, and then connect the entries to the supporting evidence.

\begin{table}[H]
\centering
\begin{tabular}{clr}
(i) & \emph{Short} CoTs, \emph{clean} labels, \emph{tight} latency & \DisPRM \\
\hdashline[1pt/1pt]
(ii) & \emph{Long} CoTs / \emph{frequent error recoveries} & \GenORM{} if compute permits; else \DisORM \\
\hdashline[1pt/1pt]
(iii) & \emph{Mixed/shifting domains} & \GenORM \\
\hdashline[1pt/1pt]
(iv) & \emph{High label noise} & \makecell[r]{ORM\\{\small \color{gray} PRMs amplify early errors}} \\
\hdashline[1pt/1pt]
(v) & {\emph{Strict compute/latency}} & \makecell[r]{\DisORM/\DisPRM\\{\small \color{gray} \GenORM and \GenPRM add sampling overhead}}\\
\hdashline[1pt/1pt]
(vi) & \emph{Limited training data} & \makecell[r]{\GenORM/\GenPRM\\{\small \color{gray} Higher sample efficiency}} \\
\end{tabular}
\end{table}

\paragraph{From findings to recommendations.}

Rows (i) and (ii) follow from our CoT-length analysis in \Cref{sec:cot_length}. When CoTs are short and process labels are clean, the per-step signal exploited by \DisPRM is informative and its discriminative head keeps inference cheap, matching the math-domain regime where \DisPRM is competitive (\Cref{fig:math_tts}). As $T$ grows, however, \Cref{thm:main_paper_dprm,thm:main_paper_gprm-log} show that the log-error lower bounds of \DisPRM and \GenPRM grow linearly with $T$, while \Cref{thm:main_paper_orm} bounds the ORM error independently of $T$. This is corroborated in \Cref{fig:length_analysis} (length-vs-improvement correlations of $-0.058$/$-0.608$ for \DisPRM/\GenPRM versus $0.316$/$0.399$ for \DisORM/\GenORM) and by the ``aha'' analysis in \Cref{fig:math_aha_overall}, where PRMs systematically miss recoveries from earlier mistakes, hence the ORM recommendation in row (ii).

Rows (iii)-(vi) reflect the multi-domain and resource trade-offs surfaced in \Cref{sec:label_noise,sec:experiments}. For mixed or shifting domains, \emph{consensus filtering} prunes long CoTs and induces a length-distribution shift that disproportionately hurts \GenPRM (largest Wasserstein distance in \Cref{tab:wasserstein_distance}, with a correlation of $-0.625$ between distance and per-domain improvement in \Cref{fig:distance_vs_improvement}), motivating \GenORM as the default. Under high label noise, the PRM800K noise-injection study (\Cref{fig:full_math_noise_analysis}) shows that \DisPRM is highly sensitive to per-step label flips and that even the more robust \GenPRM inherits the monotonicity assumption that amplifies early errors, so ORMs are the safer choice.

Finally, \GenORM/\GenPRM are slower than discriminative variants because they autoregressively generate verification tokens with $M$ rollouts (\cf\ \Cref{eq:genprm_test}), while \DisORM/\DisPRM score each CoT in a single forward pass. On MMLU-Pro with CoTs from Llama-3.1-8B-Instruct, the normalized inference times are 0.0005/0.0005/0.0571/0.0306 seconds per CoT for \DisORM/\DisPRM/\GenORM/\GenPRM, respectively, making \DisORM/\DisPRM preferable under strict compute or latency budgets. Although \GenPRM is $1.87\times$ faster than \GenORM due to early stopping, it may miss ``aha'' CoTs; thus, we recommend \GenORM when reliability is prioritized, especially in medical and legal domains. In contrast, generative variants better exploit LLM priors and rationale-style supervision, improving sample efficiency with limited training data (\Cref{fig:domain_difference} and \Cref{sec:additional_experiments}), so we recommend them in row (vi).

\paragraph{Limitations and future work.}
While we present a thorough analysis of four reward model variants, our study has several limitations:
(i) All models are trained via \textbf{supervised fine–tuning}. One could instead use a generative verifier to roll out rationales and treat agreement between their verdict and the GT label as a reward signal for reinforcement learning (RL). Because using RL to train verifiers/reward models is uncommon and introduces additional confounders, we exclude RL-based training from our analysis.
(ii) Owing to computational constraints, we adopt \textbf{LoRA adapters} rather than full-parameter fine-tuning. This choice may affect performance and scaling behavior. However, we expect the qualitative trends to hold.
(iii) Following most of the PRM literature~\citep{prm800k, versaprm}, we do not consider \textbf{tool use}, however, \citet{verification-tool} showed that tool use can help reduce auto-label noise. In future work, we plan to extend our analysis to broader task domains, model families, and training regimes. We also plan to explicitly study tool-augmented verification and inference pipelines.

\section*{Acknowledgement}
This work was supported by the Institute for Information \& Communications Technology Planning \& Evaluation (IITP) grants funded by the Korea government (MSIT) (No. RS-2019-II190075, Artificial Intelligence Graduate School Program (KAIST); No. RS-2022-II220713, Meta-learning Applicable to Real-world Problems), and by Center for Applied Research in Artificial Intelligence (CARAI) grant funded by DAPA and ADD (UD230017TD).

\section*{Broader Impact Statement}
This work evaluates verification strategies for test-time scaling of LLMs across multiple domains. It \textbf{does not} involve human subjects, user studies, or the collection of personally identifiable information. All datasets used are \textbf{publicly available} benchmarks and were accessed under their respective licenses. To the best of our knowledge, they do not contain sensitive personal data.

A natural direction for future work is to increase the trustworthiness of LLM outputs in real systems by verifying them, thereby reducing reasoning errors and hallucinations. Although our experiments include legal and medical-themed datasets (\eg, law and health), the models and methods are research artifacts and are \textbf{not} intended for real-world legal, medical, or other high-stakes decision-making. They should not substitute professional judgment, and any deployment in such settings would require additional domain-specific validation, safety auditing, and regulatory compliance.

\section*{Reproducibility Statement}
We believe that we provide sufficient materials, including prompts, hyperparameters, model backbones, training details, and the synthetic data generation process, throughout the main paper (\Cref{sec:related_work,sec:experimental_setups}). Additional details can be found in \Cref{sec:prompts,sec:dataset,sec:implementation_details,sec:training_examples}. Furthermore, we \textbf{publicly release} all relevant artifacts: \textbf{(i) code}, \textbf{(ii) datasets} (including any we generate), and \textbf{(iii) model checkpoints}. 
\paragraph{\href{https://github.com/db-Lee/Multi-RM}{Code}.}
\begin{itemize}[itemsep=1mm,parsep=1pt,topsep=2pt,leftmargin=*]
  \item \texttt{discriminative/}: training/inference code for \emph{discriminative} variants (\DisORM/\DisPRM), adapted from \href{https://github.com/UW-Madison-Lee-Lab/VersaPRM}{VersaPRM}~\citep{versaprm}.
  \item \texttt{generative/}: training/inference code for \emph{generative} variants (\GenORM/\GenPRM).
\end{itemize}

\paragraph{Training datasets.}
\begin{itemize}[itemsep=1mm,parsep=1pt,topsep=2pt,leftmargin=*]
  \item \href{https://huggingface.co/datasets/dongboklee/MMLU-Pro_Llama-3.1-8B-Instruct_train}{\texttt{MMLU-Pro\_Llama-3.1-8B-Instruct\_train}}: MMLU-Pro training dataset for \DisORM/\DisPRM, adapted from \href{https://github.com/UW-Madison-Lee-Lab/VersaPRM}{VersaPRM}~\citep{versaprm}.
  \item \href{https://huggingface.co/datasets/dongboklee/MMLU-Pro_Llama-3.1-8B-Instruct_gORM_train}{\texttt{MMLU-Pro\_Llama-3.1-8B-Instruct\_gORM\_train}}: MMLU-Pro training dataset for \GenORM.
  \item \href{https://huggingface.co/datasets/dongboklee/MMLU-Pro_Llama-3.1-8B-Instruct_gPRM_train}{\texttt{MMLU-Pro\_Llama-3.1-8B-Instruct\_gPRM\_train}}: MMLU-Pro training dataset for \GenPRM.
\end{itemize}

\paragraph{Test datasets.}
\begin{itemize}[itemsep=1mm,parsep=1pt,topsep=2pt,leftmargin=*]
  \item \href{https://huggingface.co/datasets/dongboklee/MMLU-Pro_Llama-3.1-8B-Instruct_test}{\texttt{MMLU-Pro\_Llama-3.1-8B-Instruct\_test}}: MMLU-Pro test dataset generated by \href{https://huggingface.co/meta-llama/Llama-3.1-8B-Instruct}{Llama-3.1-8B-Instruct}, adapted from \href{https://github.com/UW-Madison-Lee-Lab/VersaPRM}{VersaPRM}~\citep{versaprm}.
  \item \href{https://huggingface.co/datasets/dongboklee/MMLU-Pro_SmolLM3-3B_test}{\texttt{MMLU-Pro\_SmolLM3-3B\_test}}: MMLU-Pro test dataset generated by \href{https://huggingface.co/HuggingFaceTB/SmolLM3-3B}{SmolLM3-3B}.
  \item \href{https://huggingface.co/datasets/dongboklee/MMLU-Pro_Qwen2.5-7B-Instruct_test}{\texttt{MMLU-Pro\_Qwen2.5-7B-Instruct\_test}}: MMLU-Pro test dataset generated by \href{https://huggingface.co/Qwen/Qwen2.5-7B-Instruct}{Qwen2.5-7B-Instruct}.
  \item \href{https://huggingface.co/datasets/dongboklee/MMLU-Pro_gemma-2-9b-it_test}{\texttt{MMLU-Pro\_gemma-2-9b-it\_test}}: MMLU-Pro test dataset generated by \href{https://huggingface.co/google/gemma-2-9b-it}{gemma-2-9b-it}.
  \item \href{https://huggingface.co/datasets/dongboklee/MMLU-Pro_Llama-3.1-70B-Instruct_test}{\texttt{MMLU-Pro\_Llama-3.1-70B-Instruct\_test}}: MMLU-Pro test dataset generated by \href{https://huggingface.co/meta-llama/Llama-3.1-70B-Instruct}{Llama-3.1-70B-Instruct}.
  \item \href{https://huggingface.co/datasets/dongboklee/GPQA-diamond_Llama-3.1-8B-Instruct_test}{\texttt{GPQA-diamond\_Llama-3.1-8B-Instruct\_test}}: GPQA-diamond test dataset generated by \href{https://huggingface.co/meta-llama/Llama-3.1-8B-Instruct}{Llama-3.1-8B-Instruct}.
  \item \href{https://huggingface.co/datasets/dongboklee/GPQA-diamond_SmolLM3-3B_test}{\texttt{GPQA-diamond\_SmolLM3-3B\_test}}: GPQA-diamond test dataset generated by \href{https://huggingface.co/HuggingFaceTB/SmolLM3-3B}{SmolLM3-3B}.
  \item \href{https://huggingface.co/datasets/dongboklee/GPQA-diamond_Qwen2.5-7B-Instruct_test}{\texttt{GPQA-diamond\_Qwen2.5-7B-Instruct\_test}}: GPQA-diamond test dataset generated by \href{https://huggingface.co/Qwen/Qwen2.5-7B-Instruct}{Qwen2.5-7B-Instruct}.
  \item \href{https://huggingface.co/datasets/dongboklee/GPQA-diamond_gemma-2-9b-it_test}{\texttt{GPQA-diamond\_gemma-2-9b-it\_test}}: GPQA-diamond test dataset generated by \href{https://huggingface.co/google/gemma-2-9b-it}{gemma-2-9b-it}.
  \item \href{https://huggingface.co/datasets/dongboklee/GPQA-diamond_Llama-3.1-70B-Instruct_test}{\texttt{GPQA-diamond\_Llama-3.1-70B-Instruct\_test}}: GPQA-diamond test dataset generated by \href{https://huggingface.co/meta-llama/Llama-3.1-70B-Instruct}{Llama-3.1-70B-Instruct}.
  \item \href{https://huggingface.co/datasets/dongboklee/MedQA_SmolLM3-3B_test}{\texttt{MedQA\_SmolLM3-3B\_test}}: MedQA test dataset generated by \href{https://huggingface.co/HuggingFaceTB/SmolLM3-3B}{SmolLM3-3B}.
  \item \href{https://huggingface.co/datasets/dongboklee/MedQA_gemma-2-9b-it_test}{\texttt{MedQA\_gemma-2-9b-it\_test}}: MedQA test dataset generated by \href{https://huggingface.co/google/gemma-2-9b-it}{gemma-2-9b-it}.
  \item \href{https://huggingface.co/datasets/dongboklee/LEXam_SmolLM3-3B_test}{\texttt{LEXam\_SmolLM3-3B\_test}}: LEXam test dataset generated by \href{https://huggingface.co/HuggingFaceTB/SmolLM3-3B}{SmolLM3-3B}.
  \item \href{https://huggingface.co/datasets/dongboklee/LEXam_gemma-2-9b-it_test}{\texttt{LEXam\_gemma-2-9b-it\_test}}: LEXam test dataset generated by \href{https://huggingface.co/google/gemma-2-9b-it}{gemma-2-9b-it}.
\end{itemize}

\paragraph{Model checkpoints.}
\begin{itemize}[itemsep=1mm,parsep=1pt,topsep=2pt,leftmargin=*]
  \item
\href{https://huggingface.co/dongboklee/dORM-14B}{\texttt{dORM-14B}} /
\href{https://huggingface.co/dongboklee/dORM-14B-2}{\texttt{-2}} /
\href{https://huggingface.co/dongboklee/dORM-14B-3}{\texttt{-3}} /
\href{https://huggingface.co/dongboklee/dORM-14B-4}{\texttt{-4}} /
\href{https://huggingface.co/dongboklee/dORM-14B-5}{\texttt{-5}}:
\DisORM with \href{https://huggingface.co/deepseek-ai/DeepSeek-R1-Distill-Qwen-14B}{DeepSeek-R1-Distill-Qwen-14B} backbone
(seed 1/2/3/4/5).

\item
\href{https://huggingface.co/dongboklee/dPRM-14B}{\texttt{dPRM-14B}} /
\href{https://huggingface.co/dongboklee/dPRM-14B-2}{\texttt{-2}} /
\href{https://huggingface.co/dongboklee/dPRM-14B-3}{\texttt{-3}} /
\href{https://huggingface.co/dongboklee/dPRM-14B-4}{\texttt{-4}} /
\href{https://huggingface.co/dongboklee/dPRM-14B-5}{\texttt{-5}}:
\DisPRM with \href{https://huggingface.co/deepseek-ai/DeepSeek-R1-Distill-Qwen-14B}{DeepSeek-R1-Distill-Qwen-14B} backbone
(seed 1/2/3/4/5).

\item
\href{https://huggingface.co/dongboklee/gORM-14B}{\texttt{gORM-14B}} /
\href{https://huggingface.co/dongboklee/gORM-14B-2}{\texttt{-2}} /
\href{https://huggingface.co/dongboklee/gORM-14B-3}{\texttt{-3}} /
\href{https://huggingface.co/dongboklee/gORM-14B-4}{\texttt{-4}} /
\href{https://huggingface.co/dongboklee/gORM-14B-5}{\texttt{-5}}:
\GenORM with \href{https://huggingface.co/deepseek-ai/DeepSeek-R1-Distill-Qwen-14B}{DeepSeek-R1-Distill-Qwen-14B} backbone
(seed 1/2/3/4/5).

\item
\href{https://huggingface.co/dongboklee/gPRM-14B}{\texttt{gPRM-14B}} /
\href{https://huggingface.co/dongboklee/gPRM-14B-2}{\texttt{-2}} /
\href{https://huggingface.co/dongboklee/gPRM-14B-3}{\texttt{-3}} /
\href{https://huggingface.co/dongboklee/gPRM-14B-4}{\texttt{-4}} /
\href{https://huggingface.co/dongboklee/gPRM-14B-5}{\texttt{-5}}:
\GenPRM with \href{https://huggingface.co/deepseek-ai/DeepSeek-R1-Distill-Qwen-14B}{DeepSeek-R1-Distill-Qwen-14B} backbone
(seed 1/2/3/4/5).

  \item \href{https://huggingface.co/dongboklee/dORM-8B}{\texttt{dORM-8B}}: \DisORM with \href{https://huggingface.co/deepseek-ai/DeepSeek-R1-Distill-Llama-8B}{DeepSeek-R1-Distill-Llama-8B} backbone.
  \item \href{https://huggingface.co/dongboklee/dPRM-8B}{\texttt{dPRM-8B}}: \DisPRM with \href{https://huggingface.co/deepseek-ai/DeepSeek-R1-Distill-Llama-8B}{DeepSeek-R1-Distill-Llama-8B} backbone.
  \item \href{https://huggingface.co/dongboklee/gORM-8B}{\texttt{gORM-8B}}: \GenORM with \href{https://huggingface.co/deepseek-ai/DeepSeek-R1-Distill-Llama-8B}{DeepSeek-R1-Distill-Llama-8B} backbone.
  \item \href{https://huggingface.co/dongboklee/gPRM-8B}{\texttt{gPRM-8B}}: \GenPRM with \href{https://huggingface.co/deepseek-ai/DeepSeek-R1-Distill-Llama-8B}{DeepSeek-R1-Distill-Llama-8B} backbone.

  \item \href{https://huggingface.co/dongboklee/dORM-qwen}{\texttt{dORM-qwen}}: \DisORM with \href{https://huggingface.co/Qwen/Qwen3-8B}{Qwen3-8B} backbone.
  \item \href{https://huggingface.co/dongboklee/dPRM-qwen}{\texttt{dPRM-qwen}}: \DisPRM with \href{https://huggingface.co/Qwen/Qwen3-8B}{Qwen3-8B} backbone.
  \item \href{https://huggingface.co/dongboklee/gORM-qwen}{\texttt{gORM-qwen}}: \GenORM with \href{https://huggingface.co/Qwen/Qwen3-8B}{Qwen3-8B} backbone.
  \item \href{https://huggingface.co/dongboklee/gPRM-qwen}{\texttt{gPRM-qwen}}: \GenPRM with \href{https://huggingface.co/Qwen/Qwen3-8B}{Qwen3-8B} backbone.
\end{itemize}


\bibliography{main}

\begin{thebibliography}{65}
\providecommand{\natexlab}[1]{#1}
\providecommand{\url}[1]{\texttt{#1}}
\expandafter\ifx\csname urlstyle\endcsname\relax
  \providecommand{\doi}[1]{doi: #1}\else
  \providecommand{\doi}{doi: \begingroup \urlstyle{rm}\Url}\fi

\bibitem[Achiam et~al.(2023)Achiam, Adler, Agarwal, Ahmad, Akkaya, Aleman, Almeida, Altenschmidt, Altman, Anadkat, et~al.]{achiam2023gpt}
Josh Achiam, Steven Adler, Sandhini Agarwal, Lama Ahmad, Ilge Akkaya, Florencia~Leoni Aleman, Diogo Almeida, Janko Altenschmidt, Sam Altman, Shyamal Anadkat, et~al.
\newblock {GPT}-4 technical report.
\newblock \emph{arXiv preprint arXiv:2303.08774}, 2023.

\bibitem[Adler et~al.(2024)Adler, Agarwal, Aithal, Anh, Bhattacharya, Brundyn, Casper, Catanzaro, Clay, Cohen, et~al.]{preference3}
Bo~Adler, Niket Agarwal, Ashwath Aithal, Dong~H Anh, Pallab Bhattacharya, Annika Brundyn, Jared Casper, Bryan Catanzaro, Sharon Clay, Jonathan Cohen, et~al.
\newblock Nemotron-4 340b technical report.
\newblock \emph{arXiv preprint arXiv:2406.11704}, 2024.

\bibitem[Albalak et~al.(2024)Albalak, Elazar, Xie, Longpre, Lambert, Wang, Muennighoff, Hou, Pan, Jeong, et~al.]{albalak2024survey}
Alon Albalak, Yanai Elazar, Sang~Michael Xie, Shayne Longpre, Nathan Lambert, Xinyi Wang, Niklas Muennighoff, Bairu Hou, Liangming Pan, Haewon Jeong, et~al.
\newblock A survey on data selection for language models.
\newblock \emph{arXiv preprint arXiv:2402.16827}, 2024.

\bibitem[Bakouch et~al.(2025)Bakouch, Ben~Allal, Lozhkov, Tazi, Tunstall, Patiño, Beeching, Roucher, Reedi, Gallouédec, Rasul, Habib, Fourrier, Kydlicek, Penedo, Larcher, Morlon, Srivastav, Lochner, Nguyen, Raffel, von Werra, and Wolf]{bakouch2025smollm3}
Elie Bakouch, Loubna Ben~Allal, Anton Lozhkov, Nouamane Tazi, Lewis Tunstall, Carlos~Miguel Patiño, Edward Beeching, Aymeric Roucher, Aksel~Joonas Reedi, Quentin Gallouédec, Kashif Rasul, Nathan Habib, Clémentine Fourrier, Hynek Kydlicek, Guilherme Penedo, Hugo Larcher, Mathieu Morlon, Vaibhav Srivastav, Joshua Lochner, Xuan-Son Nguyen, Colin Raffel, Leandro von Werra, and Thomas Wolf.
\newblock {SmolLM3: smol, multilingual, long-context reasoner}.
\newblock \url{https://huggingface.co/blog/smollm3}, 2025.

\bibitem[Bavaresco et~al.(2025)Bavaresco, Bernardi, Bertolazzi, Elliott, Fern{\'a}ndez, Gatt, Ghaleb, Giulianelli, Hanna, Koller, Martins, Mondorf, Neplenbroek, Pezzelle, Plank, Schlangen, Suglia, Surikuchi, Takmaz, and Testoni]{bavaresco2024llms}
Anna Bavaresco, Raffaella Bernardi, Leonardo Bertolazzi, Desmond Elliott, Raquel Fern{\'a}ndez, Albert Gatt, Esam Ghaleb, Mario Giulianelli, Michael Hanna, Alexander Koller, Andre Martins, Philipp Mondorf, Vera Neplenbroek, Sandro Pezzelle, Barbara Plank, David Schlangen, Alessandro Suglia, Aditya~K Surikuchi, Ece Takmaz, and Alberto Testoni.
\newblock {LLM}s instead of human judges? a large scale empirical study across 20 {NLP} evaluation tasks.
\newblock In Wanxiang Che, Joyce Nabende, Ekaterina Shutova, and Mohammad~Taher Pilehvar (eds.), \emph{Proceedings of the 63rd Annual Meeting of the Association for Computational Linguistics (Volume 2: Short Papers)}, pp.\  238--255, Vienna, Austria, July 2025. Association for Computational Linguistics.
\newblock ISBN 979-8-89176-252-7.
\newblock \doi{10.18653/v1/2025.acl-short.20}.
\newblock URL \url{https://aclanthology.org/2025.acl-short.20/}.

\bibitem[Charniak \& Johnson(2005)Charniak and Johnson]{charniak-johnson-2005-coarse}
Eugene Charniak and Mark Johnson.
\newblock Coarse-to-fine n-best parsing and maxent discriminative reranking.
\newblock In \emph{Proceedings of the 43rd Annual Meeting of the Association for Computational Linguistics}, pp.\  173--180. Association for Computational Linguistics, 2005.

\bibitem[Cobbe et~al.(2021)Cobbe, Kosaraju, Bavarian, Chen, Jun, Kaiser, Plappert, Tworek, Hilton, Nakano, et~al.]{cobbe2021training}
Karl Cobbe, Vineet Kosaraju, Mohammad Bavarian, Mark Chen, Heewoo Jun, Lukasz Kaiser, Matthias Plappert, Jerry Tworek, Jacob Hilton, Reiichiro Nakano, et~al.
\newblock Training verifiers to solve math word problems.
\newblock \emph{arXiv preprint arXiv:2110.14168}, 2021.

\bibitem[Comanici et~al.(2025)Comanici, Bieber, Schaekermann, Pasupat, Sachdeva, Dhillon, Blistein, Ram, Zhang, Rosen, et~al.]{comanici2025gemini}
Gheorghe Comanici, Eric Bieber, Mike Schaekermann, Ice Pasupat, Noveen Sachdeva, Inderjit Dhillon, Marcel Blistein, Ori Ram, Dan Zhang, Evan Rosen, et~al.
\newblock Gemini 2.5: Pushing the frontier with advanced reasoning, multimodality, long context, and next generation agentic capabilities.
\newblock \emph{arXiv preprint arXiv:2507.06261}, 2025.

\bibitem[Cui et~al.(2023)Cui, Ning, Li, Chen, Yan, Li, Ling, Tian, and Yuan]{cui2023chatlaw}
Jiaxi Cui, Munan Ning, Zongjian Li, Bohua Chen, Yang Yan, Hao Li, Bin Ling, Yonghong Tian, and Li~Yuan.
\newblock {ChatLaw}: A multi-agent collaborative legal assistant with knowledge graph enhanced mixture-of-experts large language model.
\newblock \emph{arXiv preprint arXiv:2306.16092}, 2023.

\bibitem[Dong et~al.(2023)Dong, Xiong, Goyal, Zhang, Chow, Pan, Diao, Zhang, Shum, and Zhang]{rejection2}
Hanze Dong, Wei Xiong, Deepanshu Goyal, Yihan Zhang, Winnie Chow, Rui Pan, Shizhe Diao, Jipeng Zhang, Kashun Shum, and Tong Zhang.
\newblock {RAFT}: Reward ranked finetuning for generative foundation model alignment.
\newblock \emph{Transactions on Machine Learning Research (TMLR)}, 2023.

\bibitem[Dong et~al.(2024)Dong, Xiong, Pang, Wang, Zhao, Zhou, Jiang, Sahoo, Xiong, and Zhang]{preference1}
Hanze Dong, Wei Xiong, Bo~Pang, Haoxiang Wang, Han Zhao, Yingbo Zhou, Nan Jiang, Doyen Sahoo, Caiming Xiong, and Tong Zhang.
\newblock {RLHF} workflow: From reward modeling to online {RLHF}.
\newblock \emph{Transactions on Machine Learning Research (TMLR)}, 2024.
\newblock ISSN 2835-8856.

\bibitem[Dubey et~al.(2024)Dubey, Jauhri, Pandey, Kadian, Al-Dahle, Letman, Mathur, Schelten, Yang, Fan, et~al.]{llama3}
Abhimanyu Dubey, Abhinav Jauhri, Abhinav Pandey, Abhishek Kadian, Ahmad Al-Dahle, Aiesha Letman, Akhil Mathur, Alan Schelten, Amy Yang, Angela Fan, et~al.
\newblock The llama 3 herd of models.
\newblock \emph{arXiv preprint arXiv:2407.21783}, 2024.

\bibitem[Fan et~al.(2026)Fan, Ni, Merane, Tian, Hermstr{\"u}wer, Huang, Akhtar, Salimbeni, Geering, Dreyer, et~al.]{fan2025lexam}
Yu~Fan, Jingwei Ni, Jakob Merane, Yang Tian, Yoan Hermstr{\"u}wer, Yinya Huang, Mubashara Akhtar, Etienne Salimbeni, Florian Geering, Oliver Dreyer, et~al.
\newblock {LEX}am: Benchmarking legal reasoning on 340 law exams.
\newblock \emph{International Conference on Learning Representations (ICLR)}, 2026.

\bibitem[Fei et~al.(2024)Fei, Shen, Zhu, Zhou, Han, Huang, Zhang, Chen, Yin, Shen, Ge, and Ng]{fei2023lawbench}
Zhiwei Fei, Xiaoyu Shen, Dawei Zhu, Fengzhe Zhou, Zhuo Han, Alan Huang, Songyang Zhang, Kai Chen, Zhixin Yin, Zongwen Shen, Jidong Ge, and Vincent Ng.
\newblock {L}aw{B}ench: Benchmarking legal knowledge of large language models.
\newblock In Yaser Al-Onaizan, Mohit Bansal, and Yun-Nung Chen (eds.), \emph{Proceedings of the 2024 Conference on Empirical Methods in Natural Language Processing}, pp.\  7933--7962, Miami, Florida, USA, November 2024. Association for Computational Linguistics.
\newblock \doi{10.18653/v1/2024.emnlp-main.452}.
\newblock URL \url{https://aclanthology.org/2024.emnlp-main.452/}.

\bibitem[Gao et~al.(2025)Gao, Song, Yang, Cai, Miao, Dong, Li, Ma, Chen, Xu, Tang, Wang, Zan, Quan, Zhang, Sha, Zhang, Ren, Liu, and Chang]{omni-math_dataset}
Bofei Gao, Feifan Song, Zhe Yang, Zefan Cai, Yibo Miao, Qingxiu Dong, Lei Li, Chenghao Ma, Liang Chen, Runxin Xu, Zhengyang Tang, Benyou Wang, Daoguang Zan, Shanghaoran Quan, Ge~Zhang, Lei Sha, Yichang Zhang, Xuancheng Ren, Tianyu Liu, and Baobao Chang.
\newblock Omni-{MATH}: A universal olympiad level mathematic benchmark for large language models.
\newblock \emph{International Conference on Learning Representations (ICLR)}, 2025.

\bibitem[{Gemma Team} et~al.(2024){Gemma Team}, Riviere, Pathak, Sessa, Hardin, Bhupatiraju, Hussenot, Mesnard, Shahriari, Ram{\'e}, et~al.]{team2024gemma}
{Gemma Team}, Morgane Riviere, Shreya Pathak, Pier~Giuseppe Sessa, Cassidy Hardin, Surya Bhupatiraju, L{\'e}onard Hussenot, Thomas Mesnard, Bobak Shahriari, Alexandre Ram{\'e}, et~al.
\newblock Gemma 2: Improving open language models at a practical size.
\newblock \emph{arXiv preprint arXiv:2408.00118}, 2024.

\bibitem[Gou et~al.(2024)Gou, Shao, Gong, yelong shen, Yang, Duan, and Chen]{verification-tool}
Zhibin Gou, Zhihong Shao, Yeyun Gong, yelong shen, Yujiu Yang, Nan Duan, and Weizhu Chen.
\newblock {CRITIC}: Large language models can self-correct with tool-interactive critiquing.
\newblock \emph{International Conference on Learning Representations (ICLR)}, 2024.

\bibitem[Guha et~al.(2023)Guha, Nyarko, Ho, R{\'e}, Chilton, Chohlas-Wood, Peters, Waldon, Rockmore, Zambrano, et~al.]{guha2023legalbench}
Neel Guha, Julian Nyarko, Daniel Ho, Christopher R{\'e}, Adam Chilton, Alex Chohlas-Wood, Austin Peters, Brandon Waldon, Daniel Rockmore, Diego Zambrano, et~al.
\newblock {LegalBench}: A collaboratively built benchmark for measuring legal reasoning in large language models.
\newblock \emph{Advances in neural information processing systems (NeurIPS)}, 2023.

\bibitem[Gulcehre et~al.(2023)Gulcehre, Paine, Srinivasan, Konyushkova, Weerts, Sharma, Siddhant, Ahern, Wang, Gu, et~al.]{rejection1}
Caglar Gulcehre, Tom~Le Paine, Srivatsan Srinivasan, Ksenia Konyushkova, Lotte Weerts, Abhishek Sharma, Aditya Siddhant, Alex Ahern, Miaosen Wang, Chenjie Gu, et~al.
\newblock Reinforced self-training (rest) for language modeling.
\newblock \emph{arXiv preprint arXiv:2308.08998}, 2023.

\bibitem[Guo et~al.(2025)Guo, Yang, Zhang, Song, Zhang, Xu, Zhu, Ma, Wang, Bi, et~al.]{guo2025deepseek}
Daya Guo, Dejian Yang, Haowei Zhang, Junxiao Song, Ruoyu Zhang, Runxin Xu, Qihao Zhu, Shirong Ma, Peiyi Wang, Xiao Bi, et~al.
\newblock {DeepSeek-R1}: Incentivizing reasoning capability in llms via reinforcement learning.
\newblock \emph{arXiv preprint arXiv:2501.12948}, 2025.

\bibitem[He et~al.(2024)He, Luo, Bai, Hu, Thai, Shen, Hu, Han, Huang, Zhang, Liu, Qi, Liu, and Sun]{olympiadbench_dataset}
Chaoqun He, Renjie Luo, Yuzhuo Bai, Shengding Hu, Zhen Thai, Junhao Shen, Jinyi Hu, Xu~Han, Yujie Huang, Yuxiang Zhang, Jie Liu, Lei Qi, Zhiyuan Liu, and Maosong Sun.
\newblock {O}lympiad{B}ench: A challenging benchmark for promoting {AGI} with olympiad-level bilingual multimodal scientific problems.
\newblock In Lun-Wei Ku, Andre Martins, and Vivek Srikumar (eds.), \emph{Proceedings of the 62nd Annual Meeting of the Association for Computational Linguistics (Volume 1: Long Papers)}, pp.\  3828--3850, Bangkok, Thailand, August 2024. Association for Computational Linguistics.
\newblock \doi{10.18653/v1/2024.acl-long.211}.
\newblock URL \url{https://aclanthology.org/2024.acl-long.211/}.

\bibitem[Hendrycks et~al.(2021)Hendrycks, Burns, Kadavath, Arora, Basart, Tang, Song, and Steinhardt]{math_dataset}
Dan Hendrycks, Collin Burns, Saurav Kadavath, Akul Arora, Steven Basart, Eric Tang, Dawn Song, and Jacob Steinhardt.
\newblock Measuring mathematical problem solving with the {MATH} dataset.
\newblock \emph{Neural Information Processing Systems Datasets and Benchmarks Track (Round 2)}, 2021.

\bibitem[Hu et~al.(2022)Hu, Shen, Wallis, Allen-Zhu, Li, Wang, Wang, Chen, et~al.]{hu2022lora}
Edward~J Hu, Yelong Shen, Phillip Wallis, Zeyuan Allen-Zhu, Yuanzhi Li, Shean Wang, Lu~Wang, Weizhu Chen, et~al.
\newblock Lora: Low-rank adaptation of large language models.
\newblock \emph{International Conference on Learning Representations (ICLR)}, 2022.

\bibitem[Jin et~al.(2021)Jin, Pan, Oufattole, Weng, Fang, and Szolovits]{jin2021disease}
Di~Jin, Eileen Pan, Nassim Oufattole, Wei-Hung Weng, Hanyi Fang, and Peter Szolovits.
\newblock What disease does this patient have? a large-scale open domain question answering dataset from medical exams.
\newblock \emph{Applied Sciences}, 11\penalty0 (14):\penalty0 6421, 2021.

\bibitem[Kantorovich(1960)]{kantorovich1960mathematical}
Leonid~V Kantorovich.
\newblock Mathematical methods of organizing and planning production.
\newblock \emph{Management science}, 6\penalty0 (4):\penalty0 366--422, 1960.

\bibitem[Khalifa et~al.(2023)Khalifa, Logeswaran, Lee, Lee, and Wang]{khalifa2023grace}
Muhammad Khalifa, Lajanugen Logeswaran, Moontae Lee, Honglak Lee, and Lu~Wang.
\newblock Grace: Discriminator-guided chain-of-thought reasoning.
\newblock \emph{arXiv preprint arXiv:2305.14934}, 2023.

\bibitem[Khalifa et~al.(2025)Khalifa, Agarwal, Logeswaran, Kim, Peng, Lee, Lee, and Wang]{thinkprm}
Muhammad Khalifa, Rishabh Agarwal, Lajanugen Logeswaran, Jaekyeom Kim, Hao Peng, Moontae Lee, Honglak Lee, and Lu~Wang.
\newblock Process reward models that think.
\newblock \emph{arXiv preprint arXiv:2504.16828}, 2025.

\bibitem[Kojima et~al.(2022)Kojima, Gu, Reid, Matsuo, and Iwasawa]{kojima2022large}
Takeshi Kojima, Shixiang~Shane Gu, Machel Reid, Yutaka Matsuo, and Yusuke Iwasawa.
\newblock Large language models are zero-shot reasoners.
\newblock \emph{Advances in neural information processing systems (NeurIPS)}, 2022.

\bibitem[Kung et~al.(2023)Kung, Cheatham, Medenilla, Sillos, De~Leon, Elepa{\~n}o, Madriaga, Aggabao, Diaz-Candido, Maningo, et~al.]{kung2023performance}
Tiffany~H Kung, Morgan Cheatham, Arielle Medenilla, Czarina Sillos, Lorie De~Leon, Camille Elepa{\~n}o, Maria Madriaga, Rimel Aggabao, Giezel Diaz-Candido, James Maningo, et~al.
\newblock Performance of chatgpt on usmle: potential for ai-assisted medical education using large language models.
\newblock \emph{PLoS digital health}, 2\penalty0 (2):\penalty0 e0000198, 2023.

\bibitem[Kwon et~al.(2023)Kwon, Li, Zhuang, Sheng, Zheng, Yu, Gonzalez, Zhang, and Stoica]{kwon2023efficient}
Woosuk Kwon, Zhuohan Li, Siyuan Zhuang, Ying Sheng, Lianmin Zheng, Cody~Hao Yu, Joseph~E. Gonzalez, Hao Zhang, and Ion Stoica.
\newblock Efficient memory management for large language model serving with pagedattention.
\newblock In \emph{Proceedings of the ACM SIGOPS 29th Symposium on Operating Systems Principles}, 2023.

\bibitem[Lightman et~al.(2024)Lightman, Kosaraju, Burda, Edwards, Baker, Lee, Leike, Schulman, Sutskever, and Cobbe]{prm800k}
Hunter Lightman, Vineet Kosaraju, Yura Burda, Harri Edwards, Bowen Baker, Teddy Lee, Jan Leike, John Schulman, Ilya Sutskever, and Karl Cobbe.
\newblock Let's verify step by step.
\newblock \emph{International Conference Learning Representations (ICLR)}, 2024.

\bibitem[Liu et~al.(2023)Liu, Iter, Xu, Wang, Xu, and Zhu]{liu2023g}
Yang Liu, Dan Iter, Yichong Xu, Shuohang Wang, Ruochen Xu, and Chenguang Zhu.
\newblock {G}-eval: {NLG} evaluation using gpt-4 with better human alignment.
\newblock In Houda Bouamor, Juan Pino, and Kalika Bali (eds.), \emph{Proceedings of the 2023 Conference on Empirical Methods in Natural Language Processing}, pp.\  2511--2522, Singapore, December 2023. Association for Computational Linguistics.
\newblock \doi{10.18653/v1/2023.emnlp-main.153}.
\newblock URL \url{https://aclanthology.org/2023.emnlp-main.153/}.

\bibitem[Loshchilov \& Hutter(2019)Loshchilov and Hutter]{adamw}
Ilya Loshchilov and Frank Hutter.
\newblock Decoupled weight decay regularization.
\newblock \emph{International Conference on Learning Representations (ICLR)}, 2019.

\bibitem[Madaan et~al.(2023)Madaan, Tandon, Gupta, Hallinan, Gao, Wiegreffe, Alon, Dziri, Prabhumoye, Yang, et~al.]{madaan2023self}
Aman Madaan, Niket Tandon, Prakhar Gupta, Skyler Hallinan, Luyu Gao, Sarah Wiegreffe, Uri Alon, Nouha Dziri, Shrimai Prabhumoye, Yiming Yang, et~al.
\newblock Self-refine: Iterative refinement with self-feedback.
\newblock \emph{Advances in Neural Information Processing Systems (NeurIPS)}, 36:\penalty0 46534--46594, 2023.

\bibitem[Meng et~al.(2024)Meng, Xia, and Chen]{preference2}
Yu~Meng, Mengzhou Xia, and Danqi Chen.
\newblock {SimPO}: Simple preference optimization with a reference-free reward.
\newblock \emph{Advances in Neural Information Processing Systems (NeurIPS)}, 2024.

\bibitem[Ouyang et~al.(2022)Ouyang, Wu, Jiang, Almeida, Wainwright, Mishkin, Zhang, Agarwal, Slama, Ray, et~al.]{rl2}
Long Ouyang, Jeffrey Wu, Xu~Jiang, Diogo Almeida, Carroll Wainwright, Pamela Mishkin, Chong Zhang, Sandhini Agarwal, Katarina Slama, Alex Ray, et~al.
\newblock Training language models to follow instructions with human feedback.
\newblock \emph{Advances in neural information processing systems (NeurIPS)}, 2022.

\bibitem[{Qwen Team}(2024{\natexlab{a}})]{qwen2.5}
{Qwen Team}.
\newblock Qwen2.5: A party of foundation models, September 2024{\natexlab{a}}.
\newblock URL \url{https://qwenlm.github.io/blog/qwen2.5/}.

\bibitem[{Qwen Team}(2024{\natexlab{b}})]{qwq-32b-preview}
{Qwen Team}.
\newblock {QwQ}: Reflect deeply on the boundaries of the unknown, November 2024{\natexlab{b}}.
\newblock URL \url{https://qwenlm.github.io/blog/qwq-32b-preview/}.

\bibitem[{Qwen Team}(2025)]{qwq32b}
{Qwen Team}.
\newblock {QwQ-32B}: Embracing the power of reinforcement learning, March 2025.
\newblock URL \url{https://qwenlm.github.io/blog/qwq-32b/}.

\bibitem[Rein et~al.(2024)Rein, Hou, Stickland, Petty, Pang, Dirani, Michael, and Bowman]{rein2024gpqa}
David Rein, Betty~Li Hou, Asa~Cooper Stickland, Jackson Petty, Richard~Yuanzhe Pang, Julien Dirani, Julian Michael, and Samuel~R Bowman.
\newblock {GPQA}: A graduate-level google-proof {Q\&A} benchmark.
\newblock \emph{Conference on Language Modeling (COLM)}, 2024.

\bibitem[Setlur et~al.(2025)Setlur, Nagpal, Fisch, Geng, Eisenstein, Agarwal, Agarwal, Berant, and Kumar]{setlur2024rewarding}
Amrith Setlur, Chirag Nagpal, Adam Fisch, Xinyang Geng, Jacob Eisenstein, Rishabh Agarwal, Alekh Agarwal, Jonathan Berant, and Aviral Kumar.
\newblock Rewarding progress: Scaling automated process verifiers for {LLM} reasoning.
\newblock \emph{International Conference on Learning Representations (ICLR)}, 2025.

\bibitem[She et~al.(2025)She, Liu, Liu, Chen, Huang, and Huang]{she2025rprm}
Shuaijie She, Junxiao Liu, Yifeng Liu, Jiajun Chen, Xin Huang, and Shujian Huang.
\newblock {R}-{PRM}: Reasoning-driven process reward modeling.
\newblock In Christos Christodoulopoulos, Tanmoy Chakraborty, Carolyn Rose, and Violet Peng (eds.), \emph{Proceedings of the 2025 Conference on Empirical Methods in Natural Language Processing}, pp.\  13438--13451, Suzhou, China, November 2025. Association for Computational Linguistics.
\newblock ISBN 979-8-89176-332-6.
\newblock \doi{10.18653/v1/2025.emnlp-main.679}.
\newblock URL \url{https://aclanthology.org/2025.emnlp-main.679/}.

\bibitem[Singhal et~al.(2023)Singhal, Azizi, Tu, Mahdavi, Wei, Chung, Scales, Tanwani, Cole-Lewis, Pfohl, et~al.]{singhal2023large}
Karan Singhal, Shekoofeh Azizi, Tao Tu, S~Sara Mahdavi, Jason Wei, Hyung~Won Chung, Nathan Scales, Ajay Tanwani, Heather Cole-Lewis, Stephen Pfohl, et~al.
\newblock Large language models encode clinical knowledge.
\newblock \emph{Nature}, 620\penalty0 (7972):\penalty0 172--180, 2023.

\bibitem[Singhal et~al.(2025)Singhal, Tu, Gottweis, Sayres, Wulczyn, Amin, Hou, Clark, Pfohl, Cole-Lewis, et~al.]{singhal2025toward}
Karan Singhal, Tao Tu, Juraj Gottweis, Rory Sayres, Ellery Wulczyn, Mohamed Amin, Le~Hou, Kevin Clark, Stephen~R Pfohl, Heather Cole-Lewis, et~al.
\newblock Toward expert-level medical question answering with large language models.
\newblock \emph{Nature Medicine}, 31\penalty0 (3):\penalty0 943--950, 2025.

\bibitem[Snell et~al.(2025)Snell, Lee, Xu, and Kumar]{snell2024scaling}
Charlie~Victor Snell, Jaehoon Lee, Kelvin Xu, and Aviral Kumar.
\newblock Scaling {LLM} test-time compute optimally can be more effective than scaling parameters for reasoning.
\newblock \emph{International Conference on Learning Representations (ICLR)}, 2025.

\bibitem[Uesato et~al.(2022)Uesato, Kushman, Kumar, Song, Siegel, Wang, Creswell, Irving, and Higgins]{uesato2022solving}
Jonathan Uesato, Nate Kushman, Ramana Kumar, Francis Song, Noah Siegel, Lisa Wang, Antonia Creswell, Geoffrey Irving, and Irina Higgins.
\newblock Solving math word problems with process-and outcome-based feedback.
\newblock \emph{arXiv preprint arXiv:2211.14275}, 2022.

\bibitem[Wang et~al.(2023)Wang, Liang, Meng, Sun, Shi, Li, Xu, Qu, and Zhou]{wang2023chatgpt}
Jiaan Wang, Yunlong Liang, Fandong Meng, Zengkui Sun, Haoxiang Shi, Zhixu Li, Jinan Xu, Jianfeng Qu, and Jie Zhou.
\newblock Is {C}hat{GPT} a good {NLG} evaluator? a preliminary study.
\newblock In Yue Dong, Wen Xiao, Lu~Wang, Fei Liu, and Giuseppe Carenini (eds.), \emph{Proceedings of the 4th New Frontiers in Summarization Workshop}, pp.\  1--11, Singapore, December 2023. Association for Computational Linguistics.
\newblock \doi{10.18653/v1/2023.newsum-1.1}.
\newblock URL \url{https://aclanthology.org/2023.newsum-1.1/}.

\bibitem[Wang et~al.(2024{\natexlab{a}})Wang, Li, Shao, Xu, Dai, Li, Chen, Wu, and Sui]{wang2024mathshepherd}
Peiyi Wang, Lei Li, Zhihong Shao, Runxin Xu, Damai Dai, Yifei Li, Deli Chen, Yu~Wu, and Zhifang Sui.
\newblock Math-shepherd: Verify and reinforce {LLM}s step-by-step without human annotations.
\newblock In \emph{Proceedings of the 62nd Annual Meeting of the Association for Computational Linguistics (ACL)}, pp.\  9426--9439, Bangkok, Thailand, 2024{\natexlab{a}}.
\newblock \doi{10.18653/v1/2024.acl-long.510}.

\bibitem[Wang et~al.(2024{\natexlab{b}})Wang, Kulikov, Golovneva, Yu, Yuan, Dwivedi-Yu, Pang, Fazel-Zarandi, Weston, and Li]{wang2024self}
Tianlu Wang, Ilia Kulikov, Olga Golovneva, Ping Yu, Weizhe Yuan, Jane Dwivedi-Yu, Richard~Yuanzhe Pang, Maryam Fazel-Zarandi, Jason Weston, and Xian Li.
\newblock Self-taught evaluators.
\newblock \emph{arXiv preprint arXiv:2408.02666}, 2024{\natexlab{b}}.

\bibitem[Wang et~al.(2024{\natexlab{c}})Wang, Ma, Zhang, Ni, Chandra, Guo, Ren, Arulraj, He, Jiang, et~al.]{mmlu-pro}
Yubo Wang, Xueguang Ma, Ge~Zhang, Yuansheng Ni, Abhranil Chandra, Shiguang Guo, Weiming Ren, Aaran Arulraj, Xuan He, Ziyan Jiang, et~al.
\newblock {MMLU-Pro}: A more robust and challenging multi-task language understanding benchmark.
\newblock \emph{Advances in Neural Information Processing Systems (NeurIPS)}, 2024{\natexlab{c}}.

\bibitem[Wei et~al.(2022)Wei, Wang, Schuurmans, Bosma, Xia, Chi, Le, Zhou, et~al.]{wei2022chain}
Jason Wei, Xuezhi Wang, Dale Schuurmans, Maarten Bosma, Fei Xia, Ed~Chi, Quoc~V Le, Denny Zhou, et~al.
\newblock Chain-of-thought prompting elicits reasoning in large language models.
\newblock \emph{Advances in neural information processing systems (NeurIPS)}, 35:\penalty0 24824--24837, 2022.

\bibitem[Wu et~al.(2025)Wu, Zhang, Dong, Xi, Zhao, Jin, Fan, Zhou, Lv, Zhang, et~al.]{wu2025reasoning}
Mingqi Wu, Zhihao Zhang, Qiaole Dong, Zhiheng Xi, Jun Zhao, Senjie Jin, Xiaoran Fan, Yuhao Zhou, Huijie Lv, Ming Zhang, et~al.
\newblock Reasoning or memorization? unreliable results of reinforcement learning due to data contamination.
\newblock \emph{arXiv preprint arXiv:2507.10532}, 2025.

\bibitem[Wu et~al.(2024)Wu, Sun, Li, Welleck, and Yang]{wu2024inference}
Yangzhen Wu, Zhiqing Sun, Shanda Li, Sean Welleck, and Yiming Yang.
\newblock Scaling inference computation: Compute-optimal inference for problem-solving with language models.
\newblock \emph{The 4th Workshop on Mathematical Reasoning and AI at NeurIPS'24}, 2024.

\bibitem[Yang et~al.(2025)Yang, Li, Yang, Zhang, Hui, Zheng, Yu, Gao, Huang, Lv, et~al.]{yang2025qwen3}
An~Yang, Anfeng Li, Baosong Yang, Beichen Zhang, Binyuan Hui, Bo~Zheng, Bowen Yu, Chang Gao, Chengen Huang, Chenxu Lv, et~al.
\newblock Qwen3 technical report.
\newblock \emph{arXiv preprint arXiv:2505.09388}, 2025.

\bibitem[Yao et~al.(2023{\natexlab{a}})Yao, Yu, Zhao, Shafran, Griffiths, Cao, and Narasimhan]{yao2023tree}
Shunyu Yao, Dian Yu, Jeffrey Zhao, Izhak Shafran, Tom Griffiths, Yuan Cao, and Karthik Narasimhan.
\newblock Tree of thoughts: Deliberate problem solving with large language models.
\newblock \emph{Advances in neural information processing systems (NeurIPS)}, 36:\penalty0 11809--11822, 2023{\natexlab{a}}.

\bibitem[Yao et~al.(2023{\natexlab{b}})Yao, Zhao, Yu, Du, Shafran, Narasimhan, and Cao]{yao2023react}
Shunyu Yao, Jeffrey Zhao, Dian Yu, Nan Du, Izhak Shafran, Karthik Narasimhan, and Yuan Cao.
\newblock {ReAct}: Synergizing reasoning and acting in language models.
\newblock In \emph{International Conference on Learning Representations (ICLR)}, 2023{\natexlab{b}}.

\bibitem[Yu et~al.(2024)Yu, Gao, and Wang]{yu2023ovm}
Fei Yu, Anningzhe Gao, and Benyou Wang.
\newblock {OVM}, outcome-supervised value models for planning in mathematical reasoning.
\newblock In Kevin Duh, Helena Gomez, and Steven Bethard (eds.), \emph{Findings of the Association for Computational Linguistics: NAACL 2024}, pp.\  858--875, Mexico City, Mexico, June 2024. Association for Computational Linguistics.
\newblock \doi{10.18653/v1/2024.findings-naacl.55}.
\newblock URL \url{https://aclanthology.org/2024.findings-naacl.55/}.

\bibitem[Zeng et~al.(2025)Zeng, Zhang, Wu, Classen, Chae, Ewer, Lee, Kim, Kang, Kunde, et~al.]{versaprm}
Thomas Zeng, Shuibai Zhang, Shutong Wu, Christian Classen, Daewon Chae, Ethan Ewer, Minjae Lee, Heeju Kim, Wonjun Kang, Jackson Kunde, et~al.
\newblock {VersaPRM}: Multi-domain process reward model via synthetic reasoning data.
\newblock \emph{International Conference on Machine Learning (ICML)}, 2025.

\bibitem[Zhang et~al.(2025{\natexlab{a}})Zhang, Hosseini, Bansal, Kazemi, Kumar, and Agarwal]{zhang2025generativeverifiers}
Lunjun Zhang, Arian Hosseini, Hritik Bansal, Mehran Kazemi, Aviral Kumar, and Rishabh Agarwal.
\newblock Generative verifiers: Reward modeling as next-token prediction.
\newblock \emph{International Conference on Learning Representations (ICLR)}, 2025{\natexlab{a}}.

\bibitem[Zhang et~al.(2025{\natexlab{b}})Zhang, Zheng, Wu, Zhang, Lin, Yu, Liu, Zhou, and Lin]{zhang2025lessons}
Zhenru Zhang, Chujie Zheng, Yangzhen Wu, Beichen Zhang, Runji Lin, Bowen Yu, Dayiheng Liu, Jingren Zhou, and Junyang Lin.
\newblock The lessons of developing process reward models in mathematical reasoning.
\newblock In Wanxiang Che, Joyce Nabende, Ekaterina Shutova, and Mohammad~Taher Pilehvar (eds.), \emph{Findings of the Association for Computational Linguistics: ACL 2025}, pp.\  10495--10516, Vienna, Austria, July 2025{\natexlab{b}}. Association for Computational Linguistics.
\newblock ISBN 979-8-89176-256-5.
\newblock \doi{10.18653/v1/2025.findings-acl.547}.
\newblock URL \url{https://aclanthology.org/2025.findings-acl.547/}.

\bibitem[Zhao et~al.(2025)Zhao, Liu, Zhang, Zhou, Gao, Li, Lyu, Qian, Qi, Li, et~al.]{zhao2025genprm}
Jian Zhao, Runze Liu, Kaiyan Zhang, Zhimu Zhou, Junqi Gao, Dong Li, Jiafei Lyu, Zhouyi Qian, Biqing Qi, Xiu Li, et~al.
\newblock {GenPRM}: Scaling test-time compute of process reward models via generative reasoning.
\newblock \emph{arXiv preprint arXiv:2504.00891}, 2025.

\bibitem[Zheng et~al.(2024)Zheng, Zhang, Zhang, Lin, Lu, Yu, Liu, Zhou, and Lin]{zheng2024processbench}
Chujie Zheng, Zhenru Zhang, Beichen Zhang, Runji Lin, Keming Lu, Bowen Yu, Dayiheng Liu, Jingren Zhou, and Junyang Lin.
\newblock {PROCESSBENCH}: Identifying process errors in mathematical reasoning.
\newblock \emph{arXiv preprint arXiv:2412.06559}, 2024.

\bibitem[Zheng et~al.(2023)Zheng, Chiang, Sheng, Zhuang, Wu, Zhuang, Lin, Li, Li, Xing, et~al.]{zheng2023judging}
Lianmin Zheng, Wei-Lin Chiang, Ying Sheng, Siyuan Zhuang, Zhanghao Wu, Yonghao Zhuang, Zi~Lin, Zhuohan Li, Dacheng Li, Eric Xing, et~al.
\newblock Judging {LLM}-as-a-judge with {MT}-bench and chatbot arena.
\newblock \emph{Advances in neural information processing systems (NeurIPS)}, 2023.

\bibitem[Zhu et~al.(2025)Zhu, Wang, and Wang]{zhu2023judgelm}
Lianghui Zhu, Xinggang Wang, and Xinlong Wang.
\newblock Judge{LM}: Fine-tuned large language models are scalable judges.
\newblock \emph{International Conference on Learning Representations (ICLR)}, 2025.

\bibitem[Ziegler et~al.(2019)Ziegler, Stiennon, Wu, Brown, Radford, Amodei, Christiano, and Irving]{rl1}
Daniel~M. Ziegler, Nisan Stiennon, Jeffrey Wu, Tom~B. Brown, Alec Radford, Dario Amodei, Paul~F. Christiano, and Geoffrey Irving.
\newblock Fine-tuning language models from human preferences.
\newblock \emph{arXiv preprint arXiv:1909.08593}, 2019.

\end{thebibliography}
\bibliographystyle{tmlr}

\clearpage
\appendix

\crefalias{section}{appendix}
\crefalias{subsection}{appendix}

\section*{Appendix Overview}

This appendix provides supplementary materials to support the main paper as follows:

\begin{itemize}[itemsep=1mm,parsep=1pt,topsep=2pt,leftmargin=*]
    \item \textbf{Theoretical Analysis} (\Cref{sec:theoretical_analysis}): details notations, assumptions, and proofs for \Cref{thm:main_paper_orm,thm:main_paper_dprm,thm:main_paper_gprm-log}.

    \item \textbf{Prompts} (\Cref{sec:prompts}): presents the detailed prompt formats.

    \item \textbf{Datasets} (\Cref{sec:dataset}): describes the datasets used in our experiments.

    \item \textbf{Implementation Details} (\Cref{sec:implementation_details}): provides implementation details, such as (i) backbones for reward models, (ii) hyperparameters, and (iii) verification CoTs for \GenORM and \GenPRM.

    \item \textbf{Training Examples} (\Cref{sec:training_examples}): contains training examples including verification CoTs of \GenORM and \GenPRM in the law domain of MMLU-pro.

    \item \textbf{Detailed Results on MMLU-Pro} (\Cref{sec:additional_experiments}): includes the complete results of \Cref{sec:experimental_results} (omitted in the main paper due to the space limit), such as per-domain results on MMLU-pro using weighted majority voting.

    \item \textbf{Additional Analysis} (\Cref{sec:additional_analysis}): includes the complete results of \Cref{sec:analysis}.

\end{itemize}

\section{Theoretical Analysis}\label{sec:theoretical_analysis}
\subsection{Analysis on Log-Error Bound}\label{sec:log_error_bound}
\paragraph{Notation.} We assume that a correct final step, $y=z_T=1$, implies all previous steps are correct. Define the stepwise conditional probabilities $u_t\coloneqq \Pr(z_t=1\mid x, z_1=1, \ldots, z_{t-1}=1)$ for $t\in [T]$. By the chain rule and the assumption, the true reward function,
\begin{equation*}
    f(x)= p(y=1\mid x)= p(z_T=1\mid x) = p(z_{1:T}=1\mid x)= \prod_{t=1}^T u_t(x)
\end{equation*}
and we write $\zeta(x)\coloneqq \log f(x)=\sum_{t=1}^T \log u_t(x)$. For \DisPRM, we define the stepwise conditional distribution $\hat{u}_t(x)\coloneqq \hat{f}_{\DisPRM}(x_{1:t})$ and use product for the aggregation, \ie, $\hat{f}_{\DisPRM}(x)\coloneqq \prod_{t=1}^T \hat{u}_t(x)$. Similarly, we define the conditional distribution  $F_t(x, v_{\leq t}) \in [0,1]$ to be the \GenPRM's normalized probability that step $t$ is correct given the verification prefix, \ie $\hat{f}_{\GenPRM}(x)\coloneqq \mathbb{E}_{v_{1:L^+}}[\prod_{t=1}^T F_t(x, v_{\leq t})]$. To bound log probability, we assume there is $\varsigma\in(0,1/2]$ such that all probabilities/predictors appearing inside logarithms are clipped into $[\varsigma,1-\varsigma]$. Hence all logs are finite and $\lvert \log(\cdot)\rvert \leq \log (1/\varsigma)$.

\paragraph{Error terms.} 

\begin{enumerate}[itemsep=1mm,parsep=1pt,topsep=2pt,leftmargin=*]
    \item  \DisPRM: Define $\delta_t \coloneqq \log \hat{u}_t - \log u_t$ (evaluated at the appropriate prefixes), and 
\begin{equation*}
    m_t \coloneqq \mathbb{E}[\delta_t \mid x], \quad \xi_t \coloneqq \delta_t - m_t,
\end{equation*}
so $\mathbb{E}[\xi_t\mid x]=0$.
\item \DisORM or \GenORM: Let $\epsilon \in \{\epsilon_d,\epsilon_g\}$,
\begin{equation*}
\epsilon_d \coloneqq \log \hat f_{\DisORM}(x) - \log f(x), \quad
\epsilon_g \coloneqq \log \hat f_{\GenORM}(x) - \log f(x),
\end{equation*}
and decompose 
\begin{equation*}
    \bar m \coloneqq \mathbb{E}[\epsilon\mid x],\quad \bar \xi \coloneqq \epsilon - \bar m, \quad
\beta_{\mathrm{orm}}^2 \coloneqq \mathbb{E}[\bar m^2],
\end{equation*} so that $\mathbb{E}[\bar \xi\mid x]=0$.

\item \GenPRM: For a single rollout $v_{1:L^+}\sim p_{\GenPRM}(\cdot\mid x)$, define
\begin{equation*}\tilde{u}_t \coloneqq F_t(x,v_{\le t}), \quad \tilde{f}_{\GenPRM}(x) \coloneqq \prod_{t=1}^T \tilde u_t.
\end{equation*}
The sampled \GenPRM log-error is
\begin{equation*}\Delta_{\GenPRM}
~\coloneqq~ \log \tilde{f}_{\GenPRM}(x) - \zeta(x)
= \sum_{t=1}^T \delta_t^{(g)}, \quad \delta_t^{(g)} \coloneqq \log \tilde u_t - \log u_t.
\end{equation*}
Let 
\begin{equation*}
    m_t^{(g)} \coloneqq \mathbb{E}[\delta_t^{(g)} \mid x], \quad
\xi_t^{(g)} \coloneqq \delta_t^{(g)} - m_t^{(g)},
\end{equation*} so that $\mathbb{E}[\xi_t^{(g)}\mid x]=0$.
\vspace{-0.05in}
\end{enumerate}  

\paragraph{Assumptions.} There exist constants $\sigma^2>0$, $\rho\ge 0$, and an integer $k \ge 0$ (independent of $T$) such that for all $x$,
\begin{enumerate}[itemsep=1mm,parsep=1pt,topsep=2pt,leftmargin=*]
    
    \item \text{(Average variance floors)} $\frac{1}{T}\sum_{t=1}^T \Var(\xi_t\mid x)\ge \sigma^2$, $\frac{1}{T}\sum_{t=1}^T\Var(\xi_t^{(g)}\mid x)\ \ge\ \sigma^2 + \tau^2$.
    
    \item \text{(Local error dependence)} For any step $s, t$, errors are only correlated within a local context window $k$, such that $\Cov(\xi_s, \xi_t\mid x) = 0$ for $|s - t| > k$. Within this window, the anti-correlation is bounded by $\Cov(\xi_s, \xi_t\mid x) \ge -\rho$. (The same holds for $\xi^{(g)}$).
    
    \item\text{(Positive slope)} $\sigma^2 > 2k\rho$.   
\end{enumerate}
Unlike a strict per-step floor, the average variance condition naturally accommodates deterministic or purely algebraic reasoning steps (where variance is near zero), provided the cumulative chain injects a proportional amount of noise. Similarly, the local dependence assumption reflects bounded memory in autoregressive generation. While a model may self-correct recent errors within a window $k$, distant steps become conditionally independent. Notice that bounded local dependence implies $\sum_{1\le s<t\le T}\Cov(\xi_s,\xi_t\mid x) \ge -k\rho T$. 

For \GenPRM with \emph{sampled} verification CoTs, sampling contributes per-step noise: $\frac{1}{T}\sum_{t=1}^T\Var(\xi_t^{(g)}\mid x)\ge \sigma^2+\tau^2$ for some $\tau^2>0$. For ORMs, assume $\Var(\bar\xi\mid x)\le \tau_{\mathrm{orm}}^2<\infty$ (no $T$-dependence).

\begin{thm}[Log-error bound of \DisORM or \GenORM]
\label{thm:orm}
Let $\epsilon\in\{\epsilon_d,\epsilon_g\}$ and write $\epsilon=\bar m+\bar \xi$ with $\mathbb{E}[\bar \xi\mid x]=0$. If $\Var(\bar \xi\mid x)\le \tau_{\mathrm{orm}}^2$ (independent of $T$), then
\[
\mathbb{E}[\epsilon^2]\ =\ \mathbb{E}[\Var(\bar \xi\mid x)]\;+\;\mathbb{E}[\bar m^2]
\ \le\ \tau_{\mathrm{orm}}^2\;+\;\beta_{\mathrm{orm}}^2,
\]
a bound that does not depend on the CoT length $T$.
\end{thm}

\begin{thm}[Log-error lower bound of \DisPRM]
\label{thm:dprm}
Let $\Delta_{\DisPRM}\coloneqq \log \hat{f}_{\DisPRM}(x)-\zeta(x)$. 
Under the assumptions above,
\[
\mathbb{E}\big[\Delta_{\DisPRM}^2\big]
\ \ge\ 
(\sigma^2-2k\rho)T .
\]
\end{thm}

\begin{thm}[Log-error lower bound of \GenPRM]
\label{thm:gprm-log}
Under the assumptions above,
\[
\mathbb{E}\!\big[\Delta_{\GenPRM}^2\big]
~\ge~
(\sigma^2+\tau^2-2k\rho)T .
\]
\end{thm}

\paragraph{Jensen-gap representation (mean predictor).}
Let $L(x,v)\coloneqq \sum_{t=1}^T \log F_t(x,v_{\le t})$ and
$K_x(\theta)\coloneqq \log \mathbb{E}[e^{\theta L}\mid x]$. Define the mean predictor $\mu(x)\coloneqq \mathbb{E}[e^{L}\mid x]$ and $\Delta_{\mathrm{mean}}(x)\coloneqq \log\mu(x)-\zeta(x)$. Then with $B^{(g)}(x)\coloneqq \mathbb{E}[L\mid x]-\zeta(x)$, we have the exact decomposition
\begin{equation*}
\Delta_{\mathrm{mean}}(x)=B^{(g)}(x)+\delta_J(x),\quad \delta_J(x)=K_x(1)-K_x'(0)=\int_0^1 (1-\theta)\,\Var_{\theta}(L\mid x)\,d\theta\ge 0,
\end{equation*}
where $\Var_{\theta}$ denotes variance under the exponentially tilted law $d\mathbb{P}_{\theta}\propto e^{\theta L}d\mathbb{P}$, \ie,
$d\mathbb{P}_\theta(v)=\mathbbm{1}\{M(\theta)>0\}\,e^{\theta L(x,v)}\,M(\theta)^{-1} d\mathbb{P}(v)$ with $M(\theta)\coloneqq\mathbb{E}[e^{\theta L}\mid x]$.

\begin{thm}[Log-error lower bound of mean-\GenPRM]
\label{thm:meanprm}
Assume the conditions of Theorem~\ref{thm:gprm-log}. In addition, suppose there exists $\kappa\in(0,1]$ such that for all $\theta\in[0,1]$,
\[
\Var_{\theta}(L\mid x)\ \ge\ \kappa\,\Var(L\mid x).
\]
Then, for every $x$,
\[
\Delta_{\mathrm{mean}}(x)\ \ge\ B^{(g)}(x)\; +\; \frac{\kappa}{2}\,\Var(L\mid x)
\ \ge\ B^{(g)}(x)\; +\; \frac{\kappa}{2}\Big((\sigma^2+\tau^2-2k\rho)T\Big).
\]
Consequently,
\[
\mathbb{E}[\Delta_{\mathrm{mean}}]\ \ge\ \frac{\kappa}{2}\Big((\sigma^2+\tau^2-2k\rho)T\Big)\; -\; \sqrt{\mathbb{E}[B^{(g)}(x)^2]},
\quad
\mathbb{E}[\Delta_{\mathrm{mean}}^2]\ \ge\ \left(\max\{0, \mathbb{E}[\Delta_{\mathrm{mean}}] \} \right)^2.
\]
\end{thm}

\paragraph{Takeaways.}
Under mild anti-correlation and variance-floor assumptions, \DisPRM\ and sampled \GenPRM\ incur log-error that grows at least linearly in the CoT length $T$, and the additional sampling noise $\tau^2$ makes \GenPRM\ strictly worse. In contrast, ORM estimators admit error bounds that are independent of $T$ provided the conditional noise is bounded, which makes them preferable for long CoTs. For mean-\GenPRM, the Jensen gap introduces a strictly nonnegative bias that scales with the variance of $L$ and hence with $T$, so even a calibrated predictor ($B^{(g)} =0$) exhibits error that increases with chain length. All proofs are deferred to~\Cref{app:proof}.

\subsection{Proofs}
\label{app:proof}

\paragraph{Proof of \texorpdfstring{\Cref{thm:orm}}{}}

\begin{proof}
By the conditional bias--variance decomposition (law of total variance),
\[
\mathbb{E}[\epsilon^2]
= \mathbb{E}\big[\Var(\epsilon\mid x)\big] + \mathbb{E}\big[(\mathbb{E}[\epsilon\mid x])^2\big]
= \mathbb{E}\big[\Var(\bar \xi\mid x)\big] + \mathbb{E}\big[\bar m^2\big].
\]
The assumption $\Var(\bar \xi\mid x)\le \tau_{\mathrm{orm}}^2$ for all $x$ gives
$\mathbb{E}[\Var(\bar \xi\mid x)]\le \tau_{\mathrm{orm}}^2$, and by definition
$\beta_{\mathrm{orm}}^2=\mathbb{E}[\bar m^2]$.
\end{proof}

\paragraph{Proof of \texorpdfstring{\Cref{thm:dprm}}{}.}
\begin{proof}
Let 
\[
B \coloneqq \sum_{t=1}^T m_t,\qquad 
N \coloneqq \sum_{t=1}^T \xi_t,
\]
so $\Delta_{\DisPRM}=B+N$ with $\mathbb{E}[N\mid x]=0$. 
By the tower property,
\begin{align*}
\mathbb{E}\!\left[\Delta_{\DisPRM}^2\right]
&=\mathbb{E}\!\big[\mathbb{E}[(B+N)^2\mid x]\big] 
= \mathbb{E}\!\big[B^2 + 2B \mathbb{E}[N \mid x] + \mathbb{E}[N^2\mid x]\big]\\
&=\mathbb{E}\!\big[\Var(N\mid x)\big]+\mathbb{E}[B^2] \ \ge\ \mathbb{E}\!\big[\Var(N\mid x)\big].
\end{align*}
Expanding the variance yields:
\[
\Var(N\mid x)=\sum_{t=1}^T\Var(\xi_t\mid x)
+2\!\!\sum_{1\le s<t\le T}\!\!\Cov(\xi_s, \xi_t\mid x).
\]
By local dependence, $\Cov(\xi_s, \xi_t\mid x)=0$ when $|s-t|>k$. Hence only pairs with $t-s\le k$ contribute. For each $t$, there are at most $k$ indices $s<t$ with $t-s\le k$, so the number of contributing pairs is at most $kT$. Since each such covariance is bounded below by $-\rho$, we have
\[
\sum_{1\le s<t\le T}\Cov(\xi_s, \xi_t\mid x)\ge -k\rho T,
\]
and therefore $2\sum_{s<t}\Cov(\cdot)\ge -2k\rho T$.
Combining the variance floor and the covariance bound, for every fixed $x$,
\begin{align*}
\Var(N\mid x)
&= \sum_{t=1}^T\Var(\xi_t\mid x)
+2\sum_{1\le s<t\le T}\Cov(\xi_s,\xi_t\mid x) \\
&\ge \sum_{t=1}^T\Var(\xi_t\mid x) - 2k\rho T \\
&\ge T\sigma^2 - 2k\rho T .
\end{align*}
Taking expectations over $x$ preserves inequalities, so
\[
\mathbb{E}\big[\Var(N\mid x)\big]
\ \ge\ 
\mathbb{E}\big[T\sigma^2-2k\rho T\big]
=
(\sigma^2-2k\rho)T.
\]
Substituting back into the earlier bound
$\mathbb{E}[\Delta_{\DisPRM}^2]\ge \mathbb{E}[\Var(N\mid x)]$
yields
\[
\mathbb{E}\!\left[\Delta_{\DisPRM}^2\right]\ \ge\ (\sigma^2-2k\rho)T,
\]
as claimed.
\end{proof}

\paragraph{Proof of \texorpdfstring{\Cref{thm:gprm-log}}{}}

\begin{proof}
Decompose
\[
\Delta_{\GenPRM}
= \sum_{t=1}^T \delta_t^{(g)}
= \underbrace{\sum_{t=1}^T m_t^{(g)}}_{\eqqcolon\,B^{(g)}}
+ \underbrace{\sum_{t=1}^T \xi_t^{(g)}}_{\eqqcolon\,N^{(g)}}.
\]
Conditional mean-zero $\mathbb{E}[N^{(g)}\mid x]=0$ implies
\[
\mathbb{E}\!\left[(\Delta_{\GenPRM})^2\right]
= \mathbb{E}\!\left[\Var\!\left(N^{(g)}\mid x\right)\right]
+ \mathbb{E}\!\left[(B^{(g)})^2\right]
\ \ge\ \mathbb{E}\!\left[\Var\!\left(N^{(g)}\mid x\right)\right].
\]
Now expand $\Var(N^{(g)}\mid x)$ using the average variance floor and local error dependence:
\begin{align*}
\Var(N^{(g)}\mid x)
&= \sum_{t=1}^T \Var(\xi_t^{(g)}\mid x)
+ 2\sum_{1\le s<t\le T}\Cov(\xi_s^{(g)},\xi_t^{(g)}\mid x) \\
&\ge T(\sigma^2+\tau^2) - 2k\rho T .
\end{align*}
Taking expectations in $x$ gives the stated bound.
\end{proof}

\paragraph{Proof of \texorpdfstring{\Cref{thm:meanprm}}{}}

\begin{proof}
\textbf{1) Exponential tilting and log-mgf.} 
Define $M(\theta)\coloneqq \mathbb{E}[e^{\theta L}\mid x]$ and $K_x(\theta)\coloneqq \log M(\theta)$.
Since $e^{\theta L}\in(0,1]$ for $\theta\in[0,1]$ and $\mathbb{E}[|L|^2]<\infty$, dominated convergence yields
$M'(\theta)=\mathbb{E}[L e^{\theta L}\mid x]$ and $M''(\theta)=\mathbb{E}[L^2 e^{\theta L}\mid x]$.
Let $d\mathbb{P}_\theta(C)\coloneqq e^{\theta L(x,C)} M(\theta)^{-1} d\mathbb{P}(C)$ and
$\mathbb{E}_\theta[\cdot]\coloneqq \mathbb{E}[\cdot\, e^{\theta L}]/M(\theta)$.
Then
\[
K_x'(\theta)=\frac{M'(\theta)}{M(\theta)}=\mathbb{E}_\theta[L\mid x],
\qquad
K_x''(\theta)=\frac{M''(\theta)M(\theta)-(M'(\theta))^2}{M(\theta)^2}
=\Var_\theta(L\mid x).
\]

\textbf{2) Jensen-gap identity.} Taylor with integral remainder at $\theta=0$ gives
\[
K_x(1)=K_x(0)+K_x'(0)+\int_0^1 (1-\theta)K_x''(\theta)\,d\theta.
\]
Since $K_x(0)=0$ and $K_x'(0)=\mathbb{E}[L\mid x]$, we obtain
\[
\log\mu(x)=\mathbb{E}[L\mid x]+\int_0^1 (1-\theta)\,\Var_\theta(L\mid x)\,d\theta.
\]
By definition of the mean predictor, 
$$
\Delta_{\mathrm{mean}}(x)= \log\mu(x)-\zeta(x), \text{ where } \mu(x)=\mathbb{E}[e^L\mid x].
$$
Plugging $\log\mu(x)= \Delta_{\mathrm{mean}}(x) + \zeta(x)$ with $B^{(g)}(x)\coloneqq \mathbb{E}[L\mid x]-\zeta(x)$, this yields
\[
\Delta_{\mathrm{mean}}(x)=B^{(g)}(x)+\delta_J(x),\quad \delta_J(x)\coloneqq \int_0^1 (1-\theta)\,\Var_\theta(L\mid x)\,d\theta\;\ge 0.
\]

\textbf{3) Lower bound on $\delta_J$ and variance linkage.} By tilt-stability,
\[
\delta_J(x)\ge \frac{\kappa}{2}\,\Var(L\mid x).
\]
Moreover, since $L=\zeta(x)+\Delta_{\GenPRM}=\zeta(x)+B^{(g)}+N^{(g)}$ with $\mathbb{E}[N^{(g)}\mid x]=0$, and since $\zeta(x)$ and $B^{(g)}(x)$ are constants when conditioning on $x$, we have
\[
\Var(L\mid x)=\Var(N^{(g)}\mid x).
\]
Expanding and using the average variance floors and local error dependence conditions (as in Theorem~\ref{thm:gprm-log}),
\begin{equation*}
\Var(N^{(g)}\mid x)\ \ge\ T(\sigma^2+\tau^2) - 2k\rho T.    
\end{equation*}
Combining this gives the pointwise bound
\[
\Delta_{\mathrm{mean}}(x)\ \ge\ B^{(g)}(x)+\frac{\kappa}{2}\Big(T(\sigma^2+\tau^2)-2k\rho T\Big).
\]

\textbf{4) Expectations and MSE.} Taking expectations over $x$ and applying Cauchy--Schwarz to $\mathbb{E}[B^{(g)}(x)]$ yields
\[
\mathbb{E}[\Delta_{\mathrm{mean}}]\ \ge\ \frac{\kappa}{2}\Big((\sigma^2+\tau^2-2k\rho)T\Big) - \sqrt{\mathbb{E}[B^{(g)}(x)^2]}.
\]
Finally, Jensen's inequality gives $\left(\max\{0, \mathbb{E}[\Delta_{\mathrm{mean}}]\}\right)^2\le \mathbb{E}[\Delta_{\mathrm{mean}}^2]$, so the MSE bound follows. In the calibrated case $B^{(g)}\equiv 0$, the stated simplified bounds hold.
\end{proof}

\section{Prompts}\label{sec:prompts}

In this section, we present prompt formats used in this work:

\begin{itemize}[itemsep=1mm,parsep=1pt,topsep=2pt,leftmargin=*]
\item \Cref{prompt:versaprm_prompt-math}: \textbf{User prompt format for generating CoTs} on GSM8K~\citep{cobbe2021training} and MATH~\citep{math_dataset}.

\item \Cref{prompt:versaprm_prompt}: \textbf{User prompt format for generating CoTs} on MMLU-Pro~\citep{mmlu-pro} proposed by \citet{versaprm}.

\item \Cref{prompt:versaprm_system}: \textbf{System prompt format for auto-labeling process labels} on MMLU-Pro~\citep{mmlu-pro} proposed by \citet{versaprm}.

\item \Cref{prompt:versaprm_user}: \textbf{User prompt format for auto-labeling process labels} on MMLU-Pro~\citep{mmlu-pro} proposed by \citet{versaprm}.

\item \Cref{prompt:orm_style_user}: \textbf{Prompt format of \GenORM}~\citep{zhang2025generativeverifiers}. We use this format for both generating synthetic verification-CoTs and training/evaluation of \GenORM.

\item \Cref{prompts:prm_style_data_generation}: \textbf{Prompt format for generating verification-CoTs} for \GenPRM following \citet{thinkprm}.

\item \Cref{prompt:prm_style_user}: \textbf{Prompt format of \GenPRM} for training and evaluation.
\end{itemize}

\begin{figure}[H]
\vspace{-0.2in}
\begin{tcolorbox}[llmprompt]
\small
\textbf{[user]}
Solve the following math problem efficiently and clearly:
\newline
\newline
- For simple problems (2 steps or fewer):
\newline
Provide a concise solution with minimal explanation.
\newline
\newline
- For complex problems (3 steps or more):
\newline
Use this step-by-step format:
\newline
\#\# Step 1: [Concise description]
\newline
[Brief explanation and calculations]
\newline
\#\# Step 2: [Concise description]
\newline
[Brief explanation and calculations]
\newline
\texttt{[OMITTED...]}
\newline
\newline
Regardless of the approach, always conclude with:
\newline
\verb|Therefore, the final answer is: $\\boxed{answer}$|.
\newline
\newline
I hope it is correct. Where \texttt{[answer]} is just the final number or expression that solves the problem.
\newline
\newline
[Problem] \newline
\textit{\{problem\}} \newline
\textbf{[/user]}
\textbf{[assistant]}
\end{tcolorbox}
\vspace{-0.1in}
\caption{\textbf{User prompt format for generating CoTs} on GSM8K~\citep{cobbe2021training} and MATH~\citep{math_dataset}.}
\label{prompt:versaprm_prompt-math}
\end{figure}

\begin{figure}[H]
\vspace{-0.2in}
\begin{tcolorbox}[llmprompt]
\small
\textbf{[user]}
Given the following question and candidate answers, choose the best answer.
\newline
[Question]
\newline
\textit{\{question \#1\}}
\newline
\textbf{[/user]}
\textbf{[assistant]}
\newline
\textit{\{assistant response \#1\}}
\newline
\textbf{[/assistant]}
\newline
\newline
\textbf{[user]}
Given the following question and candidate answers, choose the best answer.
\newline
[Question]
\newline
\textit{\{question \#2\}}
\newline
\textbf{[/user]}
\textbf{[assistant]}
\newline
\textit{\{assistant response \#2\}}
\newline
\textbf{[/assistant]}
\newline
\texttt{[OMITTED...]}
\newline
\newline
\textbf{[user]}
Given the following question and candidate answers, choose the best answer.
\newline
[Question] \newline
\textit{\{question\}} \newline
\textbf{[/user]}
\textbf{[assistant]}

\end{tcolorbox}
\vspace{-0.1in}
\caption{\textbf{User prompt format for generating CoTs} on MMLU-Pro~\citep{mmlu-pro} proposed by \citet{versaprm}}
\label{prompt:versaprm_prompt}
\end{figure}

\begin{figure}[H]
\begin{tcolorbox}[llmprompt]
\small
\textbf{[system]}
You are an experienced evaluator specializing in assessing the quality of reasoning steps in problem-solving. Your task is to find the first BAD step in a student's solution to a multiple choice question.

You will judge steps as GOOD, OK, or BAD based on the following criteria:

\textbf{1. GOOD Step}  
A step is classified as GOOD if it meets all of these criteria:
\begin{itemize}
  \item \textbf{Correct}: Everything stated is accurate and aligns with known principles or the given problem.
  \item \textbf{Verifiable}: The step can be verified using common knowledge, simple calculations, or a quick reference (e.g., recalling a basic theorem). If verifying requires extensive effort (e.g., detailed calculations or obscure references), mark it BAD instead.
  \item \textbf{Appropriate}: The step fits logically within the context of the preceding steps. If a prior mistake exists, a GOOD step can correct it.
  \item \textbf{Insightful}: The step demonstrates reasonable problem-solving direction. Even if ultimately progressing in the wrong direction, it is acceptable as long as it represents a logical approach.
\end{itemize}

\textbf{2. OK Step}  
A step is classified as OK if it is:
\begin{itemize}
  \item \textbf{Correct and Verifiable}: Contains no errors and can be verified.
  \item \textbf{Unnecessary or Redundant}: Adds little value, such as restating prior information or providing basic encouragement (e.g., “Good job!”).
  \item \textbf{Partially Progressing}: Makes some progress toward the solution but lacks decisive or significant advancement.
\end{itemize}

\textbf{3. BAD Step}  
A step is classified as BAD if it:
\begin{itemize}
  \item \textbf{Is Incorrect}: Contains factual errors, misapplies concepts, derives an incorrect result, or contradicts the ground truth answer.
  \item \textbf{Is Hard to Verify}: Requires significant effort to confirm due to poor explanation.
  \item \textbf{Is Off-Topic}: Includes irrelevant or nonsensical information.
  \item \textbf{Derails}: Leads to dead ends, circular reasoning, or unreasonable approaches.
\end{itemize}

\textbf{Task Description} \newline
You will be provided with:
\begin{enumerate}
  \item A Multiple Choice Question
  \item A Ground Truth Answer
  \item A Student's Step-by-Step Solution, where each step is enclosed with tags and indexed from 0.
\end{enumerate}

Once you identify a BAD step, return the index of the earliest BAD step. Otherwise, return the index of -1 (which denotes all steps are GOOD or OK).  
Please put your final answer (i.e., the index) in \verb|\boxed{}|.
\textbf{[/system]}
\end{tcolorbox}
\vspace{-0.1in}
\caption{\textbf{System prompt format for auto-labeling process labels} on MMLU-Pro~\citep{mmlu-pro} proposed by \citet{versaprm}}
\label{prompt:versaprm_system}
\end{figure}
\begin{figure}[H]
\begin{tcolorbox}[llmprompt]
\small
\textbf{[user]}
The following is a multiple choice question and its ground truth answer. You are also given a student’s solution (split into steps, enclosed with tags and indexed from 0): \newline

[Multiple Choice Question] \newline
\textit{\{question\}} \newline

[Ground Truth Answer] \newline
\textit{\{answer\}} \newline

[Student Solution] \newline
\textit{\{solution\}} \newline

\textbf{[/user]} 
\textbf{[assistant]} The first BAD step index is:
\end{tcolorbox}
\vspace{-0.1in}
\caption{\textbf{User prompt format for auto-labeling process labels} on MMLU-Pro~\citep{mmlu-pro} proposed by \citet{versaprm}}
\label{prompt:versaprm_user}
\end{figure}

\begin{figure}[H]
\vspace{-0.1in}
\begin{tcolorbox}[llmprompt]
\small
\textbf{[user]}
You are a \{\texttt{category}\} teacher. Grade the solution, verifying correctness step by step. \newline
At the end of Solution verification, when you give your final grade, write it in the form ``Verification: Is the answer correct (Yes/No)? X'', where X is either Yes or No.\newline

[\{\texttt{Category}\} Problem] \newline
\{\texttt{problem}\} \newline

[Solution] \newline
\{\texttt{solution}\} \newline \textbf{[/user]} 
\textbf{[assistant]} \textbf{[think]} Let's verify step by step:
\end{tcolorbox}
\vspace{-0.2in}
\caption{\textbf{Prompt format of \GenORM}~\citep{zhang2025generativeverifiers}. We use this format for both generating synthetic verification-CoTs and training/evaluation of \GenORM.}
\label{prompt:orm_style_user}
\end{figure}

\begin{figure}[H]
\begin{tcolorbox}[llmprompt]
\small
\textbf{[user]}
You are given a \textit{\{category\}} problem and a proposed multiple-step solution (with a step on each line): \newline

[\textit{\{Category\}} Problem] \newline
\textit{\{question\}} \newline

[Solution] \newline
\textit{\{solution\}} \newline

Review and critique the proposed solution steps and determine whether each step is correct.  
If the solution is incomplete, only critique the steps that are provided.  
Your output must be in the following format: \newline

Step 1: The step is \verb|\boxed{correct/incorrect}| \\
Step 2: The step is \verb|\boxed{correct/incorrect}| \\
\vdots \\
Step $n$: The step is \verb|\boxed{correct/incorrect}| \newline

Once you find an incorrect step, you should stop since you do not need to analyze the remaining steps.  
If the solution is incomplete, only verify the provided steps. \textbf{[/user]} \textbf{[assistant]} \textbf{[think]} Let's verify step by step:
\end{tcolorbox}
\vspace{-0.2in}
\caption{\textbf{Prompt format for generating verification-CoTs} for \GenPRM following \citet{thinkprm}.}
\label{prompts:prm_style_data_generation}
\end{figure}

\begin{figure}[H]
\vspace{-0.1in}
\begin{tcolorbox}[llmprompt]
\small
\textbf{[user]}
You are given a \{\texttt{category}\} problem and a proposed step-by-step solution: \newline

[\{\texttt{category}\} Problem] \newline
\{\texttt{problem}\} \newline

[Solution] \newline
\{\texttt{solution}\} \newline

Review and critique each step in the proposed solution to determine whether each step is correct.  
If the solution is incomplete, only verify the provided steps.
\textbf{[/user]} \textbf{[assistant]} \textbf{[think]} Let's verify step by step:
\end{tcolorbox}
\vspace{-0.2in}
\caption{\textbf{Prompt format of \GenPRM} for training and evaluation.}
\vspace{-0.2in}
\label{prompt:prm_style_user}
\end{figure}

\section{Dataset}\label{sec:dataset}
In this section, we provide more details on the datasets.
\paragraph{Math Datasets.}
For the math domain, we use the widely adopted \textbf{PRM800K}~\citep{prm800k} for training, where the process labels $z_{1:T}$ are human-annotated. For training ORMs, we set the outcome label $y=\mathbbm{1}\!\big(z_{1:T}=\mathbf{1}_T\big)$ (rather than $y=\mathbbm{1}(\hat{a}(r_T)=a)$), since PRM800K provides high-quality ground-truth process labels. As a testbed, we use \href{https://huggingface.co/datasets/Qwen/ProcessBench}{\textbf{ProcessBench}}~\citep{zheng2024processbench}, which comprises four splits: 400 CoTs from GSM8K~\citep{cobbe2021training}, 1K from Math~\citep{math_dataset}, 1K from Omni-Math~\citep{omni-math_dataset}, and 1K from OlympiadBench~\citep{olympiadbench_dataset}. We evaluate outcome verification by predicting $y\in\{0,1\}$ using the \texttt{final\_answer\_correct} field. We also generate $N=16$ CoTs per question with \href{https://huggingface.co/Qwen/Qwen2.5-7B-Instruct}{Qwen2.5-7B-Instruct}~\citep{qwen2.5} to assess test-time scaling (TTS).

\begin{table}[H]
\caption{Dataset statistics for each domain of MMLU-pro~\citep{mmlu-pro}. We report the number of questions, the number of CoTs, and the average number of CoTs per question for both training and test splits.}
\vspace{-1em}
\label{tab:dataset-stats}
\resizebox{\textwidth}{!}{
\begin{tabular}{lcccccc}
\toprule
\multirow{2}{*}{\textbf{Domain}} & \multicolumn{3}{c}{\textbf{Training Set}} & \multicolumn{3}{c}{\textbf{Test Set}} \\
\cmidrule(lr){2-4} \cmidrule(lr){5-7}
 & \textbf{\# Questions} & \textbf{\# CoTs} & \textbf{Avg. CoTs / Q} & \textbf{\# Questions} & \textbf{\# CoTs} & \textbf{Avg. CoTs / Q} \\
\midrule
Law & 500 & 7,806 & 15.61 & 145 & 18,537 & 127.84 \\
Psychology & 498 & 7,901 & 15.87 & 150 & 19,164 & 127.76 \\
Chemistry & 500 & 6,537 & 13.07 & 150 & 15,981 & 106.54 \\
Biology & 417 & 6,420 & 15.40 & 130 & 16,441 & 126.47 \\
Physics & 500 & 6,680 & 13.36 & 150 & 16,460 & 109.73 \\
History & 81 & 1,275 & 15.74 & 150 & 19,159 & 127.73 \\
Economics & 500 & 7,749 & 15.50 & 150 & 18,911 & 126.07 \\
Math & 500 & 6,940 & 13.88 & 150 & 17,014 & 113.43 \\
Business & 489 & 6,969 & 14.25 & 149 & 17,344 & 116.40 \\
Philosophy & 199 & 3,125 & 15.70 & 149 & 18,844 & 126.47 \\
Health & 456 & 7,202 & 15.79 & 140 & 17,862 & 127.59 \\
Engineering & 500 & 6,032 & 12.06 & 150 & 15,708 & 104.72 \\
Computer Science & 110 & 1,638 & 14.89 & 150 & 18,429 & 122.86 \\
Other & 500 & 7,824 & 15.65 & 150 & 18,982 & 126.55 \\
\midrule
\textbf{Total} & 5,750 & 84,098 & 14.63 & 2,063 & 248,836 & 120.62 \\
\bottomrule
\end{tabular}
}
\vspace{-0.15in}
\end{table}

\paragraph{Multi-domain datasets.} For the multi-domain setting, we adopt \textbf{MMLU-Pro}~\citep{mmlu-pro}, a 10-choice benchmark spanning 14 domains: law, psychology, chemistry, biology, physics, history, economics, math, business, philosophy, health, engineering, computer science, and other. As shown in \Cref{tab:dataset-stats}, the corpus includes 5{,}750 training and 2{,}063 evaluation questions. For each question, \citet{versaprm} generate 16/128 CoTs for training/evaluation with \href{https://huggingface.co/meta-llama/Llama-3.1-8B-Instruct}{Llama-3.1-8B-Instruct}~\citep{llama3}, and auto-label reasoning steps (\ie, process labels) using \href{https://huggingface.co/meta-llama/Llama-3.1-70B-Instruct}{Llama-3.1-70B-Instruct} with prompts in \Cref{prompt:versaprm_system,prompt:versaprm_user}; please see \citet{versaprm} for more details. To assess generalization across CoTs from different $p_{\mathtt{LLM}}$, we also generate 16 CoTs per evaluation question using \href{https://huggingface.co/HuggingFaceTB/SmolLM3-3B}{SmolLM3-3B}~\citep{bakouch2025smollm3}, \href{https://huggingface.co/Qwen/Qwen2.5-7B-Instruct}{Qwen2.5-7B-Instruct}, \href{https://huggingface.co/google/gemma-2-9b-it}{gemma-2-9b-it}~\citep{team2024gemma}, and \href{https://huggingface.co/meta-llama/Llama-3.1-70B-Instruct}{Llama-3.1-70B-Instruct}, spanning diverse model sizes and families.

\section{Implementation Details}\label{sec:implementation_details}
In this section, we provide implementation details omitted from the main paper due to space limits.

\paragraph{Backbones for reward models.}
Following \citet{zhang2025generativeverifiers} and \citet{thinkprm}, we use \href{https://huggingface.co/deepseek-ai/DeepSeek-R1-Distill-Qwen-1.5B}{R1-Distill-Qwen-1.5B} and \href{https://huggingface.co/deepseek-ai/DeepSeek-R1-Distill-Qwen-7B}{R1-Distill-Qwen-7B}~\citep{guo2025deepseek} for the math domain, and \href{https://huggingface.co/deepseek-ai/DeepSeek-R1-Distill-Llama-8B}{R1-Distill-Llama-8B} and \href{https://huggingface.co/deepseek-ai/DeepSeek-R1-Distill-Qwen-14B}{R1-Distill-Qwen-14B} for the multi-domain setting, as reward-model backbones. We also use \href{https://huggingface.co/Qwen/Qwen3-8B}{Qwen3-8B} as the backbone to assess whether the results hold for non-distilled backbones in the multi-domain setting. Note that VersaPRM~\citep{versaprm} originally used \href{https://huggingface.co/meta-llama/Llama-3.1-8B-Instruct}{Llama-3.1-8B-Instruct} as the reward-model backbone for \DisPRM; for a fair comparison, we use \href{https://huggingface.co/collections/deepseek-ai/deepseek-r1-678e1e131c0169c0bc89728d}{R1-Distill
 models} and \href{https://huggingface.co/Qwen/Qwen3-8B}{Qwen3-8B} for both \DisORM and \DisPRM.

\begin{table}[H]
\centering
\vspace{-0.1in}
\caption{\small\textbf{Summary of hyperparameters}.}
\label{tab:hyperparameters}
\vspace{-0.1in}
\resizebox{\textwidth}{!}{%
\begin{tabular}{@{}l|ccc|cccccc|ccc@{}}
\toprule
\multirow{2}{*}{\textbf{Method}} & 
\multicolumn{3}{c|}{\textbf{LoRA}} & 
\multicolumn{6}{c|}{\textbf{Training}} & 
\multicolumn{3}{c}{\textbf{Inference}} \\ 
\cmidrule(lr){2-4}\cmidrule(lr){5-10}\cmidrule(lr){11-13}
& Rank $r$ & $\alpha$ & Dropout $p$ 
& Batch & Optim. & Epochs & LR & Decay & Scheduler 
& Package & Temp. $\tau$ & $M$ \\
\midrule
\DisORM \& \DisPRM & 16 & 32 & 0.1 & 16 & AdamW & 1 & $10^{-4}$ & $10^{-2}$ & Cosine & - & - & - \\
\GenORM \& \GenPRM & 32 & 16 & 0.1 & 16 & AdamW & 1 & $10^{-4}$ & $10^{-2}$  & Linear & vLLM & 0.6 & 10 or 16 \\
\bottomrule
\end{tabular}%
}
\vspace{-0.1in}
\end{table}

\paragraph{Hyperparameters.}
We apply LoRA~\citep{hu2022lora} for parameter-efficient fine-tuning, optimize with AdamW~\citep{adamw}, and use vLLM~\citep{kwon2023efficient} for fast inference.
At inference, we sample $M{=}16$ verification CoTs for the math domain and $M{=}10$ for the multi-domain setting.
Hyperparameters are summarized in \Cref{tab:hyperparameters}: for \DisORM/\DisPRM we adopt those of \citet{versaprm}, and for \GenORM/\GenPRM we follow \citet{thinkprm}.
Note that in preliminary experiments we set $r{=}32$ and $\alpha{=}16$ for \DisORM/\DisPRM to compare fairly with \GenORM/\GenPRM (also using $r{=}32$ and $\alpha{=}16$). However, we observed an \textbf{overall performance degradation} (\eg, $\approx$2\%), so we follow the settings of \citet{versaprm}.
The hyperparameters in \Cref{tab:hyperparameters} are shared across all experiments and we do not perform exhaustive tuning\footnote{In \Cref{tab:hyperparameter_search}, we report the results of sweeps over the learning rate and LoRA rank for \DisPRM/\GenPRM.}.
We report means over five independent runs, except for experiments using \href{https://huggingface.co/deepseek-ai/DeepSeek-R1-Distill-Llama-8B}{R1-Distill-Llama-8B} and \href{https://huggingface.co/Qwen/Qwen3-8B}{Qwen3-8B} reward backbones.


\paragraph{Verification CoTs for \GenORM and \GenPRM.}
Following \citet{thinkprm}, we sample 4 different verification CoTs for each question $q$ and CoT $r_{1:T}$ pair in the training dataset by prompting \href{https://huggingface.co/Qwen/QwQ-32B}{QwQ-32B}~\citep{qwq32b} with \texttt{temperature}=0.6, \texttt{top\_k}=20, \texttt{top\_p}=0.95, and \texttt{min\_p}=0 using the formats in \Cref{prompt:orm_style_user,prompts:prm_style_data_generation}. Note that \citet{thinkprm} originally used \href{https://huggingface.co/Qwen/QwQ-32B-Preview}{QwQ-32B-Preview}~\citep{qwq-32b-preview}. In preliminary experiments, we found \href{https://huggingface.co/Qwen/QwQ-32B}{QwQ-32B} more likely to follow instructions and produce more parsable verification CoTs (\eg, 1K vs. 7K for \GenPRM in the law domain), so we use \href{https://huggingface.co/Qwen/QwQ-32B}{QwQ-32B} throughout.

For the math domain we set \texttt{category} as \texttt{math}; for the multi-domain setting we use $\texttt{category}\in$\{\texttt{law}, \ldots, \texttt{computer science}\} and leave it blank for \texttt{other}. For additional benchmarks, such as GPQA-diamond, MedQA, and LEXam, we also leave \texttt{category} blank. We discard any verification CoT that (i) has \textbf{unparsable labels}; (ii) contains \textbf{Chinese characters}; (iii) exceeds \textbf{the token limit}—4{,}096 for math~\citep{thinkprm} or 8{,}192 for multi-domain; or (iv) whose parsed labels are \textbf{inconsistent with the targets} (\eg, $y$ or $z_{1:T}$), corresponding to the \emph{consensus filtering} in \Cref{sec:reward_models}. We also balance the counts of \texttt{Yes}/\texttt{No} examples. The resulting training sets contain 34{,}286 CoTs for \GenORM and 35{,}666 CoTs for \GenPRM in the math domain, and 171{,}780 (\GenORM) and 94{,}156 (\GenPRM) in the multi-domain setting.

\section{Training Examples}\label{sec:training_examples}
In this section, we present training examples including verification CoTs:

\begin{itemize}[itemsep=1mm,parsep=1pt,topsep=2pt,leftmargin=*]
\item \Cref{prompt:orm_style_example}: \textbf{Training example of \GenORM} on the law domain of MMLU-pro~\citep{mmlu-pro}.

\item \Cref{prompt:prm_style_example}: \textbf{Training example of \GenPRM} on the law domain of MMLU-pro~\citep{mmlu-pro}.

\end{itemize}

\newpage

\begin{figure}[H]
\begin{tcolorbox}[llmprompt]
\small
\textbf{[user]} \\
You are a \{\texttt{category}\} teacher. Grade the solution, verifying correctness step by step. \newline
At the end of Solution verification, when you give your final grade, write it in the form ``Verification: Is the answer correct (Yes/No)? X'', where X is either Yes or No.\newline

[Law Problem] \\
Question: A shopper in a store sees that a display case is about to fall on a second shopper's head. The shopper does nothing to warn or assist the second shopper, who is a stranger to her. She could have taken action to protect the second shopper without harming herself. Did the shopper have a duty to take action to try and protect the second shopper from injury?

A. No, the shopper is not required to intervene unless she is a store employee or security personnel.

B. Yes, the shopper must take action under the \"Good Samaritan\" law.

C. Yes, the fact that a person sees that action is needed for another's aid or protection does impose a legal duty to act.
\texttt{[OMITTED...]} \\

[Solution] \\

The shopper could have taken action to protect the second shopper without harming herself, so the issue is whether there is a legal duty to intervene. \\

The ``Good Samaritan'' law protects individuals from liability for failing to provide assistance, so B is incorrect. \\

There is no special relationship between the shopper and the second shopper, so I is incorrect. \\

\texttt{[OMITTED...]} \\

Therefore, the shopper is not required to intervene, and the correct answer is D. \\

The answer is (D).

\textbf{[/user]} \textbf{[assistant]} \textbf{[think]} \\

Let's verify step by step: 

The question is about whether a shopper has a duty to act to prevent harm to another stranger when she can do so without risk. The solution concludes D is correct. \\

First, the solution correctly dismisses B because Good Samaritan laws typically protect those who do help, not impose a duty to act. That seems right. \\

For option I, the solution says no special relationship exists. Since the shopper and victim are strangers, that's accurate. So I is wrong. \\

\texttt{[OMITTED...]} \\

The solution mentions ``duty to rescue'' in certain circumstances like special relationships or being first to notice. However, in the problem, there's no indication the shopper was first. So D holds.\\

Wait, but the problem states ``she could have taken action without harming herself,'' which is a key point. Some jurisdictions might have a duty if it's safe, but traditionally, common law doesn't. The solution's conclusion aligns with that. \\

Therefore, the answer D is correct. The solution's reasoning is sound, dismissing other options properly. The final answer D matches the analysis. So the solution is correct. \\

\textbf{[/think]} \\

Verification: Is the answer correct (Yes/No)? Yes
\textbf{[/assistant]}
\end{tcolorbox}
\vspace{-0.2in}
\caption{\textbf{Training example of \GenORM} on the law domain of MMLU-Pro~\citep{mmlu-pro}.}
\label{prompt:orm_style_example}
\end{figure}
\begin{figure}[H]
\vspace{-0.1in}
\begin{tcolorbox}[llmprompt]
\small
\textbf{[user]}
You are given a law problem and a proposed step-by-step solution:\\

[Law Problem] \\
Question: A plaintiff sued a defendant for injuries that the plaintiff suffered when he was struck by the defendant's car. At trial, the plaintiff testified that he was walking across the street inside a crosswalk at the time the defendant's car hit him. \texttt{[OMITTED...]} In support of the defendant's testimony, the defendant calls a neighbor to testify that she has observed the plaintiff frequently during the years that she has lived on that street, and that the plaintiff "always crosses the street outside the crosswalk." The neighbor's testimony is likely to be \\
A. admitted, because it tends to show that the plaintiff was careless. \\
\texttt{[OMITTED...]} \\
G. admitted, because it is probative of the plaintiff's habit of crossing the street outside the crosswalk. \\
H. excluded, because it is irrelevant to the specific incident. \\
I. admitted, because it provides context to the plaintiff's behavior. \\

[Solution] \\
Step 1: The neighbor's testimony is about the plaintiff's habit of crossing the street outside the crosswalk, which is relevant to the case. \\
Step 2: Habit evidence is admissible to prove that a person regularly acts in conformity with the habit. \\
Step 3: The neighbor's testimony is an example of habit evidence, as it describes the plaintiff's regular behavior of crossing the street outside the crosswalk. \\
Step 4: The answer is (G). \\

Review and critique each step in the proposed solution to determine whether each step is correct. If the solution is incomplete, only verify the provided steps. \textbf{[/user]} \textbf{[assistant]} \textbf{[think]} \\
Let's verify step by step: \\

First, the problem is about whether the neighbor's testimony about the plaintiff's habit is admissible. The proposed solution says step 1 is correct because the testimony is about habit and relevant. That seems right since the case hinges on where the plaintiff was crossing. \\

Step 2 states habit evidence is admissible to show conformity. Under Federal Rules of Evidence, habit (as opposed to character) is admissible. So step 2 is correct. \\

Step 3 says the neighbor's testimony is habit evidence. But wait, habit requires a regular, routine practice. The neighbor says ``frequently'' over years, which might qualify as a habit. So step 3 is correct. \\

Step 4 concludes G is correct. So all steps seem okay. Wait, but maybe ``frequently'' isn't enough? Some jurisdictions require more specific regularity. But generally, habit evidence can be admissible if it's a settled pattern. So steps are correct. Hmm, but let me check again. The options: G says ``probative of plaintiff's habit''. So yes, the steps are correct. The answer should be G. So all steps are correct. \\
\textbf{[/think]} \\

Step 1: The step is \verb|\boxed{correct}| \\
Step 2: The step is \verb|\boxed{correct}| \\
Step 3: The step is \verb|\boxed{correct}| \\
Step 4: The step is \verb|\boxed{correct}| \\
Is the solution correct? Yes \\
\textbf{[/assistant]}
\end{tcolorbox}
\vspace{-0.2in}
\caption{\textbf{Training example of \GenPRM} on the law domain of MMLU-Pro~\citep{mmlu-pro}.}
\label{prompt:prm_style_example}
\end{figure}

\section{Detailed Results on MMLU-Pro}\label{sec:additional_experiments}
\begin{figure}[H]
\includegraphics[height=0.7cm]{images/main_legend.pdf}
\medskip
\vspace{-0.12in}
\centering
\includegraphics[width=0.95\textwidth]{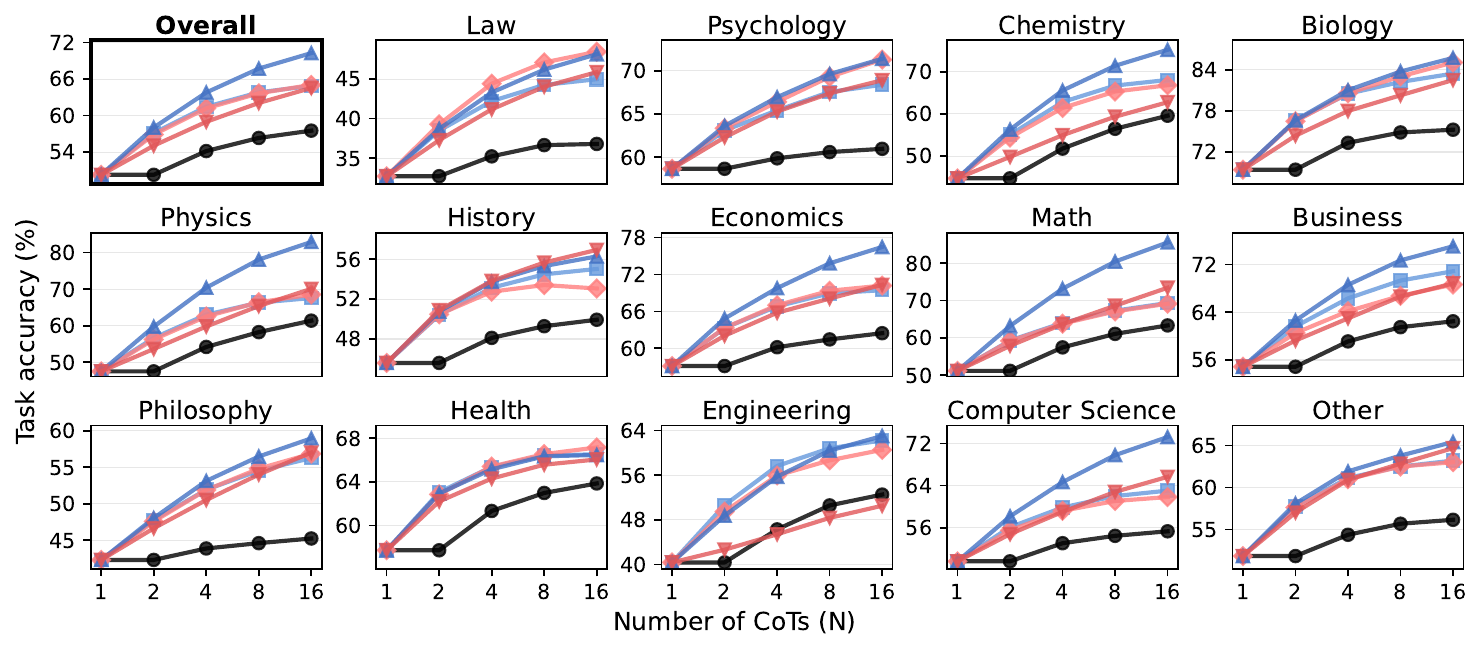}
\vspace{-0.15in}
\caption{\small \textbf{Best-of-$N$ results using \href{https://huggingface.co/meta-llama/Llama-3.1-8B-Instruct}{Llama-3.1-8B-Instruct}} on MMLU-Pro with \href{https://huggingface.co/deepseek-ai/DeepSeek-R1-Distill-Qwen-14B}{R1-Distill-Qwen-14B} backbone for reward models.}
\label{fig:llama_8B_14B_bon}
\end{figure}

\begin{figure}[H]
\vspace{-0.5in}
\centering
\includegraphics[width=0.95\textwidth]{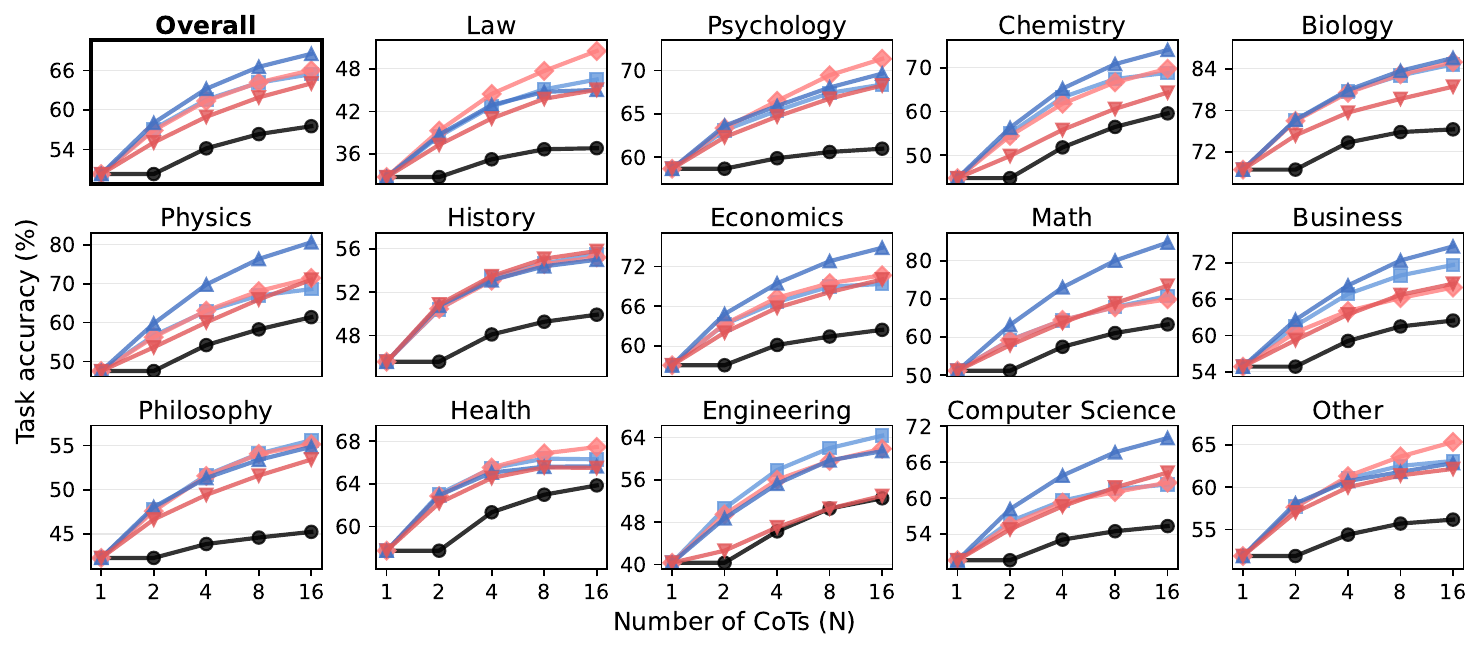}
\vspace{-0.15in}
\caption{\small \textbf{Weighted majority voting results using \href{https://huggingface.co/meta-llama/Llama-3.1-8B-Instruct}{Llama-3.1-8B-Instruct}} on MMLU-Pro with \href{https://huggingface.co/deepseek-ai/DeepSeek-R1-Distill-Qwen-14B}{R1-Distill-Qwen-14B} backbone for reward models.}
\label{fig:llama_8B_14B_wmv}
\vspace{-0.1in}
\end{figure}

\begin{figure}[H]
\vspace{-0.3in}
\centering
\includegraphics[width=0.95\textwidth]{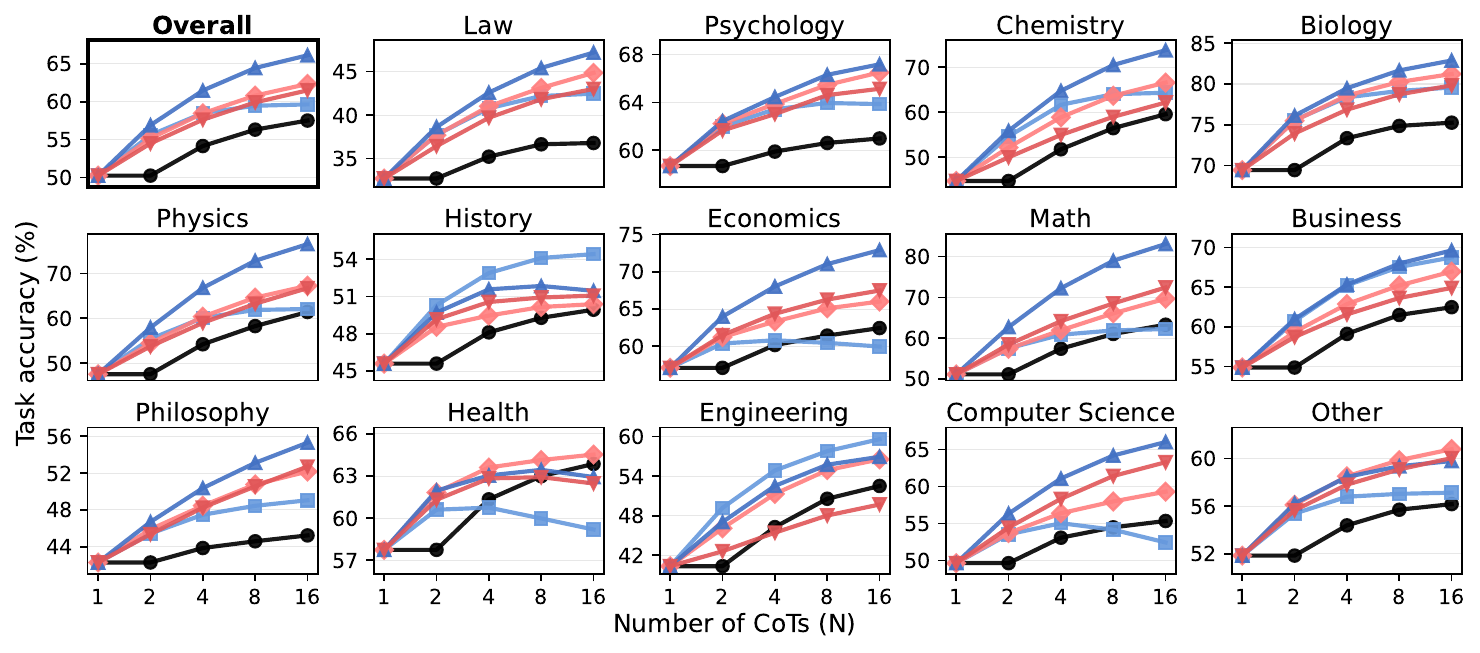}
\vspace{-0.15in}
\caption{\small \textbf{Best-of-$N$ results using \href{https://huggingface.co/meta-llama/Llama-3.1-8B-Instruct}{Llama-3.1-8B-Instruct}} on MMLU-Pro~\citep{mmlu-pro} with \href{https://huggingface.co/deepseek-ai/DeepSeek-R1-Distill-Llama-8B}{R1-distill-Llama-8B} backbone for reward models.}
\label{fig:llama_8B_8B_bon}
\vspace{-0.1in}
\end{figure}

\begin{figure}[H]
\vspace{-0.3in}
\centering
\includegraphics[width=0.95\textwidth]{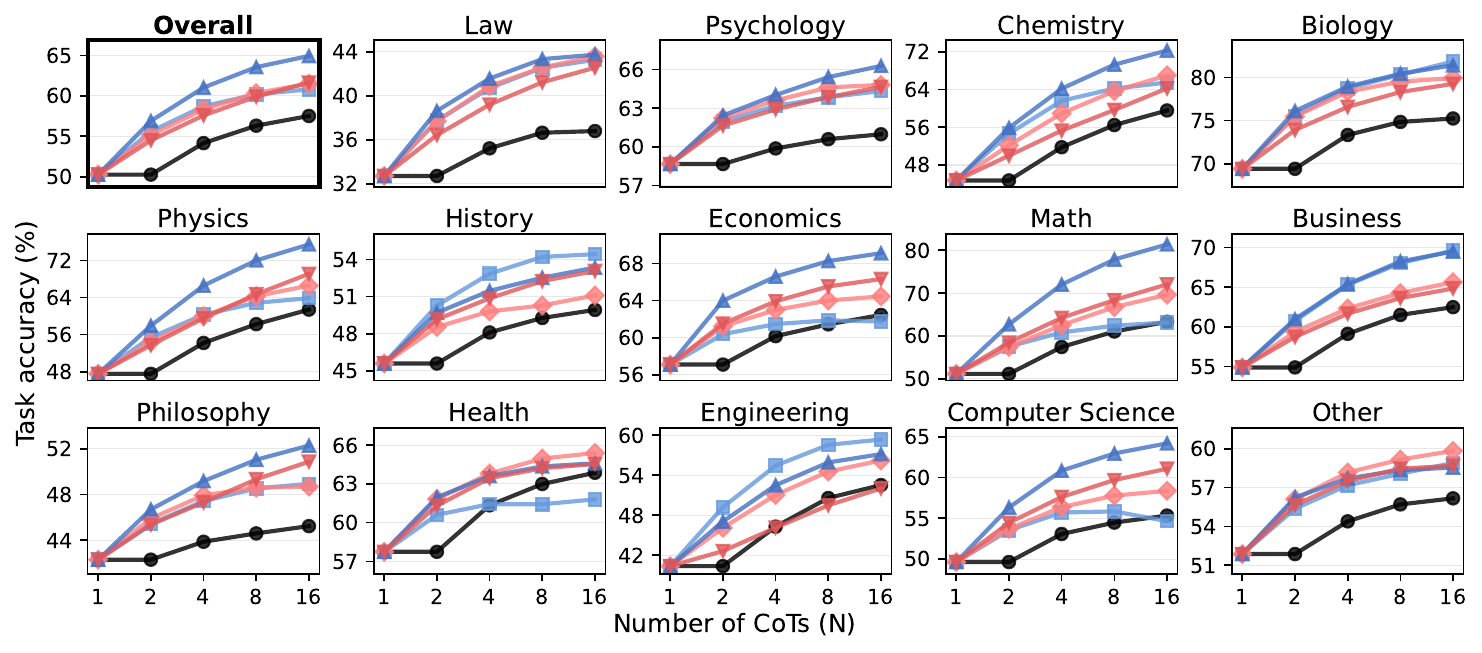}
\vspace{-0.15in}
\caption{\small \textbf{Weighted majority voting results using \href{https://huggingface.co/meta-llama/Llama-3.1-8B-Instruct}{Llama-3.1-8B-Instruct}} on MMLU-Pro~\citep{mmlu-pro} with \href{https://huggingface.co/deepseek-ai/DeepSeek-R1-Distill-Llama-8B}{R1-Distill-Llama-8B} backbone for reward models.}
\label{fig:llama_8B_8B_wmv}
\vspace{-0.1in}
\end{figure}

\begin{figure}[H]
\vspace{-0.5in}
\centering
\includegraphics[width=0.95\textwidth]{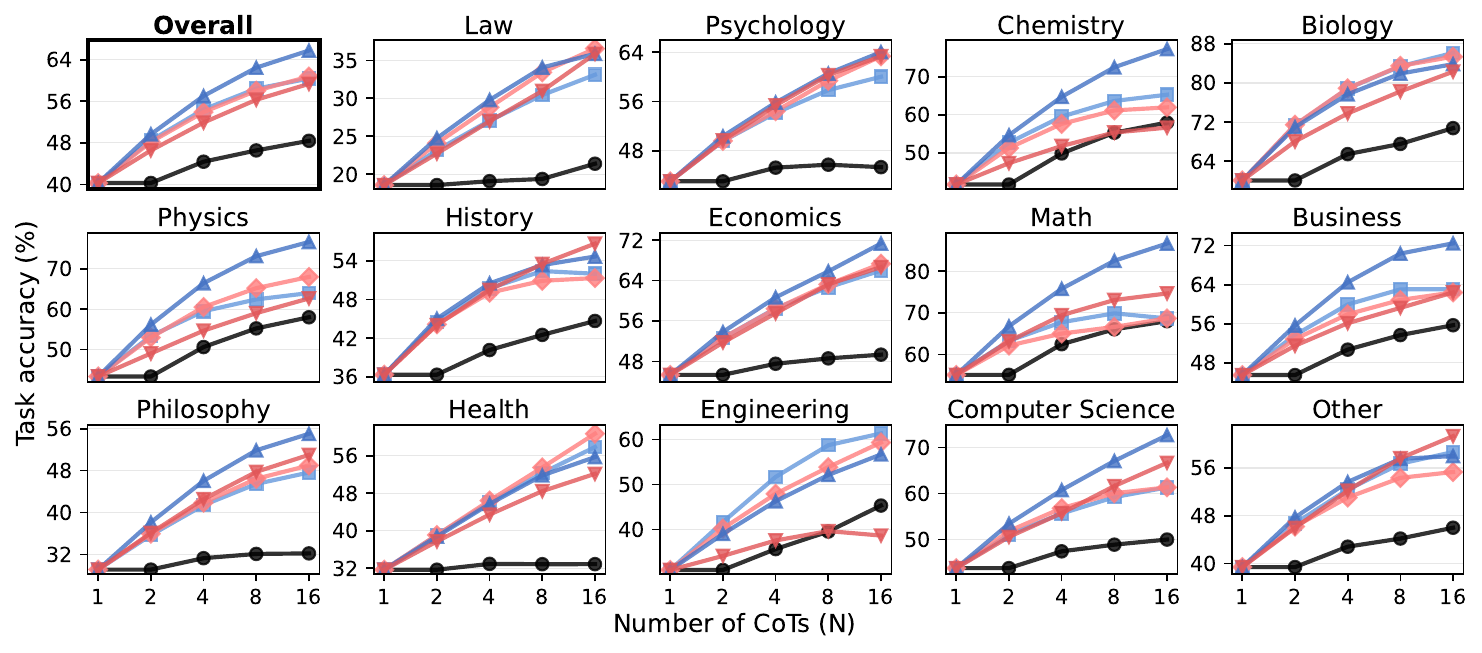}
\vspace{-0.15in}
\caption{\small \textbf{Best-of-$N$ results using \href{https://huggingface.co/HuggingFaceTB/SmolLM3-3B}{SmolLM3-3B}} on MMLU-Pro with \href{https://huggingface.co/deepseek-ai/DeepSeek-R1-Distill-Qwen-14B}{R1-Distill-Qwen-14B} backbone for reward models.}
\label{fig:smollm_bon}
\vspace{-0.1in}
\end{figure}

\begin{figure}[H]
\vspace{-0.3in}
\centering
\includegraphics[width=0.95\textwidth]{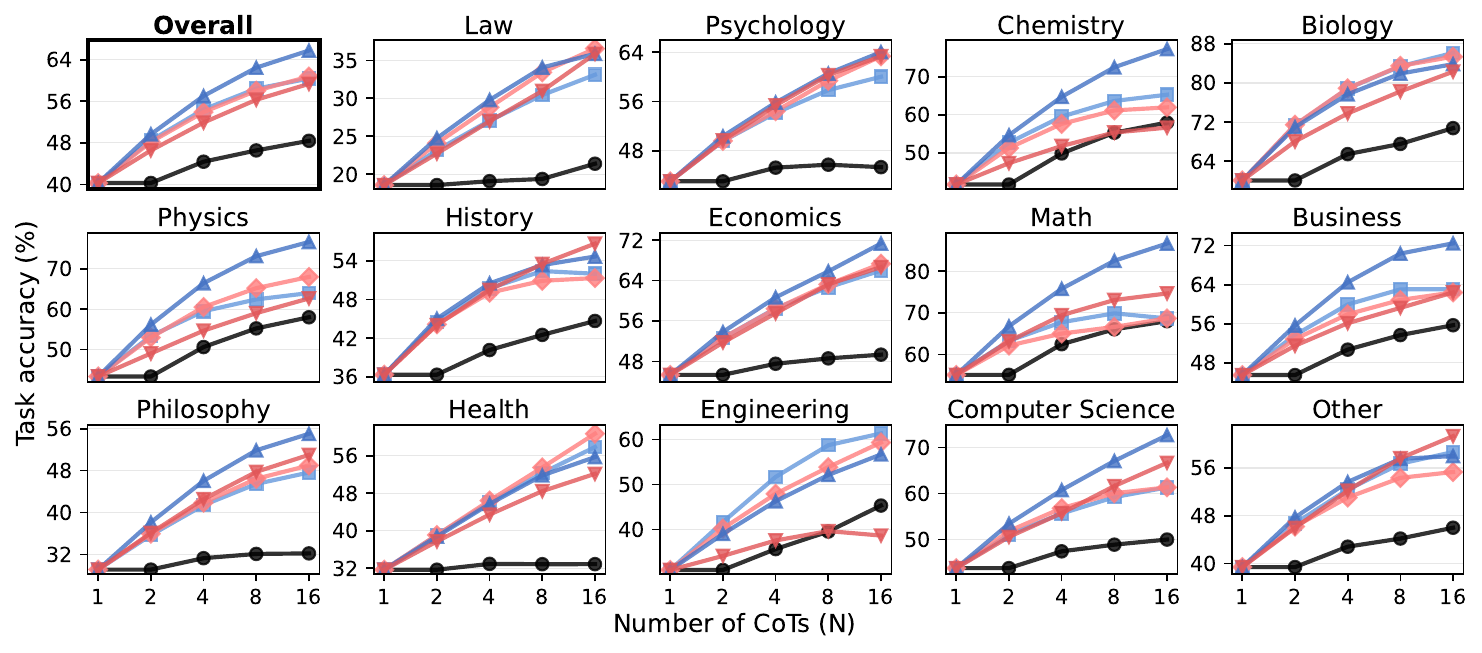}
\vspace{-0.15in}
\caption{\small \textbf{Weighted majority voting results using \href{https://huggingface.co/HuggingFaceTB/SmolLM3-3B}{SmolLM3-3B}} on MMLU-Pro with \href{https://huggingface.co/deepseek-ai/DeepSeek-R1-Distill-Qwen-14B}{R1-Distill-Qwen-14B} backbone for reward models.}
\label{fig:smollm_wmv}
\vspace{-0.1in}
\end{figure}

\begin{figure}[H]
\vspace{-0.3in}
\centering
\includegraphics[width=0.95\textwidth]{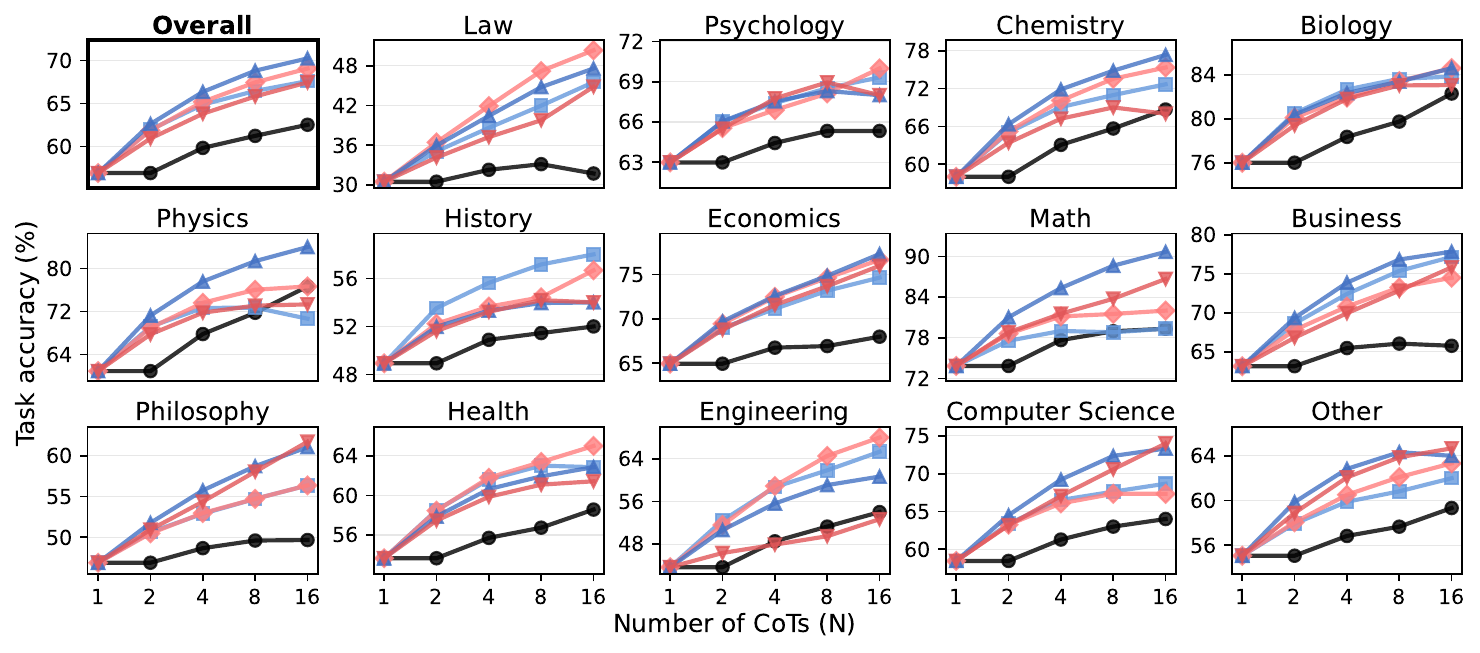}
\vspace{-0.15in}
\caption{\small \textbf{Best-of-$N$ results using \href{https://huggingface.co/Qwen/Qwen2.5-7B-Instruct}{Qwen2.5-7B-Instruct}} on MMLU-Pro with \href{https://huggingface.co/deepseek-ai/DeepSeek-R1-Distill-Qwen-14B}{R1-Distill-Qwen-14B} backbone for reward models.}
\label{fig:qwen_bon}
\vspace{-0.1in}
\end{figure}

\begin{figure}[H]
\vspace{-0.5in}
\centering
\includegraphics[width=0.95\textwidth]{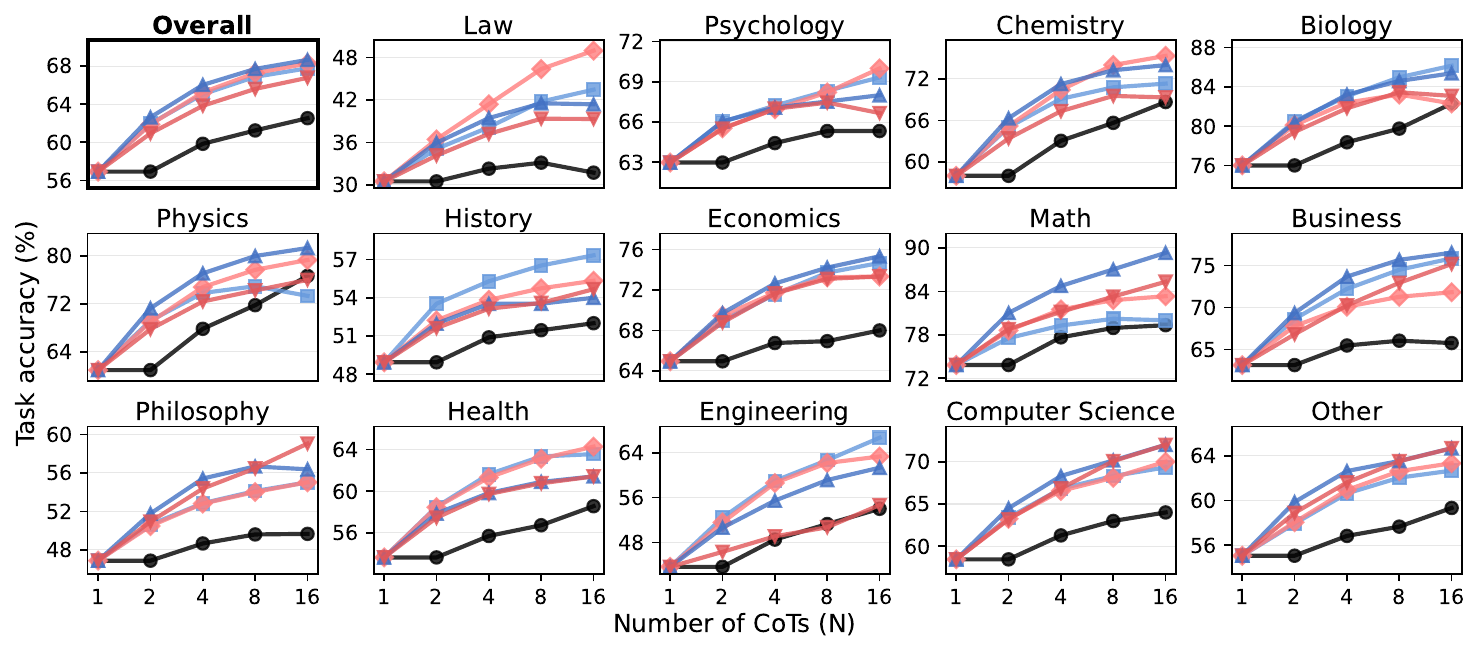}
\vspace{-0.15in}
\caption{\small \textbf{Weighted majority voting results using \href{https://huggingface.co/Qwen/Qwen2.5-7B-Instruct}{Qwen2.5-7B-Instruct}} on MMLU-Pro with \href{https://huggingface.co/deepseek-ai/DeepSeek-R1-Distill-Qwen-14B}{R1-Distill-Qwen-14B} backbone for reward models.}
\label{fig:qwen_wmv}
\vspace{-0.1in}
\end{figure}

\begin{figure}[H]
\vspace{-0.3in}
\centering
\includegraphics[width=0.95\textwidth]{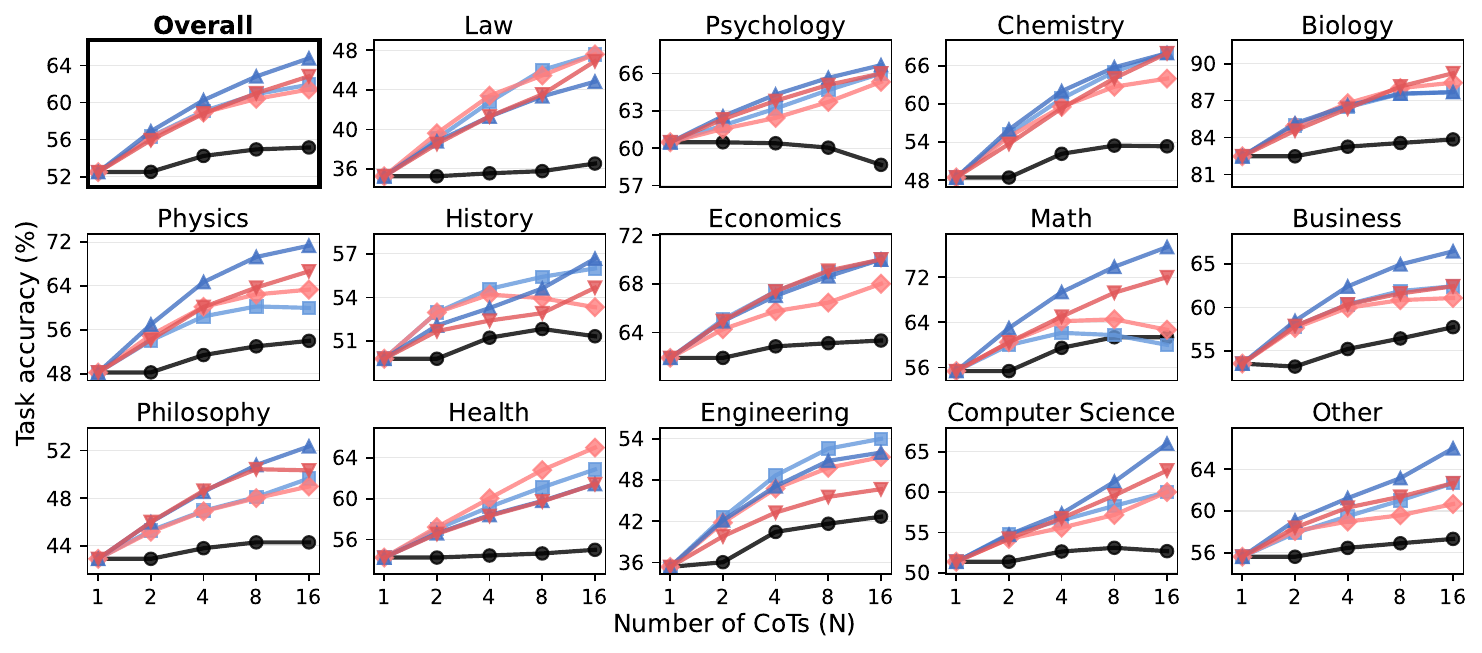}
\vspace{-0.15in}
\caption{\small \textbf{Best-of-$N$ results using \href{https://huggingface.co/google/gemma-2-9b-it}{gemma-2-9b-it}} on MMLU-Pro with \href{https://huggingface.co/deepseek-ai/DeepSeek-R1-Distill-Qwen-14B}{R1-Distill-Qwen-14B} backbone for reward models.}
\label{fig:gemma_bon}
\vspace{-0.1in}
\end{figure}

\begin{figure}[H]
\vspace{-0.3in}
\centering
\includegraphics[width=0.95\textwidth]{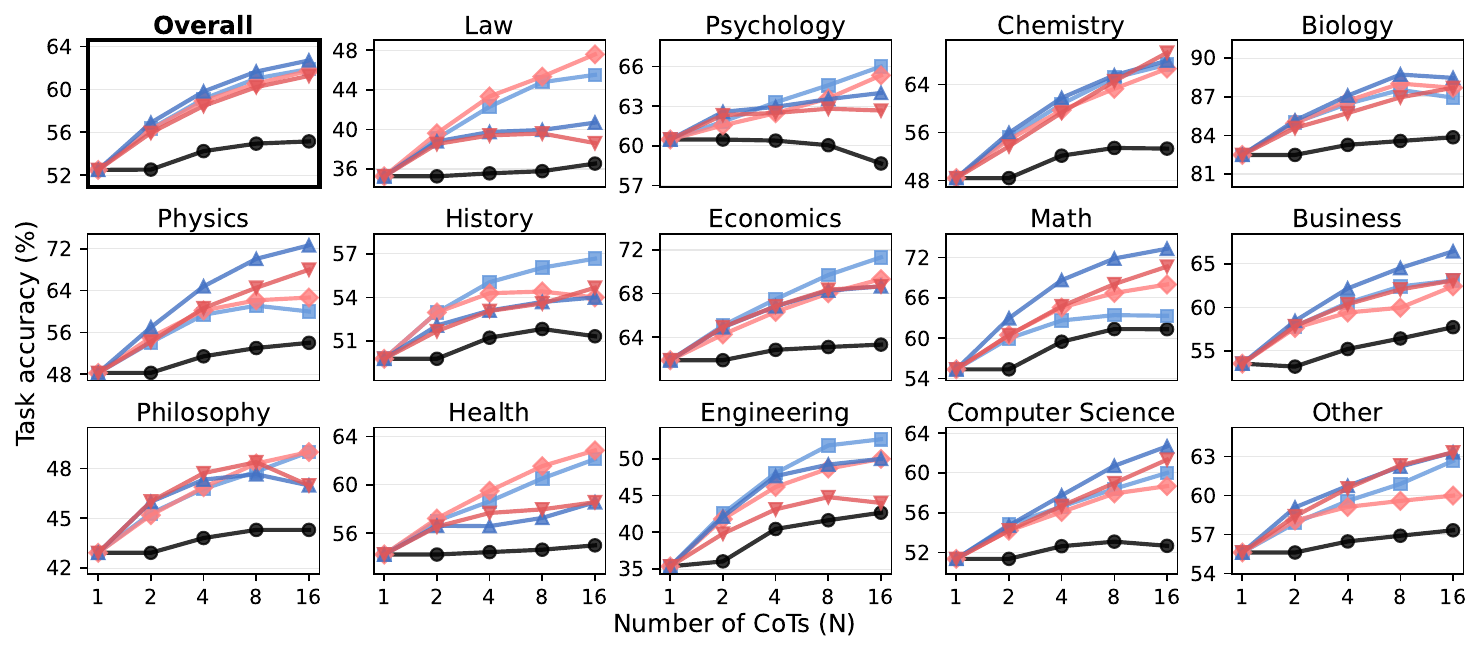}
\vspace{-0.15in}
\caption{\small \textbf{Weighted majority voting results using \href{https://huggingface.co/google/gemma-2-9b-it}{gemma-2-9b-it}} on MMLU-Pro with \href{https://huggingface.co/deepseek-ai/DeepSeek-R1-Distill-Qwen-14B}{R1-Distill-Qwen-14B} backbone for reward models.}
\label{fig:gemma_wmv}
\vspace{-0.1in}
\end{figure}

\begin{figure}[H]
\vspace{-0.5in}
\centering
\includegraphics[width=0.95\textwidth]{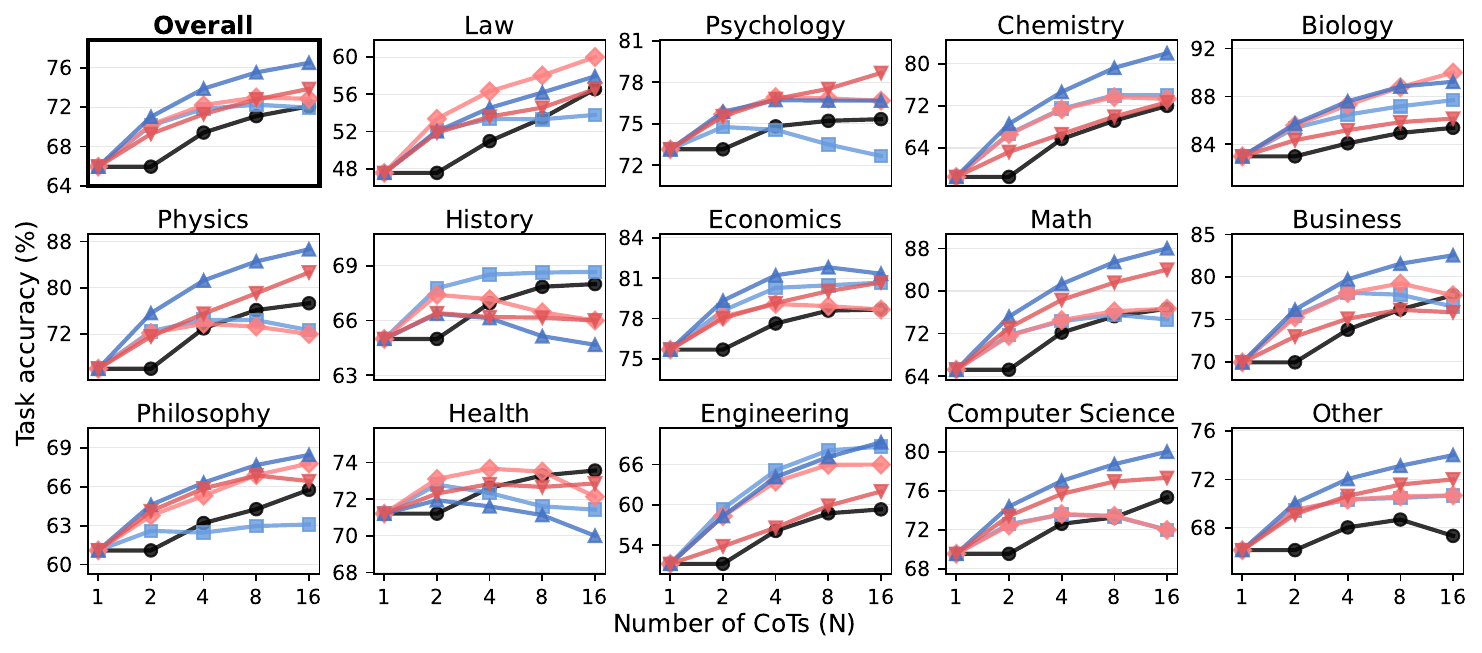}
\vspace{-0.15in}
\caption{\small \textbf{Best-of-$N$ results using \href{https://huggingface.co/meta-llama/Llama-3.1-70B-Instruct}{Llama-3.1-70B-Instruct}} on MMLU-Pro with \href{https://huggingface.co/deepseek-ai/DeepSeek-R1-Distill-Qwen-14B}{R1-Distill-Qwen-14B} backbone for reward models.}
\label{fig:llama_70B_bon}
\vspace{-0.1in}
\end{figure}

\begin{figure}[H]
\vspace{-0.3in}
\centering
\includegraphics[width=0.95\textwidth]{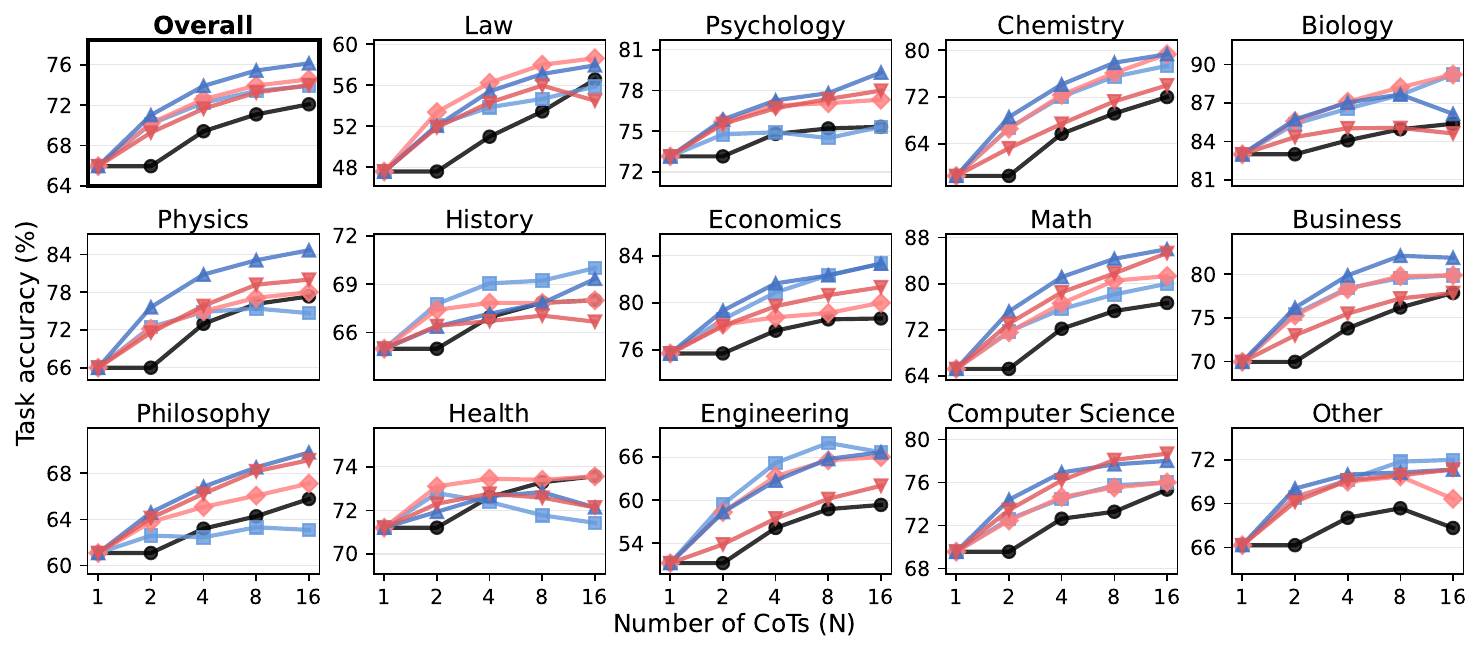}
\vspace{-0.15in}
\caption{\small \textbf{Weighted majority voting results using \href{https://huggingface.co/meta-llama/Llama-3.1-70B-Instruct}{Llama-3.1-70B-Instruct}} on MMLU-Pro with \href{https://huggingface.co/deepseek-ai/DeepSeek-R1-Distill-Qwen-14B}{R1-Distill-Qwen-14B} backbone for reward models.}
\label{fig:llama_70B_wmv}
\vspace{-0.1in}
\end{figure}

\begin{figure}[H]
\vspace{-0.3in}
\centering
\includegraphics[width=0.95\textwidth]{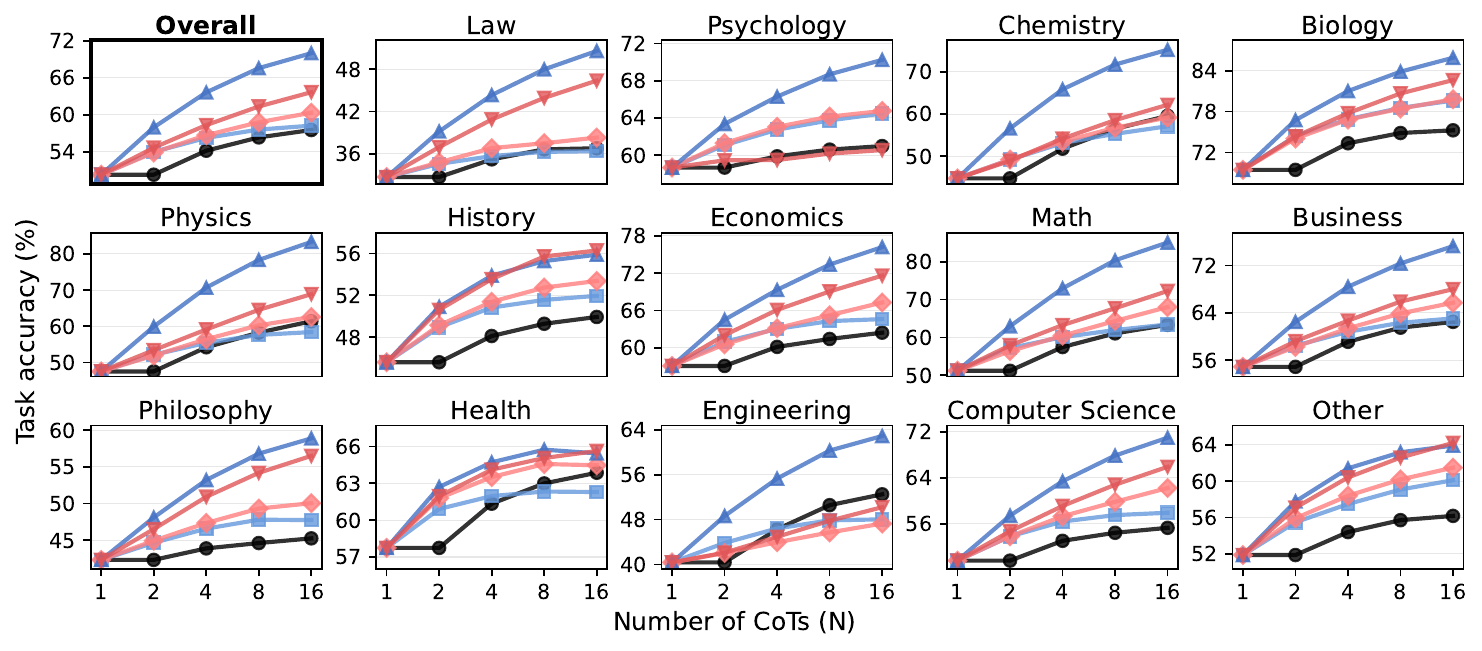}
\vspace{-0.15in}
\caption{\small \textbf{Best-of-$N$ performance using \href{https://huggingface.co/meta-llama/Llama-3.1-8B-Instruct}{Llama-3.1-8B-Instruct} when trained and evaluated on each domain} of MMLU-Pro with \href{https://huggingface.co/deepseek-ai/DeepSeek-R1-Distill-Qwen-14B}{R1-Distill-Qwen-14B} backbone for reward models.}
\label{fig:domain_specialization_bon}
\vspace{-0.1in}
\end{figure}

\begin{figure}[H]
\vspace{-0.5in}
\centering
\includegraphics[width=0.95\textwidth]{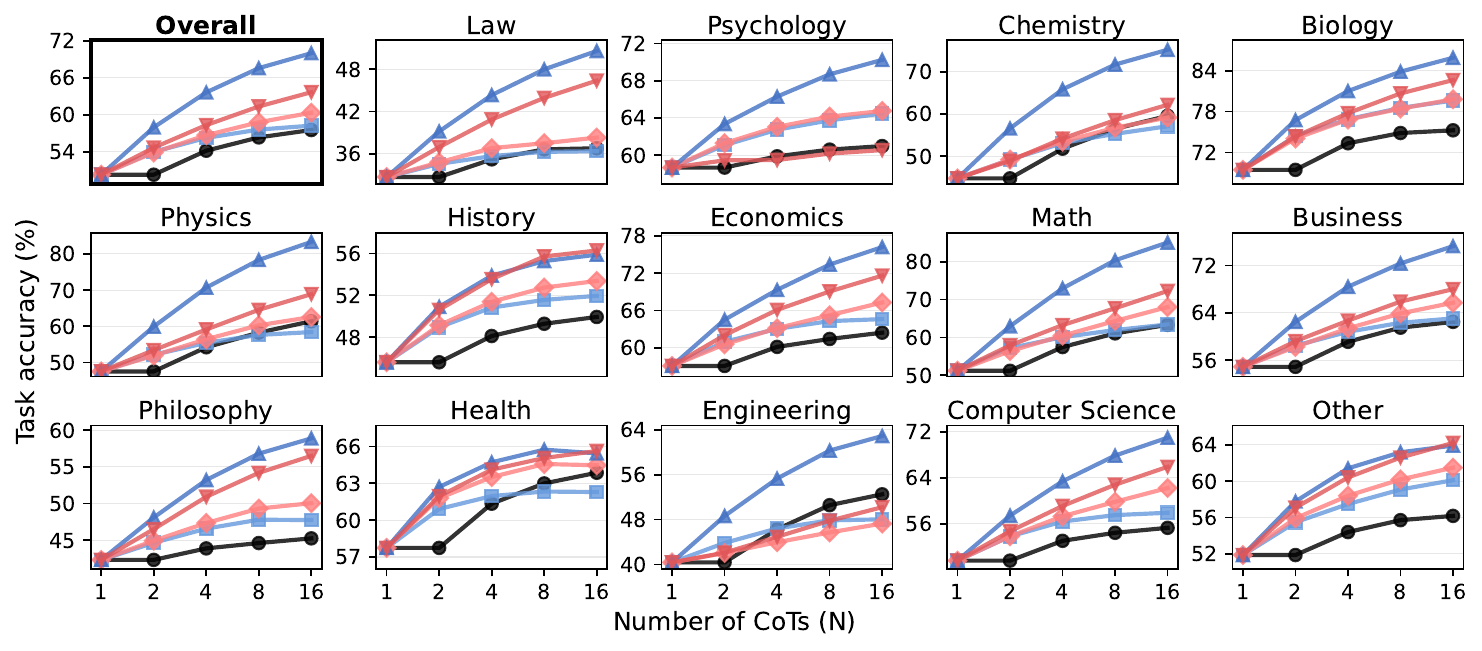}
\vspace{-0.15in}
\caption{\small \textbf{Weighted majority voting performance using \href{https://huggingface.co/meta-llama/Llama-3.1-8B-Instruct}{Llama-3.1-8B-Instruct} when trained and evaluated on each domain} of MMLU-Pro with \href{https://huggingface.co/deepseek-ai/DeepSeek-R1-Distill-Qwen-14B}{R1-Distill-Qwen-14B} backbone for reward models.}
\label{fig:domain_specialization_wmv}
\vspace{-0.1in}
\end{figure}

In this section, we present detailed results on MMLU-Pro~\citep{mmlu-pro} as follows:
\begin{itemize}[itemsep=1mm,parsep=1pt,topsep=2pt,leftmargin=*]

\item \Cref{fig:llama_8B_14B_bon}: \textbf{Best-of-$N$ results using \href{https://huggingface.co/meta-llama/Llama-3.1-8B-Instruct}{Llama-3.1-8B-Instruct}} on MMLU-Pro with \href{https://huggingface.co/deepseek-ai/DeepSeek-R1-Distill-Qwen-14B}{R1-Distill-Qwen-14B} backbone for reward models.

\item \Cref{fig:llama_8B_14B_wmv}: \textbf{Weighted majority voting results using \href{https://huggingface.co/meta-llama/Llama-3.1-8B-Instruct}{Llama-3.1-8B-Instruct}} on MMLU-Pro with \href{https://huggingface.co/deepseek-ai/DeepSeek-R1-Distill-Qwen-14B}{R1-Distill-Qwen-14B} backbone for reward models.

\item \Cref{fig:llama_8B_8B_bon}: \textbf{Best-of-$N$ results using \href{https://huggingface.co/meta-llama/Llama-3.1-8B-Instruct}{Llama-3.1-8B-Instruct}} on MMLU-Pro with \href{https://huggingface.co/deepseek-ai/DeepSeek-R1-Distill-Llama-8B}{R1-Distill-Llama-8B} backbone for reward models.

\item \Cref{fig:llama_8B_8B_wmv}: \textbf{Weighted majority voting results using \href{https://huggingface.co/meta-llama/Llama-3.1-8B-Instruct}{Llama-3.1-8B-Instruct}} on MMLU-Pro with \href{https://huggingface.co/deepseek-ai/DeepSeek-R1-Distill-Llama-8B}{R1-Distill-Llama-8B} backbone for reward models.

\item \Cref{fig:smollm_bon}: \textbf{Best-of-$N$ results using \href{https://huggingface.co/HuggingFaceTB/SmolLM3-3B}{SmolLM3-3B}} on MMLU-Pro with \href{https://huggingface.co/deepseek-ai/DeepSeek-R1-Distill-Qwen-14B}{R1-Distill-Qwen-14B} backbone for reward models.

\item \Cref{fig:smollm_wmv}: \textbf{Weighted majority voting results using \href{https://huggingface.co/HuggingFaceTB/SmolLM3-3B}{SmolLM3-3B}} on MMLU-Pro with \href{https://huggingface.co/deepseek-ai/DeepSeek-R1-Distill-Qwen-14B}{R1-Distill-Qwen-14B} backbone for reward models.

\item \Cref{fig:qwen_bon}: \textbf{Best-of-$N$ results using \href{https://huggingface.co/Qwen/Qwen2.5-7B-Instruct}{Qwen2.5-7B-Instruct}} on MMLU-Pro with \href{https://huggingface.co/deepseek-ai/DeepSeek-R1-Distill-Qwen-14B}{R1-Distill-Qwen-14B} backbone for reward models.

\item \Cref{fig:qwen_wmv}: \textbf{Weighted majority voting results using \href{https://huggingface.co/Qwen/Qwen2.5-7B-Instruct}{Qwen2.5-7B-Instruct}} on MMLU-Pro with \href{https://huggingface.co/deepseek-ai/DeepSeek-R1-Distill-Qwen-14B}{R1-Distill-Qwen-14B} backbone for reward models.

\item \Cref{fig:gemma_bon}: \textbf{Best-of-$N$ results using \href{https://huggingface.co/google/gemma-2-9b-it}{gemma-2-9b-it}} on MMLU-Pro with \href{https://huggingface.co/deepseek-ai/DeepSeek-R1-Distill-Qwen-14B}{R1-Distill-Qwen-14B} backbone for reward models.

\item \Cref{fig:gemma_wmv}: \textbf{Weighted majority voting results using \href{https://huggingface.co/google/gemma-2-9b-it}{gemma-2-9b-it}} on MMLU-Pro with \href{https://huggingface.co/deepseek-ai/DeepSeek-R1-Distill-Qwen-14B}{R1-Distill-Qwen-14B} backbone for reward models.

\item \Cref{fig:llama_70B_bon}: \textbf{Best-of-$N$ results using \href{https://huggingface.co/meta-llama/Llama-3.1-70B-Instruct}{Llama-3.1-70B-Instruct}} on MMLU-Pro with \href{https://huggingface.co/deepseek-ai/DeepSeek-R1-Distill-Qwen-14B}{R1-Distill-Qwen-14B} backbone for reward models.

\item \Cref{fig:llama_70B_wmv}: \textbf{Weighted majority voting results using \href{https://huggingface.co/meta-llama/Llama-3.1-8B-Instruct}{Llama-3.1-8B-Instruct}} on MMLU-Pro with \href{https://huggingface.co/deepseek-ai/DeepSeek-R1-Distill-Qwen-14B}{R1-Distill-Qwen-14B} backbone for reward models.

\item \Cref{fig:domain_specialization_bon}: \textbf{Best-of-$N$ results using \href{https://huggingface.co/meta-llama/Llama-3.1-8B-Instruct}{Llama-3.1-8B-Instruct} when trained and evaluated on each domain} of MMLU-Pro with \href{https://huggingface.co/deepseek-ai/DeepSeek-R1-Distill-Qwen-14B}{R1-Distill-Qwen-14B} backbone for reward models.

\item \Cref{fig:domain_specialization_wmv}: \textbf{Weighted majority voting results using \href{https://huggingface.co/meta-llama/Llama-3.1-8B-Instruct}{Llama-3.1-8B-Instruct} when trained and evaluated on each domain} of MMLU-Pro with \href{https://huggingface.co/deepseek-ai/DeepSeek-R1-Distill-Qwen-14B}{R1-Distill-Qwen-14B} backbone for reward models.

\vspace{-0.05in}
\end{itemize}


\section{Additional Analysis}\label{sec:additional_analysis}
In this section, we present additional analysis on the failure of PRMs.

\begin{figure}[H]
\vspace{-0.3in}
\includegraphics[height=0.7cm]{images/main_legend.pdf}
\medskip
\vspace{-0.12in}
\centering
\includegraphics[width=1\textwidth]{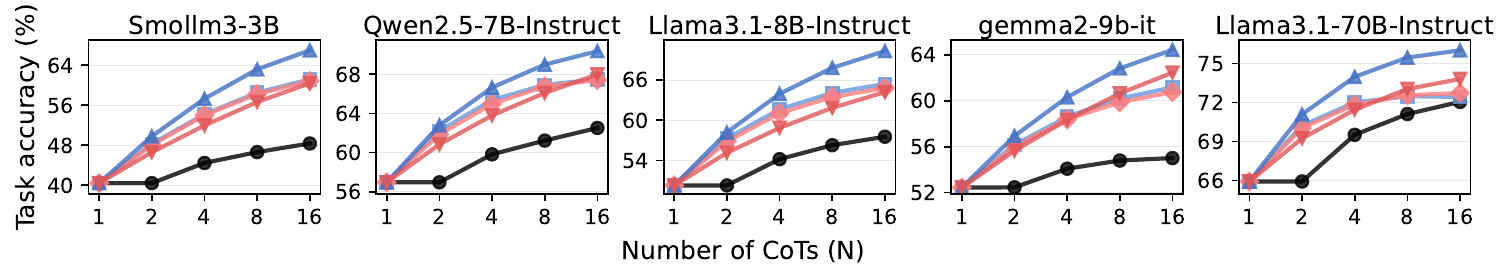}
\vspace{-0.3in}
\caption{\small Overall Best-of-$N$ results using five different $p_\mathtt{LLM}$ on MMLU-Pro with \textbf{\href{https://huggingface.co/Qwen/Qwen3-8B}{Qwen3-8B}} backbone for reward models.}
\label{fig:overall_bon_qwen}
\vspace{-0.1in}
\end{figure}

\begin{itemize}[itemsep=1mm,parsep=1pt,topsep=2pt,leftmargin=*]

\item \Cref{fig:overall_bon_qwen}: Overall Best-of-$N$ results using five different $p_\mathtt{LLM}$ on MMLU-Pro with \textbf{\href{https://huggingface.co/Qwen/Qwen3-8B}{Qwen3-8B}} backbone for reward models.
\item \Cref{tab:hyperparameter_search}:
Best-of-$N$ results (overall) on MMLU-Pro using \href{https://huggingface.co/meta-llama/Llama-3.1-8B-Instruct}{Llama-3.1-8B-Instruct} by \textbf{varying learning rate and LoRA rank $r$ for PRM variants}. We use \href{https://huggingface.co/deepseek-ai/DeepSeek-R1-Distill-Qwen-14B}{R1-Distill-Qwen-14B} backbone for reward models.


\item \Cref{tab:wasserstein_distance_math}: \textbf{Wasserstein distance in the math domain} before and after filtering for \GenORM and \GenPRM.

\item \Cref{tab:wasserstein_distance}: \textbf{Wasserstein distance in the multi-domain setting} before and after filtering for \GenORM and \GenPRM. To reduce the CoT-length distribution shift (\ie, the Wasserstein distance) for \GenPRM, we apply (i) \textbf{label refinement} using \href{https://blog.google/technology/google-deepmind/google-gemini-ai-update-december-2024/}{Gemini-2.0 Flash}~\citep{comanici2025gemini}: due to a parsing issue, \textbf{59.96\%} of process labels are replaced; and (ii) \textbf{relaxation} of the \textit{consensus filtering} rule: when $y=1$, we keep the verification CoTs $v_{1:L^+}$ with $\hat{z}_{1:T} = 1_T$, and when $y=0$, we keep $v_{1:L^+}$ if there exists $t \in \{1,\ldots,T\}$ such that $z_t = 0$.

\item \Cref{tab:survival_proportion}:
\textbf{Surviving proportion (\%) of CoTs} on the train split of MMLU-Pro. 
We compare \GenORM and \GenPRM under (i) \textbf{label refinement} using \href{https://blog.google/technology/google-deepmind/google-gemini-ai-update-december-2024/}{Gemini-2.0 Flash}~\citep{comanici2025gemini}, and (ii) \textbf{relaxed} \textit{consensus filtering}.

\item \Cref{tab:label_refinement_and_relaxed_filtering}:
Best-of-$N$ results on MMLU-Pro using \href{https://huggingface.co/meta-llama/Llama-3.1-8B-Instruct}{Llama-3.1-8B-Instruct} with (i) \textbf{label refinement} or (ii) \textbf{relaxed filtering}. We use \href{https://huggingface.co/deepseek-ai/DeepSeek-R1-Distill-Qwen-14B}{R1-Distill-Qwen-14B} as the backbone for reward models. 

\item \Cref{tab:mmlu_pro_aggregation_smollm3_3b}:
Best-of-$N$ results on MMLU-Pro using CoTs generated by \href{https://huggingface.co/HuggingFaceTB/SmolLM3-3B}{SmolLM3-3B}. We use \href{https://huggingface.co/deepseek-ai/DeepSeek-R1-Distill-Qwen-14B}{R1-Distill-Qwen-14B} as the backbone for reward models.

\item \Cref{tab:mmlu_pro_aggregation_qwen2_5_7b}:
Best-of-$N$ results on MMLU-Pro using CoTs generated by \href{https://huggingface.co/Qwen/Qwen2.5-7B-Instruct}{Qwen2.5-7B-Instruct}. We use \href{https://huggingface.co/deepseek-ai/DeepSeek-R1-Distill-Qwen-14B}{R1-Distill-Qwen-14B} as the backbone for reward models.

\item \Cref{tab:mmlu_pro_aggregation_llama3_1_8b}:
Best-of-$N$ results on MMLU-Pro using CoTs generated by \href{https://huggingface.co/meta-llama/Llama-3.1-8B-Instruct}{Llama-3.1-8B-Instruct}. We use \href{https://huggingface.co/deepseek-ai/DeepSeek-R1-Distill-Qwen-14B}{R1-Distill-Qwen-14B} as the backbone for reward models.

\item \Cref{tab:mmlu_pro_aggregation_gemma2_9b}:
Best-of-$N$ results on MMLU-Pro using CoTs generated by \href{https://huggingface.co/google/gemma-2-9b-it}{gemma2-9b-it}. We use \href{https://huggingface.co/deepseek-ai/DeepSeek-R1-Distill-Qwen-14B}{R1-Distill-Qwen-14B} as the backbone for reward models.

\item \Cref{tab:mmlu_pro_aggregation_llama3_1_70b}:
Best-of-$N$ results on MMLU-Pro using CoTs generated by \href{https://huggingface.co/meta-llama/Llama-3.1-70B-Instruct}{Llama-3.1-70B-Instruct}. We use \href{https://huggingface.co/deepseek-ai/DeepSeek-R1-Distill-Qwen-14B}{R1-Distill-Qwen-14B} as the backbone for reward models.

\end{itemize}

\begin{table}[H]
\vspace{-0.1in}
\centering
\caption{Best-of-$N$ results (overall) on MMLU-Pro using \href{https://huggingface.co/meta-llama/Llama-3.1-8B-Instruct}{Llama-3.1-8B-Instruct} by \textbf{varying learning rate and LoRA rank $r$ for PRM variants}. We use with \href{https://huggingface.co/deepseek-ai/DeepSeek-R1-Distill-Qwen-14B}{R1-Distill-Qwen-14B} backbone for reward models. The number in parentheses denotes the change.}
\vspace{-0.1in}
\label{tab:hyperparameter_search}
\small
\resizebox{\textwidth}{!}{
\begin{tabular}{lcclllll}
\toprule
\multirow{2}{*}{\textbf{Method}} & \multirow{2}{*}{\textbf{Learning rate}} & \multirow{2}{*}{$r$} & \multicolumn{5}{c}{$N$} \\
\cmidrule(lr){4-8}
 &  &  & \textbf{1} & \textbf{2} & \textbf{4} & \textbf{8} & \textbf{16} \\
\midrule
\textbf{Majority voting} & -- & -- & 50.27 & 50.27 & 54.15 & 56.14 & 57.16 \\
\midrule
\textbf{\DisORM} & $1\cdot 10^{-4}$ & 16 & 50.27 & 57.30 & 61.54 & 63.95 & 65.38 \\
\textbf{\DisPRM (default)} & $1\cdot 10^{-4}$ & 16 & 50.27 & 57.18 & 61.51 & 64.10 & 65.55 \\
\textbf{\DisPRM (changed)} & $5\cdot 10^{-5}$ & 16 & 50.27 & 56.77 \textbf{(-0.41)} & 60.84 \textbf{(-0.67)} & 62.99 \textbf{(-1.11)} & 64.04 \textbf{(-1.51)} \\
\textbf{\DisPRM (changed)} & $1\cdot 10^{-4}$ & 32 & 50.27 & 57.01 \textbf{(-0.17)} & 61.31 \textbf{(-0.20)} & 64.11 \textbf{(+0.01)} & 66.02 \textbf{(+0.47)} \\
\midrule
\textbf{\GenORM} & $1\cdot 10^{-4}$ & 32 & 50.27 & 58.24 & 63.88 & 67.82 & 70.02 \\
\textbf{\GenPRM (default)} & $1\cdot 10^{-4}$ & 32 & 50.27 & 55.24 & 59.06 & 62.10 & 64.26 \\
\textbf{\GenPRM (changed)} & $5\cdot 10^{-5}$ & 32 & 50.27 & 55.09 \textbf{(-0.15)} & 58.94 \textbf{(-0.12)} & 61.96 \textbf{(-0.14)} & 64.26 \textbf{(+0.00)} \\
\textbf{\GenPRM (changed)} & $1\cdot 10^{-4}$ & 64 & 50.27 & 54.94 \textbf{(-0.30)} & 58.73 \textbf{(-0.33)} & 61.88 \textbf{(-0.22)} & 64.55 \textbf{(+0.29)} \\
\midrule
\textbf{Pass@$N$} & -- & -- & 50.27 & 61.74 & 71.56 & 79.77 & 86.05 \\
\bottomrule
\end{tabular}
}
\vspace{-0.1in}
\end{table}

\begin{table}[H]
\centering
\caption{\small\textbf{Wasserstein distance in the math domain} before and after filtering for \GenORM and \GenPRM.
}
\label{tab:wasserstein_distance_math}
\vspace{-0.1in}
\resizebox{0.7\textwidth}{!}{%
\begin{tabular}{@{}lccccc@{}}
\toprule
 & \textbf{Overall} & \textbf{GSM8K} & \textbf{Math} & \textbf{Omni-Math} & \textbf{OlympiadBench} \\
\midrule
Train (PRM800K) & 2.760 & 5.113 & 3.813 & 2.027 & 1.514 \\
\GenORM            & 2.430 & 4.780 & 3.480 & 1.695 & 1.194 \\
\GenPRM            & 1.600 & 3.680 & 2.348 & 1.448 & 1.203 \\
\bottomrule
\end{tabular}%
}
\vspace{-0.1in}
\end{table}

\begin{table}[H]
\centering
\caption{\textbf{Wasserstein distance in the multi-domain setting} before and after filtering for \GenORM{} and \GenPRM{}. To reduce the CoT-length distribution shift (\ie, the Wasserstein distance) for \GenPRM{}, we apply (i) \textbf{label refinement} using \href{https://blog.google/technology/google-deepmind/google-gemini-ai-update-december-2024/}{Gemini-2.0 Flash}~\citep{comanici2025gemini}: due to a parsing issue, \textbf{59.96\%} of process labels are replaced; and (ii) \textbf{relaxation} of the \textit{consensus filtering} rule: when $y=1$, we keep the verification CoTs $v_{1:L^+}$ with $\hat{z}_{1:T} = 1_T$, and when $y=0$, we keep $v_{1:L^+}$ if there exists $t \in \{1,\ldots,T\}$ such that $z_t = 0$.}
\label{tab:wasserstein_distance}
\vspace{-0.1in}
\small
\resizebox{\textwidth}{!}{%
\begin{tabular}{llllll}
\toprule
 & \textbf{Overall} & \textbf{Law} & \textbf{Psychology} & \textbf{Chemistry} & \textbf{Biology} \\
\midrule
Train               & 0.202 & 0.090 & 0.203 & 0.393 & 0.264 \\
\GenORM             & 0.532 & 0.089 & 0.218 & 1.128 & 0.506 \\
\GenPRM  & 3.083 & 1.284 & 0.742 & 6.922 & 2.039 \\
\GenPRM + \textbf{label refinement} 
 & 3.265 \textbf{(+0.182)}
 & 1.397 \textbf{(+0.113)}
 & 0.893 \textbf{(+0.151)}
 & 7.183 \textbf{(+0.261)}
 & 2.416 \textbf{(+0.377)} \\
\GenPRM + \textbf{relaxed filtering}   & 2.001 \textbf{(-1.082)} & 0.885 \textbf{(-0.399)} & 0.397 \textbf{(-0.345)} & 4.939 \textbf{(-1.983)} & 1.311 \textbf{(-0.728)} \\
\midrule
 & \textbf{Physics} & \textbf{History} & \textbf{Economics} & \textbf{Math} & \textbf{Business} \\
\midrule
Train               & 0.628 & 0.069 & 0.311 & 0.167 & 0.322 \\
\GenORM             & 1.201 & 0.154 & 0.564 & 0.282 & 0.491 \\
\GenPRM  & 5.952 & 0.581 & 1.782 & 4.655 & 4.267 \\
\GenPRM + \textbf{label refinement} 
 & 6.104 \textbf{(+0.152)}
 & 0.752 \textbf{(+0.171)}
 & 2.044 \textbf{(+0.262)}
 & 4.852 \textbf{(+0.197)}
 & 4.494 \textbf{(+0.227)} \\
\GenPRM + \textbf{relaxed filtering}   & 4.371 \textbf{(-1.581)} & 0.203 \textbf{(-0.378)} & 1.094 \textbf{(-0.688)} & 2.571 \textbf{(-2.084)} & 2.777 \textbf{(-1.490)} \\
\midrule
 & \textbf{Philosophy} & \textbf{Health} & \textbf{Engineering} & \textbf{Computer science} & \textbf{Other} \\
\midrule
Train            & 0.129 & 0.105 & 1.234 & 0.353 & 0.093 \\
\GenORM             & 0.545 & 0.213 & 3.611 & 0.338 & 0.312 \\
\GenPRM  & 1.235 & 0.979 & 12.735 & 3.459 & 0.927 \\
\GenPRM + \textbf{label refinement} 
 & 1.299 \textbf{(+0.064)}
 & 1.157 \textbf{(+0.178)}
 & 13.058 \textbf{(+0.323)}
 & 3.742 \textbf{(+0.283)}
 & 1.030 \textbf{(+0.103)} \\
\GenPRM + \textbf{relaxed filtering}   & 0.505 \textbf{(-0.730)} & 0.554 \textbf{(-0.425)} & 9.536 \textbf{(-3.199)} & 2.363 \textbf{(-1.096)} & 0.460 \textbf{(-0.467)} \\
\bottomrule
\end{tabular}
}
\vspace{-0.1in}
\end{table}

\begin{table}[H]
\centering
\caption{\textbf{Surviving proportion (\%) of CoTs} on the train split of MMLU-Pro. 
We compare \GenORM and \GenPRM under (i) \textbf{label refinement} using \href{https://blog.google/technology/google-deepmind/google-gemini-ai-update-december-2024/}{Gemini-2.0 Flash}~\citep{comanici2025gemini}, and (ii) \textbf{relaxed} \textit{consensus filtering}. Please see the caption of \Cref{tab:wasserstein_distance} for more details.}
\label{tab:survival_proportion}
\vspace{-0.1in}
\small
\resizebox{0.9\textwidth}{!}{%
\begin{tabular}{llllll}
\toprule
 & \textbf{Overall} & \textbf{Law} & \textbf{Psychology} & \textbf{Chemistry} & \textbf{Biology} \\
\midrule
\GenORM                       & 51.1 & 51.6 & 28.3 & 71.9 & 42.0 \\
\GenPRM           & 28.0 & 22.7 & 22.8 & 30.1 & 30.4 \\
\GenPRM + \textbf{label refinement} 
  & 44.0 \textbf{(+16.0)}
  & 23.6 \textbf{(+0.9)}
  & 42.1 \textbf{(+19.3)}
  & 33.5 \textbf{(+3.4)}
  & 55.4 \textbf{(+25.0)} \\
\GenPRM + \textbf{relaxed filtering} 
  & 45.5 \textbf{(+17.5)}
  & 24.9 \textbf{(+2.2)}
  & 43.7 \textbf{(+20.9)}
  & 35.8 \textbf{(+5.7)}
  & 54.6 \textbf{(+24.2)} \\
\midrule
 & \textbf{Physics} & \textbf{History} & \textbf{Economics} & \textbf{Math} & \textbf{Business} \\
\midrule
\GenORM                       & 77.0 & 28.4 & 50.9 & 70.3 & 54.3 \\
\GenPRM           & 34.2 & 26.0 & 37.0 & 30.8 & 24.7 \\
\GenPRM + \textbf{label refinement} 
  & 35.8 \textbf{(+1.6)}
  & 49.3 \textbf{(+23.3)}
  & 64.9 \textbf{(+27.9)}
  & 65.5 \textbf{(+34.7)}
  & 56.0 \textbf{(+31.3)} \\
\GenPRM + \textbf{relaxed filtering} 
  & 37.3 \textbf{(+3.1)}
  & 41.0 \textbf{(+15.0)}
  & 64.1 \textbf{(+27.1)}
  & 66.7 \textbf{(+35.9)}
  & 57.8 \textbf{(+33.1)} \\
\midrule
 & \textbf{Philosophy} & \textbf{Health} & \textbf{Engineering} & \textbf{Computer Science} & \textbf{Other} \\
\midrule
\GenORM                       & 48.6 & 40.5 & 37.0 & 70.1 & 42.9 \\
\GenPRM           & 26.8 & 27.7 & 13.5 & 38.9 & 31.0 \\
\GenPRM + \textbf{label refinement} 
  & 46.6 \textbf{(+19.8)}
  & 51.3 \textbf{(+23.6)}
  & 13.9 \textbf{(+0.4)}
  & 46.9 \textbf{(+8.0)}
  & 37.5 \textbf{(+6.5)} \\
\GenPRM + \textbf{relaxed filtering} 
  & 48.0 \textbf{(+21.2)}
  & 58.7 \textbf{(+31.0)}
  & 14.1 \textbf{(+0.6)}
  & 48.8 \textbf{(+9.9)}
  & 38.6 \textbf{(+7.6)} \\
\bottomrule
\end{tabular}
}
\end{table}

\begin{table}[H]
\centering
\caption{Best-of-$N$ results on MMLU-Pro using \href{https://huggingface.co/meta-llama/Llama-3.1-8B-Instruct}{Llama-3.1-8B-Instruct} with \textbf{label refinement} or \textbf{filtering relaxation}. We use \href{https://huggingface.co/deepseek-ai/DeepSeek-R1-Distill-Qwen-14B}{R1-Distill-Qwen-14B} as the backbone for reward models. The number in parentheses denotes the change after label refinement or filtering relaxation. Please see the caption of \Cref{tab:wasserstein_distance} for more details.}
\label{tab:label_refinement_and_relaxed_filtering}
\vspace{-0.1in}
\small
\resizebox{0.9\textwidth}{!}{
\begin{tabular}{llllll}
\toprule
\multirow{2}{*}{\textbf{Method}} 
  & \multicolumn{5}{c}{$N$} \\
\cmidrule(lr){2-6}
  & \textbf{1} & \textbf{2} & \textbf{4} & \textbf{8} & \textbf{16} \\
\midrule
Majority voting
& 50.27
& 50.27
& 54.15
& 56.14
& 57.16 \\
\midrule
\DisORM
& 50.27
& 57.30
& 61.54
& 63.95
& 65.38 \\
\DisPRM
& 50.27
& 57.18
& 61.51
& 64.10
& 65.55 \\
\DisPRM + \textbf{label refinement}  
& 50.27 
& 56.99 \textbf{(-0.19)} 
& 61.41 \textbf{(-0.10)} 
& 64.38 \textbf{(+0.28)} 
& 66.57 \textbf{(+1.02)} \\
\midrule
\GenORM
& 50.27
& 58.24
& 63.88
& 67.82
& 70.02 \\
\GenPRM
& 50.27
& 55.24
& 59.06
& 62.10
& 64.26 \\
\GenPRM + \textbf{label refinement}
& 50.27 
& 54.99 \textbf{(-0.25)} 
& 58.86 \textbf{(-0.20)} 
& 62.10 \textbf{(+0.00)} 
& 64.84 \textbf{(+0.58)} \\
\GenPRM + \textbf{relaxed filtering}
& 50.27 
& 54.93 \textbf{(-0.31)} 
& 58.88 \textbf{(-0.18)} 
& 62.23 \textbf{(+0.13)} 
& 64.82 \textbf{(+0.56)} \\
\midrule
Pass@$N$
& 50.27
& 61.74
& 71.56
& 79.77
& 86.05 \\
\bottomrule
\end{tabular}
}
\end{table}

\begin{table}[H]
\centering
\caption{Best-of-$N$ results on MMLU-Pro using CoTs generated by \href{https://huggingface.co/HuggingFaceTB/SmolLM3-3B}{SmolLM3-3B}. We use \href{https://huggingface.co/deepseek-ai/DeepSeek-R1-Distill-Qwen-14B}{R1-Distill-Qwen-14B} backbone for reward models.}
\vspace{-0.1in}
\label{tab:mmlu_pro_aggregation_smollm3_3b}
\small
\begin{tabular}{lccccc}
\toprule
\multirow{2}{*}{\textbf{Method}} & \multicolumn{5}{c}{$N$} \\
\cmidrule(lr){2-6}
 & \textbf{1} & \textbf{2} & \textbf{4} & \textbf{8} & \textbf{16} \\
\midrule
MV & 40.42 & 40.42 & 44.45 & 46.63 & 48.35 \\
\midrule
\DisORM & 40.42 & 48.46 & 54.47 & 58.79 & 61.13 \\
\DisPRM (min) & 40.42 & 48.00 & 53.80 & 57.95 & 60.32 \\
\DisPRM (prod) & 40.42 & 47.51 & 53.31 & 57.62 & 60.09 \\
\DisPRM (mean) & 40.42 & 46.82 & 51.85 & 55.48 & 57.82 \\
\DisPRM (last) & 40.42 & 48.03 & 53.54 & 57.04 & 58.61 \\
\midrule
\GenORM & 40.42 & 49.55 & 56.81 & 62.14 & 65.29 \\
\GenPRM & 40.42 & 46.34 & 51.72 & 56.29 & 59.66 \\
\midrule
Pass@$N$ & 40.42 & 53.04 & 65.24 & 76.04 & 83.77 \\
\bottomrule
\end{tabular}
\end{table}

\begin{table}[H]
\centering
\caption{Best-of-$N$ results on MMLU-Pro using CoTs generated by \href{https://huggingface.co/Qwen/Qwen2.5-7B-Instruct}{Qwen2.5-7B-Instruct}. We use \href{https://huggingface.co/deepseek-ai/DeepSeek-R1-Distill-Qwen-14B}{R1-Distill-Qwen-14B} backbone for reward models.}
\vspace{-0.1in}
\label{tab:mmlu_pro_aggregation_qwen2_5_7b}
\small
\begin{tabular}{lccccc}
\toprule
\multirow{2}{*}{\textbf{Method}} & \multicolumn{5}{c}{$N$} \\
\cmidrule(lr){2-6}
 & \textbf{1} & \textbf{2} & \textbf{4} & \textbf{8} & \textbf{16} \\
\midrule
MV & 56.97 & 56.97 & 59.82 & 61.22 & 62.51 \\
\midrule
\DisORM & 56.97 & 62.08 & 65.19 & 66.88 & 67.96 \\
\DisPRM (min) & 56.97 & 61.89 & 64.98 & 66.83 & 68.02 \\
\DisPRM (prod) & 56.97 & 61.89 & 64.97 & 66.69 & 67.74 \\
\DisPRM (mean) & 56.97 & 61.02 & 63.62 & 65.28 & 66.48 \\
\DisPRM (last) & 56.97 & 61.90 & 64.84 & 66.41 & 67.42 \\
\midrule
\GenORM & 56.97 & 62.57 & 66.34 & 68.59 & 70.06 \\
\GenPRM & 56.97 & 60.88 & 63.80 & 65.75 & 67.44 \\
\midrule
Pass@$N$ & 56.97 & 66.05 & 73.58 & 79.52 & 84.23 \\
\bottomrule
\end{tabular}
\end{table}

\begin{table}[H]
\centering
\caption{Best-of-$N$ results on MMLU-Pro using CoTs generated by \href{https://huggingface.co/meta-llama/Llama-3.1-8B-Instruct}{Llama-3.1-8B-Instruct}. We use \href{https://huggingface.co/deepseek-ai/DeepSeek-R1-Distill-Qwen-14B}{R1-Distill-Qwen-14B} backbone for reward models.}
\vspace{-0.1in}
\label{tab:mmlu_pro_aggregation_llama3_1_8b}
\small
\begin{tabular}{lccccc}
\toprule
\multirow{2}{*}{\textbf{Method}} & \multicolumn{5}{c}{$N$} \\
\cmidrule(lr){2-6}
 & \textbf{1} & \textbf{2} & \textbf{4} & \textbf{8} & \textbf{16} \\
\midrule
MV & 50.27 & 50.27 & 54.19 & 56.28 & 57.54 \\
\midrule
\DisORM & 50.27 & 57.33 & 61.72 & 63.99 & 64.93 \\
\DisPRM (min) & 50.27 & 56.93 & 61.19 & 63.63 & 64.88 \\
\DisPRM (prod) & 50.27 & 56.73 & 60.96 & 63.54 & 64.91 \\
\DisPRM (mean) & 50.27 & 55.88 & 59.64 & 62.01 & 63.56 \\
\DisPRM (last) & 50.27 & 57.03 & 61.14 & 63.34 & 64.19 \\
\midrule
\GenORM & 50.27 & 58.14 & 63.76 & 67.57 & 69.96 \\
\GenPRM & 50.27 & 55.13 & 58.93 & 61.96 & 64.34 \\
\midrule
Pass@$N$ & 50.27 & 61.73 & 71.61 & 79.74 & 85.88 \\
\bottomrule
\end{tabular}
\end{table}

\begin{table}[H]
\centering
\caption{Best-of-$N$ results on MMLU-Pro using CoTs generated by \href{https://huggingface.co/google/gemma-2-9b-it}{gemma2-9b-it}. We use \href{https://huggingface.co/deepseek-ai/DeepSeek-R1-Distill-Qwen-14B}{R1-Distill-Qwen-14B} backbone for reward models.}
\vspace{-0.1in}
\label{tab:mmlu_pro_aggregation_gemma2_9b}
\small
\begin{tabular}{lccccc}
\toprule
\multirow{2}{*}{\textbf{Method}} & \multicolumn{5}{c}{$N$} \\
\cmidrule(lr){2-6}
 & \textbf{1} & \textbf{2} & \textbf{4} & \textbf{8} & \textbf{16} \\
\midrule
MV & 52.44 & 52.46 & 54.09 & 54.80 & 55.01 \\
\midrule
\DisORM & 52.44 & 56.32 & 58.96 & 60.78 & 61.93 \\
\DisPRM (min) & 52.44 & 56.17 & 58.66 & 60.18 & 61.27 \\
\DisPRM (prod) & 52.44 & 56.19 & 58.68 & 60.20 & 61.25 \\
\DisPRM (mean) & 52.44 & 55.45 & 57.62 & 59.16 & 60.08 \\
\DisPRM (last) & 52.44 & 56.09 & 58.32 & 59.63 & 60.43 \\
\midrule
\GenORM & 52.44 & 56.85 & 60.20 & 62.78 & 64.59 \\
\GenPRM & 52.44 & 55.74 & 58.46 & 60.64 & 62.22 \\
\midrule
Pass@$N$ & 52.44 & 59.17 & 64.90 & 69.68 & 73.18 \\
\bottomrule
\end{tabular}
\end{table}

\begin{table}[H]
\centering
\caption{Best-of-$N$ results on MMLU-Pro using CoTs generated by \href{https://huggingface.co/meta-llama/Llama-3.1-70B-Instruct}{Llama-3.1-70B-Instruct}. We use \href{https://huggingface.co/deepseek-ai/DeepSeek-R1-Distill-Qwen-14B}{R1-Distill-Qwen-14B} backbone for reward models.}
\vspace{-0.1in}
\label{tab:mmlu_pro_aggregation_llama3_1_70b}
\small
\begin{tabular}{lccccc}
\toprule
\multirow{2}{*}{\textbf{Method}} & \multicolumn{5}{c}{$N$} \\
\cmidrule(lr){2-6}
 & \textbf{1} & \textbf{2} & \textbf{4} & \textbf{8} & \textbf{16} \\
\midrule
MV & 65.91 & 65.91 & 69.51 & 71.11 & 72.04 \\
\midrule
\DisORM & 65.91 & 70.20 & 72.05 & 72.46 & 72.18 \\
\DisPRM (min) & 65.91 & 70.06 & 71.92 & 72.60 & 72.63 \\
\DisPRM (prod) & 65.91 & 69.97 & 71.84 & 72.40 & 72.19 \\
\DisPRM (mean) & 65.91 & 69.30 & 71.05 & 71.90 & 72.36 \\
\DisPRM (last) & 65.91 & 70.01 & 71.74 & 72.13 & 71.85 \\
\midrule
\GenORM & 65.91 & 71.02 & 73.89 & 75.42 & 76.02 \\
\GenPRM & 65.91 & 69.22 & 71.38 & 72.88 & 73.79 \\
\midrule
Pass@$N$ & 65.91 & 74.98 & 81.76 & 86.84 & 90.21 \\
\bottomrule
\end{tabular}
\end{table}



\end{document}